\documentclass[acmjacm]{style-subdirectory/acmtrans2m}
\usepackage{makeidx}
\usepackage{times}
\usepackage{epsf}

\def\backnode{\mbox{\em back-node}}
\def\xml{\mbox{\em XML}}
\def\backconstraints{\mbox{\em back}}
\def\ksgchain{\mbox{$K^*N_{cF}$-{\em chain}}}
\def\ksgchains{\mbox{$K^*N_{cF}$-{\em chains}}}
\def\fnh{\mbox{\em fh}}
\def\distribute{\mbox{\em dist}}
\def\frun{\mbox{\em f-run}}
\def\fr{\mbox{\mbox{$f$-$r$}}}
\def\hf{\mbox{$H_{<\infty}$}}
\def\generator{\mbox{\em gen}}
\def\usesb1{\mbox{\em USES$\_B_1$}}
\def\partition{\mbox{\em PARTITION}}
\def\last{\mbox{\em Last}}
\def\partition{\mbox{\em PARTITION}}
\def\prefix{\mbox{\em pfx}}
\def\sprefix{\mbox{\em s-pfx}}
\def\subfset{\mbox{\em Subf}}
\def\lleq{\mbox{$\leq _{\ell}$}}
\def\slleq{\mbox{$< _{\ell}$}}
\def\incomparable{\mbox{\em incp}}
\def\xdlrcc{{\cal MTALC}({\cal D}_{\rcc8})}
\def\setrep{\mbox{\em set-rep}}
\def\tpc{T\oplus C}
\def\tpcf{(T\oplus C)^*}
\def\kproj{\mbox{$K${\em -proj}}}
\def\qproj{\mbox{$Q${\em -proj}}}
\def\infinity{\mbox{\em Inf}}
\def\iff{\mbox{iff}}
\def\wrt{\mbox{w.r.t.}}
\def\wrt{\mbox{w.r.t.}}
\def\alphabet{\Sigma (x,N_P,N_{cF})}
\def\alphabett{\Sigma (2^{\hf},N_P,x,K,N_{cF})}
\def\alphabettt{\Sigma (2^Q,N_P,x,K,N_{cF})}
\def\alphabt{\Sigma (Q,N_P,x,K,N_{cF})}
\def\alphabtt{\Sigma (S,N_P,x,K,N_{cF})}
\def\st{\mbox{s-t}}
\def\nd{\mbox{n-d}}
\def\lp{\mbox{\em LP}}
\def\clp{\mbox{\em CLP}}
\def\dclp{\mbox{\em dCLP}}
\def\sclp{\mbox{\em sCLP}}
\def\dsclp{\mbox{\em dsCLP}}
\def\clp{\mbox{\em CLP}}
\def\closure{\mbox{\em cl}}
\def\concretefeatures{\mbox{\em cFeatures}}
\def\ncf{\mbox{\em ncf}}
\def\abstractfeatures{\mbox{\em aFeatures}}
\def\naf{\mbox{\em naf}}
\def\primitiveconcepts{\mbox{\em pConcepts}}
\def\definedconcepts{\mbox{\em dConcepts}}
\def\lits{\mbox{\em ${\cal L}$it}}
\def\consts{\mbox{\em constr}}
\def\econcepts{\mbox{\em eConcepts}}
\def\feconcepts{\mbox{\em feConcepts}}
\def\reconcepts{\mbox{\em reConcepts}}
\def\fbranchingfactor{\mbox{\em fbf}}
\def\rbranchingfactor{\mbox{\em rbf}}
\def\branchingfactor{\mbox{\em bf}}
\def\rec{\mbox{\em rec}}
\def\af{\mbox{\em af}}
\def\branchingtuple{\mbox{\em bt}}
\def\csp{\mbox{\em CSP}}
\def\queue{\mbox{\em Queue}}

\def\dnf{\mbox{\em dnf}}
\def\gdnf{\mbox{\em DNF}}

\def\dnfone{\mbox{\em dnf1}}

\def\dnftwo{\mbox{\em dnf2}}

\def\pcea{\mbox{{\em pc}$\exists\forall$}}
\def\pceaone{\mbox{{\em pc}$\exists\forall$}}
\def\mtalc{{\cal MTALC}}
\def\mtalcpq{{\cal MTALC}_{p,q}}
\def\mtalczq{{\cal MTALC}_{0,q}}
\def\tnext{\bigcirc}
\def\rccats{\mbox{$\rcc8${\em -at}}}
\def\cdalgats{\mbox{$\cdalg${\em -at}}}

\def\topspace{{\cal TS}}
\def\rtopspace{{\cal RTS}}
\def\deuxdp{\mbox{2D}{\cal P}}

\def\xdl{{\cal MTALC}({\cal D}_x)}
\def\bnxdl{{\cal BNALC}({\cal D}_x)}
\def\xdlpq{{\cal MTALC}_{p,q}({\cal D}_x)}
\def\xdlzz{{\cal MTALC}_{0,0}({\cal D}_x)}
\def\xdlzo{{\cal MTALC}_{0,1}({\cal D}_x)}
\def\xdlztrcc{{\cal MTALC}_{0,2}({\cal D}_{\rcc8})}
\def\xdlzorcc{{\cal MTALC}_{0,1}({\cal D}_{\rcc8})}

\def\xdlzq{{\cal MTALC}_{0,q}({\cal D}_x)}
\def\alcf{{\cal ALCF}}
\def\pltl{{\cal PLTL}}

\def\ctl{{\cal CTL}}
\def\ctlstar{{\cal CTL}^*}

\def\rccdl{{\cal MTALC}({\cal D}_{\rcc8})}
\def\xdlzocda{{\cal MTALC}_{0,1}({\cal D}_{\cdalg})}
\def\xdlzoatra{{\cal MTALC}_{0,1}({\cal D}_{\atra})}

\def\xat{{\mbox{x-{\em at}}}}

\def\npx{{N_P^x}}

\def\rccdc{{\mbox{\em DC}}}
\def\rccec{{\mbox{\em EC}}}
\def\rcctpp{{\mbox{\em TPP}}}
\def\rccpo{{\mbox{\em PO}}}
\def\rcceq{{\mbox{\em EQ}}}
\def\rccntpp{{\mbox{\em NTPP}}}
\def\rcctppi{{\mbox{\em TPPi}}}
\def\rccntppi{{\mbox{\em NTPPi}}}
\def\north{{\mbox{\em No}}}
\def\northeast{{\mbox{\em NE}}}
\def\east{{\mbox{\em Ea}}}
\def\southeast{{\mbox{\em SE}}}
\def\south{{\mbox{\em So}}}
\def\southwest{{\mbox{\em SW}}}
\def\west{{\mbox{\em We}}}
\def\northwest{{\mbox{\em NW}}}
\def\equal{{\mbox{\em Eq}}}
\def\allenb{<}

\def\allenm{\mbox{{\em m}}}

\def\alleno{\mbox{{\em o}}}

\def\rcc8{\mbox{${\cal RCC}$8}}
\def\cdalg{{{\cal CD}}}

\def\alc{{\cal ALC}}
\def\alcd{{\cal ALC}({\cal D})}

\def\cdalg{{\cal CDA}}

\def\rel-alg{{\cal R}}

\def\atraats{\mbox{$\atra${\em -at}}}
\def\eee{\mbox{\em eee}}

\def\err{\mbox{\em err}}
\def\lel{\mbox{\em lel}}
\def\lll{\mbox{\em lll}}

\def\lrl{\mbox{\em lrl}}

\def\orl{\mbox{\em orl}}

\def\rll{\mbox{\em rll}}

\def\rol{\mbox{\em rol}}
\def\rrl{\mbox{\em rrl}}
\def\rro{\mbox{\em rro}}
\def\rrr{\mbox{\em rrr}}

\def\BBR{{\rm I\!R}}
\def\BBN{{\rm I\!N}}
\def\BBQ{{\rm Q\mkern-10.3mu I\mkern2.5mu}}
\def\intervals{\mbox{\em Intervals}}
\def\Box{{\sqcap\mkern-12.3mu\sqcup}}
\def\cqfd{\vrule height 1.2ex depth 0ex width 1.2ex}

\def\apra{{\cal CYC}_b}
\def\atra{{\cal CYC}_t}

\def\ro1{{\cal R}_{0,1}}

\def\dligne{\mbox{d-line}}

\def\co1{{\cal C}_{O,1}}

\def\deuxdo{\mbox{2D}{\cal O}}

\acmVolume{}
\acmNumber{}
\acmYear{03}
\acmMonth{}

\newcommand{\BibTeX}{{\rm B\kern-.05em{\sc i\kern-.025em b}\kern-.08em
    T\kern-.1667em\lower.7ex\hbox{E}\kern-.125emX}}

\title{Bridging the gap between modal temporal logics and constraint-based
  QSR\footnote{Qualitative Spatial Reasoning.} as an $\alcd$ spatio-temporalisation
  with weakly cyclic TBoxes}
\author{AMAR ISLI\\
  Fachbereich Informatik, Universit\"at Hamburg}
\begin{abstract}
The aim of this work is to provide a family of qualitative theories for spatial change
in general, and for motion of spatial scenes in particular. Motion of an
\hbox{$n$-object} spatial scene is seen as a change (over time) of the (qualitative)
spatial relations between the different objects involved in the scene ---if, for
instance, the spatial relations are those of the \hbox{well-known}
\hbox{Region-Connection} Calculus $\rcc8$, then the objects of the scene are seen as
regions of a topological space, and motion of the scene as a change in the $\rcc8$
relations on the different pairs of the objects. To achieve this, we consider a
spatio-temporalisation $\xdl$, of the well-known $\alcd$ family of Description Logics
(DLs) with a concrete domain:
the $\xdl$ concepts are interpreted over infinite $k$-ary trees, with the nodes
standing for time points;
the roles split into $m+n$ immediate-successor (accessibility) relations, which are
antisymmetric and serial, and of which $m$ are general, not necessarily functional,
the other $n$ functional;
the concrete domain ${\cal D}_x$ is generated by an \mbox{$\rcc8$-like} spatial
Relation Algebra (RA) $x$.
The (long-term) goal is to design a platform for the implementation of flexible and
efficient domain-specific languages for tasks involving spatial change. In order to capture the expressiveness of most modal temporal
logics encountered in the literature, we introduce weakly cyclic Terminological Boxes (TBoxes) of
$\xdl$, whose axioms capture the decreasing property of modal temporal operators. We show the important result that satisfiability of an $\xdl$ concept
with respect to a weakly cyclic TBox is decidable in nondeterministic
exponential time, by reducing it to the emptiness
problem of a weak alternating automaton augmented with spatial constraints, which we show to remain decidable, although the accepting
condition of a run involves, additionally to the standard case,
consistency of a CSP (Constraint Satisfaction Problem) potentially
infinite. The result provides a tableaux-like satisfiability procedure 
which we will discuss.
Finally, given the importance and cognitive plausibility of continuous 
change in the real physical world, we provide a discussion showing that our decidability result extends to the case where the nodes of the $k$-ary tree-structures are
interpreted as (durative) intervals, and each of the $m+n$ roles as the {\em meets} relation of
Allen's RA of interval relations.
\end{abstract}

\category{}{}{}


\begin{document}

\setcounter{page}{1}

\begin{bottomstuff}
This
    work was supported partly by the EU project ``{\em Cognitive Vision systems}"
    (CogVis), under grant {\em CogVis IST 2000-29375}.\\
Author's address: Amar Isli,
Fachbereich Informatik, Universit\"at Hamburg,
Vogt-K\"olln-Strasse 30, D-22527 Hamburg, Germany
(isli@informatik.uni-hamburg.de). 
\end{bottomstuff}

\maketitle

\newtheorem{theorem}{Theorem}[section]
\newtheorem{conjecture}[theorem]{Conjecture}
\newtheorem{corollary}[theorem]{Corollary}
\newtheorem{proposition}[theorem]{Proposition}
\newtheorem{lemma}[theorem]{Lemma}
\newtheorem{discussion}[theorem]{Discussion}
\newtheorem{fact}[theorem]{Fact}
\newdef{definition}[theorem]{Definition}
\newdef{remark}[theorem]{Remark}
\newdef{example}[theorem]{Example}
\section{Introduction}\label{sect1}
Modal temporal logics are well-known in computer science in general, and in
Artificial Intelligence (AI) in particular. Important issues that need to
be addressed, when defining a modal temporal logic, include the ontological
issue of whether to choose points or intervals as the primitive objects, and the issue of whether
time is a total (linear) or a partial order. For more details, the reader is
referred to books such as van Benthem's \cite{vanBenthem83a}, or to survey
articles such as Vila's \cite{Vila94a}.

Qualitative Spatial Reasoning (henceforth QSR), and more generally Qualitative Reasoning
(QR), differs from quantitative reasoning by its particularity of
remaining at a description level as high as possible. In other words, QSR
sticks at the idea of ``making only as many distinctions as necessary''
\cite{Cohn97b}, idea borrowed to na\"{\i}ve physics
\cite{Hayes85b}. The core motivation behind this is that, whenever the
number of distinctions that need to be made is finite, the reasoning issue
can get rid of the calculation details of quantitative models, and be
transformed into a simple matter of symbol manipulation; in the particular
case of constraint-based QSR, this generally means a finite
RA \cite{Tarski41b} (see also
\cite{Duentsch99a,Ladkin94a,Maddux02a}), with tables recording the results of applying
the different operations to the different atoms, and the reasoning issue
reduced to a matter of table look-ups: a good illustration to this is the
well-known topological calculus $\rcc8$
\cite{Randell92a,Egenhofer91b}. One plausible way of responding to criticisms
against QSR languages (which include
Forbus, Nielsen and Faltings' \cite{Forbus91a} poverty conjecture, and Habel's
\cite{Habel95a} argument that
such languages suffer from not having ``the ability to refine discrete structures if
necessary''), and against QR languages in general, is to
define such languages according to cognitive adequacy criteria
\cite{Renz00a}. For more
details on QSR, and on QR in general, the reader is referred to survey articles such
as \cite{Cohn97b,Dague95a}.

Considered separately, modal temporal logics and constraint-based QSR have
each an important place in AI. However, an important goal for research in AI,
which has not received enough attention so far, is to define well-founded
languages combining modal temporal logics and QSR languages. Such languages
are key representational tools for tasks involving qualitative spatial change.
Examples of such tasks include
satellite-like surveillance of large-scale geographic spaces, and
(qualitative) path planning for robot navigation
\cite{Escrig98b,Freksa92b,Isli02b,Latombe91a,Schlieder93a,Zimmermann96a}.

The goal of the present work is to enhance the expressiveness of modal
temporal logics with qualitative spatial constraints. What we get is a
family of qualitative theories for spatial change in general, and for
motion of spatial scenes in particular. The family consists of
domain-specific spatio-temporal (henceforth $\st$) languages, and is
obtained by spatio-temporalising a well-known family of description logics (DLs)
with a concrete domain, known as $\alcd$ \cite{Baader91a}. $\alcd$
originated from a pure DL known as $\alc$
\cite{Schmidt-Schauss91a}, with $m\geq 0$ roles all of which are
general, not necessarily functional relations, and which Schild
\cite{Schild91a} has shown to be expressively equivalent to Halpern
and Moses' ${\cal K}_{(m)}$ modal logic \cite{Halpern85a}. $\alcd$
is obtained by adding to $\alc$
functional roles (better known as abstract features), a concrete
domain ${\cal D}$, and concrete features (which refer to objects of
the concrete domain). The spatio-temporalisation of $\alcd$ is
obtained, as the name suggests, by performing two specialisations at the same
time:
\begin{enumerate}
  \item temporalisation of the roles, so that they consist of $m+n$
    immediate-successor (accessibility) relations
    $R_1,\ldots R_m,f _1,\ldots ,f _n$, of which the $R_i$'s are general, not
    necessarily functional relations, and the $f_i$'s functional relations;
    and
  \item spatialisation of the concrete domain ${\cal D}$: the concrete domain
    is now ${\cal D}_x$, and is  generated  by a spatial RA $x$, such as the
    Region-Connection Calculus RCC8 \cite{Randell92a}.
\end{enumerate}
The final spatio-temporalisation of $\alcd$ will be
referred to as $\xdl$ ($\mtalc$ for \mbox{${\cal M}$odal}
\mbox{${\cal T}$emporal} $\alc$). To summarise, $\xdl$ verifies the
following:
\begin{enumerate}
  \item the (abstract) domain (i.e., the set of worlds in modal
    logics terminology) of $\xdl$ interpretations is a universe of
    time points;
  \item the roles consist of $m+n$ immediate-successor
    relations
    $R_1,\ldots R_m,f_1,\ldots ,f_n$, of
    which the $R_i$'s are general, not necessarily functional
    relations, and the $f_i$'s are functional relations;
  \item the roles are antisymmetric and serial, and we
    denote, as usual, the transitive closure and the
    reflexive-transitive closure of a relation $R$ by $R^+$ and
    $R^*$, respectively;
  \item the concrete domain ${\cal D}_x$ is generated by an
    $\rcc8$-like constraint-based qualitative spatial language $x$; and
  \item the concrete features are functions from the abstract domain
    onto objects of the concrete domain: in the case of $x$ being
    $\rcc8$, for instance, the objects of the concrete domain are
    regions of a topological space.
\end{enumerate}
When viewed as a domain-specific high-level vision system, a theory of
the family has the following properties:
\begin{enumerate}
  \item ``the eyes'' of the system are the concrete features: with each
    object $O_i$ of the (changing) spatial domain at hand, we associate
    one and only one concrete feature $g_i$, which is given the task of
    ``keeping an eye'' on $O_i$'s position as time flows;
  \item the (concrete) feature chains other than the concrete features,
    which consist of finite chains (compositions) of abstract features
    terminated by a concrete feature, and allow to access from a given
    node of an interpretation, the value of a concrete feature at
    another, future node, constitute ``the predictive'' engine of the
    system;
  \item the objects of the concrete domain are concrete objects of the
    spatial domain at hand; and
  \item the predicates of the concrete domain constitute ``the
    high-level memory'' of the system, able of representing knowledge on
    objects of the spatial domain at hand, as seen by the concrete
    features or predicted by the feature chains, as spatial constraints
    on tuples of the corresponding concrete features or feature chains.
\end{enumerate}
The idea of domains in LP (Logic Programming)
\cite{BenhamouF93a,Colmerauer90a,Jaffar87a,Jaffar94a,vanHentenryck89a}
has led to CLP (Constraint Logic Programming) with a specific domain,
such as $\clp (\BBQ )$ and $\clp (\BBN )$, known to be efficient
implementations of CLP with, respectively, the rationals and the
integers as a specific domain.
The motivation behind the extension of $\alc$ to $\alcd$
was similar, in that with $\alcd$ we can refer directly to objects
of the domain we are interested in, thanks to the concrete features,
and to the (concrete) feature chains in general, and record knowledge
on these objects, as ``seen'' by the concrete features or ``predicted''
by the feature chains, thanks to the predicates. This allows the
objects of the domain at hand, and the knowledge on them, to be
isolated from the knowledge on the abstract objects, sufficiently
enough to allow for an easy and efficient implementation. In this
respect, $\mtalc ({\cal D}_{\rcc8})$, for instance, will be an
implementation of $\mtalc$, the temporal component, with a concrete
domain ${\cal D}_{\rcc8}$ generated by $\rcc8$.

$\xdl$ is the result of combining a temporalisation of a pure DL
language, $\alc$ \cite{Schmidt-Schauss91a}, with a spatialisation of a
constraint-based language reflected by a concrete domain ${\cal D}$.
The discussion of the previous paragraph, on the separation of the
knowledge on the objects of the domain at hand, from the knowledge on
the abstract objects, leads to an important computational advantage in
the use of this way of getting spatio-temporal languages, instead of
combining two or more modal logics, which is known to potentially lead
to undecidable spatio-temporal languages ---even when the combined
parts are tractable \cite{Bennett02a,Bennett02b}! With DLs, it is known
that, as long as the pure DL and the constraint-based language reflected
by the concrete domain are decidable, the resulting DL with a concrete
domain is so that satisfiability of a concept $\wrt$ an acyclic TBox
is decidable \cite{Baader91a}.

Constraint-based languages candidate for generating a concrete domain
for a member of our family of spatio-temporal theories, are spatial RAs
for which the atomic relations form a decidable subset ---i.e., such
that consistency of a CSP expressed as a conjunction of $n$-ary
relations on $n$-tuples of objects, where $n$ is the arity of the RA
relations, is decidable. Examples of such RAs known in the literature
include,
the Region-Connection Calculus $\rcc8$ in \cite{Randell92a} (see also
\cite{Egenhofer91b}),
the Cardinal Directions Algebra $\cdalg$ in \cite{Frank92b},
and the rectangle algebra in \cite{Balbiani98a} (see also
\cite{Guesgen89a,Mukerjee90a}) for the binary case;
and
the RA $\atra$ of 2D orientations in \cite{Isli98a,Isli00b} for the
ternary case.

The paper, without loss of generality, will focus on two concrete
domains generated by two of the three binary spatial RAs mentioned above,
$\rcc8$ \cite{Randell92a} and $\cdalg$ \cite{Frank92b}; and on a third
concrete domain generated by the ternary spatial RA $\atra$ in
\cite{Isli98a,Isli00b}. It is known that, in the general case,
satisfiability of an $\alcd$ concept with respect to a cyclic
Terminological Box (TBox) is undecidable (see, e.g., \cite{Lutz01b}). In
this work, we define a weak form of TBox cyclicity,\footnote{Intuitively,
a TBox is weakly cyclic if all its cycles are of degree 1, reflected by a
defined concept appearing in the right hand side of the axiom defining
it (within the scope of an exitential or universal quantifier).} which captures the decreasing property of modal temporal operators.
The pure DL $\mtalc$, consisting of the temporal component of $\xdl$,
together with weakly cyclic TBoxes, captures the expressiveness of most
modal temporal logics encountered in the literature. Similarly to
{\em eventuality} formulas in modal temporal logics, some of the defined
concepts of $\xdl$ will be referred to as {\em eventuality} concepts,
for the axioms defining them describe situations that need to be
effectively satisfied sometime in the future. As an example of such
defined concepts, the concept $C$ defined by the axiom
$C\doteq p\sqcup \exists f.C$, where $p$ is an atomic proposition, and $f$
an immediate-successor accessibility relation, associating with each state
of an interpretation its immediate successor, describes the eventuality
formula $\diamondsuit p$, of, say, Propositional Linear Temporal Logic
(PLTP). For a state $s$ of an interpretation to satisfy the eventuality
formula $\diamondsuit p$, there should exist a descendent node of $s$,
along the infinite path $f^\omega$, satisfying $p$. The axiom
$C\doteq p\sqcup\exists f.C$, however, may leave $p$ unsatisfied, and
still give the false impression that it is satisfied: this is a well-known
situation in modal temporal logics, which may happen by eternally
reporting the satisfiability to the next state, and which can be get rid
of by having recourse to the theory of automata on infinite objects (see,
e.g., \cite{Isli96d,Vardi86a}). This discussion is important for the
understanding of the way are obtained the accepting states of the (weak)
alternating automaton on infinite objects, associated with the
satisfiability of an $\xdl$ concept $\wrt$ an $\xdl$ weakly cyclic TBox.

In a nondeterministic automaton on infinite words, the transition function, say $\delta$, is
so that $\delta (a,q)$, where $a$ is a letter of the alphabet and $q$ an
element of the set $Q$ of states, is a subset of $Q$; and a state $q$
accepts a word $u=av$, if and only if ($\iff$) there exists a state
$q'\in\delta (a,q)$ accepting the suffix $v$. In alternating
automata, $\delta (a,q)$, when put in a certain normal form, is a set of
subsets of $Q$; and a state $q$ accepts a word $u=av$, $\iff$ there exists a
set of states $Q'\in\delta (a,q)$, such that each state $q'$ in $Q'$
accepts the suffix $v$. This intuitive definition of acceptance works only
when no condition is imposed on the so called run of the automaton on the
the input (infinite) word. When such a condition is imposed, the acceptance
requires more: in the case of B\"uchi automata, for instance, where the
accepting condition is given by a subset ${\cal F}$ of the set of states,
the states repeated infinitely often in any branch of the run should include a state of ${\cal F}$.

The theory of alternating automata \cite{Muller87a} on
infinite words, and on infinite trees in general, is a
generalisation of the theory of nondeterministic automata \cite{Rabin69a}.
In this work, we will mostly need weak alternating automata \cite{Muller92a}.

Furthermore, we will need to extend standard alternating automata, in order to
handle interpretations ``following'' the evolution of a spatial scene
of interest over time, by recording at each node, thanks to concrete
features, the positions of the objects of the scene. The new kind of
alternating automata handle spatial constraints on concrete features,
and on (concrete) feature chains in general, which allow them to restrict the values of the
different concrete features at the different nodes of an
interpretation. As a consequence, the
accepting condition will involve, not only the states infinitely
often repeated in a run, but a CSP as well, which is potentially
infinite. We show that the ``injection'' of spatial constraints into standard
weak alternating automata does not compromise decidability of the
emptiness problem, which we show can be achieved in nondeterministic
exponential time.

We prove the important result that, satisfiability of an $\xdl$ concept
with respect to an $\xdl$ weakly cyclic TBox is decidable, by reducing it to the
emptiness problem of a weak alternating automaton augmented with
spatial constraints. A first discussion will show that the result provides an effective tableaux-like
satisfiability procedure, with the particularity of dynamically
handling spatial constraints, using constraint propagation techniques,
which allows it to potentially reduce the search space.
A second discussion will clarify how the decidability result extends to the case where the nodes of the $k$-ary
tree-structures are interpreted as (durative) intervals, and each of the $m+n$
roles as the {\em meets} relation of Allen's RA of interval relations \cite{Allen83b}.

Weakly definable languages of infinite words (or $\omega$-words), and of
infinite $k$-ary trees in general, were first defined by Rabin
\cite{Rabin70a}: a language $L$ is weakly definable $\iff$ $L$ and its complement
$\overline{L}$ are B\"uchi ($L$ is B\"uchi $\iff$ there exists a B\"uchi automaton
accepting $L$). Muller, Saoudi and Schupp \cite{Muller92a} have shown
(1) that a language is weakly definable $\iff$ there exists a weak alternating
automaton accepting it; and
(2) that a weak alternating automaton can be simulated with a B\"uchi nondeterministic
automaton (see also \cite{Muller95a}).
The emptiness problem of a B\"uchi nondeterministic automaton, in turn, is known to be
trivially decidable \cite{Rabin69a}.

Computing with alternating automata is easy. In particular, the complement
of a language accepted by an alternating automaton $M$, is the language of
the alternating automaton $M'$, obtained from $M$ by dualising the
transition function (interchanging the $\wedge$ and $\vee$ operators), and
complementing the accepting condition (see the complementation theorem in
\cite{Muller87a}).

The particularity of weak alternating automata is that, one can find a
partition $Q=Q_1\cup\cdots\cup Q_m$ of the set $Q$ of states, and a
partial order $\geq$ on this partition, so that if $q'\in\delta (a,q)$,
then there exist $i,j\in\{1,\cdots ,m\}$, such that
$q\in Q_i\wedge q'\in Q_j\wedge Q_i\geq Q_j$.
\subsection{The way from $\lp$ to distributed spatial $\clp$ ($\dsclp$)}
The theoretical framework of Logic Programming ($\lp$) is the propositional
calculus. A logic program can be seen as deciding satisfiability of a conjunction
of Horn clauses. In reality, however, one cannot always restrict the knowledge on
the real world, to Boolean combinations of atomic, Boolean propositions. One has
to face the problem of representing, and dealing with, knowledge on a specific
domain of interest, generally consisting of objects referred to by variables.
Such knowledge is represented using constraints of the form $P(x_1,\ldots ,x_n)$,
where $P$ is an $n$-ary predicate, and the $x_i$ are variables. Such constraints
are referred to as {\em domains} in $\clp$.

$\clp$ incorporates the idea of domains into $\lp$, so that, for instance, the
user can restrict the range of a variable, or of a pair of variables, either with
the use of a unary constraint (on a variable), or with the use of a binary
constraint (on a pair of variables). Search algorithms based on consistency
techniques \cite{BenhamouF93a,Colmerauer90a,Jaffar87a,Jaffar94a,vanHentenryck89a},
can then use a priori pruning during the search, generally by applying a filtering
algorithm, such as arc-consistency \cite{Mackworth77a,Montanari74a}, at each node
of the search tree, which potentially reduces the domains of the variables and of
the pairs of variables (thus reducing the subtree of the search space rooted at
the current node\footnote{By current node, we mean the node of the tree-like
search space where the search is.}). Modal temporal logics, in turn, can be seen
as a mean for distributing LP, which leads to distibuted LP (dLP). Adding
variable (and \mbox{pair-of-variable}) domains to dLP, in a similar way as in LP,
leads to distributed $\clp$ ($\dclp$). The general use, and understanding of, a
domain in $\clp$ and in $\dclp$, is as a unary constraint, of the form $R(x)$,
stating that variable $x$ is constrained to belong to the unary predicate $R$; or
as a binary constraint, of the form $S(x,y)$, stating that the pair $(x,y)$ of
variables is constrained to belong to the binary predicate $S$: $R$ is nothing
else than a subset of the whole universe of values used for variables'
instantiation, and $S$ a subset of the cross product of the universe by itself
(the universe is generally the set $\BBR$ of reals, or any of its subsets, such
as the set $\BBQ$ of rationals, the set $\BBN$ of integers, or the set $\{0,1\}$
symbolising the Booleans).

In QSR, restricting the domain of a variable to a strict subset of the (continuous) spatial domain
at hand (the two-dimensional space, for instance), has generally no practical importance. In other
words, unary constraints do not have as much importance as in traditional $\clp$. Indeed, a
conjunction of QSR constraints (in other words, a QSR CSP) is always node- and arc-consistent.
Examples of QSR constraint languages for which this is the case, include languages of binary
relations we have already mentioned: the Region-Connection Calculus in \cite{Randell92a,Egenhofer91b}, the
Cardinal Directions Algebra in \cite{Frank92b} and the rectangle algebra in
\cite{Balbiani98a,Guesgen89a,Mukerjee90a}. For these languages, the lowest local-consistency
filtering, which can potentially reduce the search space, is path consistency. This is not the end of
the story: there are QSR constraint languages for which CSPs expressed in them are already strongly
$3$-consistent (i.e., node-, arc- and path-consistent), and for which effectiveness of
local-consistency filtering starts from $4$-consistency (for instance, the ternary RA of 2D
orientations in \cite{Isli98a,Isli00b}).

It should be clear now that, in order to be able to reason qualitatively about space within $\clp$
or within $\dclp$, one has to adapt it, so that the domains reflect the reality of the spatial
domain at hand: the domains should be binary constraints, if the constraint language used for
representing spatial knowledge is binary (such as the three mentioned above), and ternary
constraints, if the language is ternary (such as the one mentioned above). To make things clearer,
suppose that the representational language is $\rcc8$. The spatial domain at hand, used for
instantiating the spatial variables, is then the set of regions of a topological space. A pair of
spatial variables of a $\clp$ program, if no restriction is given, is
related by the $\rcc8$ universal
relation, which is the set of the eight atoms of the language. A domain in this case is any subset
of the set of all eight atoms (in other words, any $\rcc8$ relation), which will restrict the
instantiations of pairs of variables, to those pairs of regions of the topological space at hand
that are related by an atom of the domain.

We refer to $\clp$ and $\dclp$ with spatial variables as described above, as $\sclp$ (spatial
$\clp$) and $\dsclp$ (distributed $\sclp$).
\subsection{Associating a weak alternating automaton with the satisfiability of a concept $\wrt$ a
weakly cyclic TBox: an overview}\label{overview}
Given an $\xdl$ concept $C$ and an $\xdl$ weakly cyclic TBox ${\cal T}$, the problem we will be interested in
is, the satisfiability of $C$ with respect to ${\cal T}$. The axioms in ${\cal T}$ are of the form $B\doteq E$,
where $B$ is a defined concept name, and $E$ an $\xdl$ concept. Using $C$, we introduce a new defined
concept name, $B_{init}$, given by the axiom $B_{init}\doteq C$. We denote by ${\cal T}'$ the TBox consisting
of ${\cal T}$ augmented with the new axiom: ${\cal T}'={\cal T}\cup\{B_{init}\doteq C\}$. The alternating automaton we
associate with the satisfiability of $C$ $\wrt$ the TBox ${\cal T}$, so that satisfiability holds $\iff$ the
language accepted by the automaton is not empty, is now almost entirely given by the TBox ${\cal T}'$: the
defined concept names represent the states of the automaton, $B_{init}$ being the initial state; the
transition function is given by the axioms themselves. However, some modification of the axioms is needed.

Given an $\xdl$ axiom $B\doteq E$ in ${\cal T}'$, the method to be proposed decomposes $E$ into some kind of
Disjunctive Normal Form, $\dnf2 (E)$, which is free of occurrences of
the form $\forall R.E$. Intuitively, the concept $E$ is satisfiable by the state consisting
of the defined concept name $B$, $\iff$ there exists an element $S$ of $\dnf2 (E)$ that is satisfiable by
$B$. An element $S$ of $\dnf2 (E)$ is a conjunction written as a set, of the form
$S_{prop}\cup S_{csp}\cup S_{\exists}$, where:
\begin{enumerate}
  \item $S_{prop}$ is a set of primitive concepts and negated primitive concepts ---it is worth
noting here that, while the defined concepts (those concept names appearing as the left hand side of an axiom) define the
states of our automaton, the primitive concepts (the other concept names) correspond to atomic propositions in, e.g.,
classical propositional calculus;
  \item $S_{csp}$ is a set of concepts of the form $\exists (u_1)\cdots (u_n).P$,
    where $u_1,\ldots ,u_n$ are feature chains and $P$ a relation (predicate) of an $n$-ary
    spatial RA; and
  \item $S_{\exists}$ is a set of concepts of the form $\exists
    R.E_1$, where $R$ is a role and $E_1$ is a concept.
\end{enumerate}
The procedure ends
with a TBox ${\cal T}'$ of which all axioms are so written. Once ${\cal T}'$ has been so written, we denote:
\begin{enumerate}
  \item by $af({\cal T}')$, the set of abstract features appearing in ${\cal T}'$; and
  \item by $rrc({\cal T}')$, the set of concepts appearing in ${\cal T}'$, of the form $\exists R.E$, with $R$ being a
    general, not necessarily functional role, and $E$ a concept.
\end{enumerate}
The alternating automaton to be associated with ${\cal T}'$, will operate on (Kripke) structures which are
infinite $m+p$-ary trees, with $m=|af({\cal T}')|$ and $p=|rrc({\cal T}')|$. Such a structure, say $t$, is associated
with a truth-value assignment function $\pi$, assigning to each node, the set of those primitive concepts
appearing in ${\cal T}'$ that are true at the node. With $t$ are also associated the concrete features
appearing in ${\cal T}'$: such a concrete feature, $g$, is mapped at each node of $t$, to a (concrete) object
of the spatial domain in consideration (e.g., a region of a topological space if the concrete domain is
generated by $\rcc8$).

The feature chains are of the form $f_1\ldots f_kg$,\footnote{Throughout the rest of the paper, a
feature chain $f_1\ldots f_kg$ is interpreted as within the Description Logics Community ---i.e., as
the composition $f_1\circ\ldots\circ f_k\circ g$: we remind the user that
$(f_1\circ\ldots\circ f_k\circ g)(x)=g(f_k(f_{k-1}(\ldots (f_2(f_1(x))))))$.} with $k\geq 0$, where the
$f_i$'s are abstract features (also known, as alluded to before, as functional roles: functions from
the abstract domain onto the abstract domain), whereas $g$ is a concrete feature (a function from the
abstract domain onto the set of objects of the concrete domain). The sets $S$ are used to label the
nodes of the search space. Informally, a run of the \mbox{tableaux-like} search space is a
\mbox{disjunction-free} subspace, obtained by selecting at each node, labelled, say, with $S$, one
element of $\dnf2 (S)$.

Let $\sigma$ be a run, $s_0$ a node of $\sigma$, and $S$ the label of $s_0$, and suppose that $S_{csp}$
contains $\exists (u_1)(u_2).P$ (we assume, without loss of generality, a concrete domain generated by
a binary spatial RA, such as $\rcc8$ \cite{Randell92a,Egenhofer91b}), with
$u_1=f_1\ldots f_kg_1$ and $u_2=f _1'\ldots f _m'g_2$.
The concept $\exists (u_1)(u_2).P$ gives birth to new nodes of the run,
$s_1=f_1(s_0),s_2=f_2(s_1),\ldots ,s_k=f_k(s_{k-1}),
 s_{k+1}=f _1'(s_0),s_{k+2}=f _2'(s_{k+1}),\ldots ,s_{k+m}=f _m'(s_{k+m-1})$; to new variables of
what could be called the (global) CSP, $\csp (\sigma )$, of $\sigma$; and to a new constraint of
$\csp (\sigma )$. The new variables are $\langle s_k,g_1\rangle$ and
$\langle s_{k+m},g_2\rangle$, which denote the values of the
concrete features $g_1$ and $g_2$ at nodes $s_k$ and $s_{k+m}$, respectively. The new constraint is
$P(\langle s_k,g_1\rangle ,\langle s_{k+m},g_2\rangle)$.
The set of all such variables together with the set of all such constraints, generated by node $s_0$,
give the $\csp$ $\csp _{\sigma}(s_0)$ of $\sigma$ at $s_0$; and the union of all
$\csp$s $\csp _s({\sigma})$, over the nodes $s$ of $\sigma$, gives $\csp (\sigma )$. As the discussion
shows, $\dsclp$ does not reduce to a mere distribution of $\sclp$, consisting of $\sclp$ at each node,
with additionally temporal precedence on the different nodes: the
feature chains make it possible to refer to the values of the
different concrete features at the different nodes of a run, and restrict
these values using spatial predicates.

The pruning process during the tableaux method will now work as follows. The search will make use
of a data structure $\queue$, which will be handled in very much the same fashion as such a data
structure is handled in local consistency algorithms, such as arc- or path-consistency
\cite{Mackworth77a,Montanari74a}, in standard $\csp$s. The data structure is initially empty. Then
whenever a new node $s$ is added to the search space, the global $\csp$ of the run being
constructed is updated, by augmenting it with (the variables and) the constraints generated, as
described above, by $s$. Once the $\csp$ has been updated, so that it includes the local $\csp$ at
the current node, the local consisteny pruning is applied by propagating the constraints in
$\queue$.
Once a run has been fully
constructed, and only then, its global $\csp$ is solved. In the case of a concrete
domain generated by a binary, $\rcc8$-like RA, the filtering is achieved with a path-consisteny
algorithm \cite{Allen83b}, and the solving of the global $\csp$, after a run has been fully
constructed, with a solution search algorithm such as Ladkin and Reinefeld's \cite{Ladkin92a}. In the
case of a concrete domain generated by a ternary spatial RA, the filtering and the solving
processes are achieved with the 4-consistency and the search algorithms in \cite{Isli98a,Isli00b}.
\subsection{The relation to Bayesian networks}
In the case of feature chains of length one (i.e., reducing to concrete features), we will discuss
how to combine the predicate concepts, of the form $\exists (g_1)(g_2).P$, with conditional
probabilities, which will make the relation, at the current state, on a pair of concrete features,
dependant only on the relation on the same pair at the previous state: the conditional probabilities
will provide, for the relation on a pair of concrete features, the probability to be $s$, given that
it was $r$ at the previous state, $r$ and $s$ being atoms of the spatial RA $x$. This will give us a
family of $\alcd$-like languages, $\bnxdl$, for probabilistic, Bayesian-network-like reasoning (see,
e.g., \cite{Pearl00a,Russell03a}), about qualitative spatio-temporal knowledge. This is particularly important
for prediction \cite{Rimey92a} in, for instance, scene interpretation in high-level computer vision.
One possibility of setting the conditional probabilities is to learn them. Another is to assume
continuous change and uniform probability distribution: the conditional probabilities can then be
derived straightforwardly from what is known in QSR as the theory of
{\em conceptual neighbourhoods} (see, e.g., \cite{Freksa92a}).
\subsection{Related work}
\subsubsection{Motion and spatial change as $\st$ histories}
According to \cite{Hayes85a}, $\st$ histories are
space-time regions traced by objects over time. For the $n$-dimensional
($\nd$ for short) space, a $\st$ history is an $n+1$-d volume. Such a history
(of an object, or of a scene in general, of the $\nd$ space) can be recorded
by associating with the flow of time a camera ``filming'' the scene. The
approach we propose respects this view of spatial change, and of
motion in particular. Each member of our family of theories is an $\alcd$-like
DL, with temporalised roles, and a spatial concrete domain. The temporalised
roles allow the DL to capture the flow of time. The spatial concrete domain,
in some sense, plays the role of a camera:
\begin{enumerate}
  \item the concrete features can be seen, as already argued, as the eyes of
    the camera (one eye per object of the spatial scene at hand); and
  \item the knowledge on the spatial scene, as perceived by the camera's eyes,
    is (qualitatively) recorded by the predicates, as spatial constraints on
    tuples of the objects involved in the scene.
\end{enumerate}
Contrary to other approaches \cite{Hazarika01a,Hazarika02a,Muller98a}, ours
makes clear the borderline between the
temporal component and the spatial component. Indeed, the general $\alcd$ framework
\cite{Baader91a} was originally inspired by domain-specific Constraint Logic
Programming ($\clp$)
\cite{BenhamouF93a,Colmerauer90a,Jaffar87a,Jaffar94a,vanHentenryck89a}, which, as
already explained, gave
birth to many efficient and flexible domain-specific implementations of $\clp$ (for
instance, $\clp (\BBR)$, $\clp (\BBQ)$, $\clp (\BBN)$, $\clp (\intervals )$). Each
theory of our family can give birth to an efficient and flexible implementation of
$\dsclp$ with a specific spatial domain (each $\rcc8$-like RA can generate such a
domain).
\subsubsection{Approaches based on multi-dimensional modal logics}
Approaches based on multi-dimensional modal logics for the representation
of $\st$ knowledge, exist in the literature
\cite{Balbiani02a,Bennett02a,Bennett02b,Gabbay02a,Wolter00a}. Their main
disadvantage is that, their spatial component, for instance, can
represent only some specific spatial knowledge (e.g., topological knowledge). In our case, whenever a new $\rcc8$-like spatial RA is found, it can be
used to generate a spatial concrete domain, and augment our family with a new theory
for spatial change. The new theory, in turn, can be implemented as an efficient and
flexible domain-specific, $\clp$-like language for tasks of the $\st$ domain. If an
implementation of a theory of the family already exists, then the implementation of
the new theory only needs to adapt the old implementation to the new concrete
domain ---which does not require much work. Another disadvantage of multi-modal
logics is that, even combining tractable modal logics may lead
to an undecidable multi-modal logic (see, for instance, \cite{Bennett02a,Bennett02b}).
\subsection{Examples of potential applications}
\subsubsection{Geographical Information Systems (GIS)}
GIS is known to be one of the priviledged application domains of constraint-based
QSR (see, for instance, \cite{Cohn97b}). Among QSR languages that have GIS as a
direct application domain, the calculus of cardinal directions in \cite{Frank92b}
from the orientational side, and the $\rcc8$ calculus \cite{Randell92a,Egenhofer91b}
from the topological side. Each of these two languages, as already discussed, can
generate one member of our family $\xdl$ of qualitative theories for spatial
change, which can be used for geographic change.
\subsubsection{High-level computer vision}
By high-level computer vision (see, e.g., \cite{Neumann83a}), we mean that it is not necessary to have knowledge on
the precise location of the different parts of, say, the moving spatial scene. Rather, we are interested in describing, qualitatively, the relative position of the different parts. The language used for the description is a qualitative language of spatial relations, in the style of the binary RAs $\rcc8$
\cite{Randell92a,Egenhofer91b} and $\cdalg$ \cite{Frank92b}, and the ternary RA
$\atra$ in \cite{Isli98a,Isli00b}: qualitative knowledge on relative position of objects of the spatial domain at hand
is represented as constraints consisting of relations of the RA on $n$-tuples of the
objects, where $n$ is the arity of the relations of the RA.
\subsubsection{Qualitative path planning for robot navigation}
There are constraint-based spatial languages in the literature considered as
well-suited for path planning for robot navigation. These include Freksa's ternary
calculus \cite{Freksa92b,Zimmermann96a} of relative orientation of 2-d points, as well as Isli and Cohn's
ternary RA $\atra$ of 2-d orientations \cite{Isli98a,Isli00b}. This is however misleading, for the languages
are spatial, and not spatio-temporal. The best they can offer is the representation
of a snapshot of a spatial change, in particular the snapshot of a motion of, say, a
spatial scene. The reason to that is that the languages do not capture the flow of
time at all. However, each can be used to generate a spatial concrete
domain for a member of our family of theories of spatial change.
\subsection{Background on binary relations}
Given a set $A$, we denote by $|A|$ the cardinality of $A$.
A binary relation, $R$, on a set $S$ is any subset of the cross product
$S\times S=\{(x,y):x,y\in S\}$.
Such a relation is reflexive $\iff$ $R(x,x)$, for all $x\in S$;
it is symmetric $\iff$, for all $x,y\in S$, $R(y,x)$, whenever $R(x,y)$;
it is transitive $\iff$, for all $x,y,z\in S$, $R(x,z)$, whenever $R(x,y)$ and $R(y,z)$;
it is irreflexive $\iff$, for all $x\in S$, $\neg R(x,x)$;
it is antisymmetric $\iff$, for all $x,y\in S$, if $R(x,y)$ and $R(y,x)$ then $y=x$; and
it is serial $\iff$, for all $x\in S$, there exists $y\in S$ such that $R(x,y)$.
The transitive (resp. reflexive-transitive) closure of $R$ is the smallest relation
$R^+$ (resp. $R^*$), which includes $R$ and is transitive (resp. reflexive and
transitive).
Finally, $R$ is functional if, for all $x\in S$, $|\{y\in S:R(x,y)\}|\leq 1$;
it is nonfunctional otherwise.
\subsection{Background on computational complexity}
The computational complexity of a given problem is a measure of 
the cost of solving it, in terms of the amount of time or space it
needs, as a function of the problem's size. A deterministic
computation is characterised by the unicity, at any time, of the step
to consider next. A nondeterministic computation is one that needs to
``guess'', among a finite number of steps, which to consider
next. There are five main complexity classes, P, PSPACE, EXP, NP, and NEXP,
which characterise, respectively, the problems that are solvable in
deterministic polynomial time, in deterministic polynomial space, in
deterministic exponential time, in nondeterministic polynomial
time, and in nondeterministic exponential time. It is known that
$\mbox{P}\subseteq\mbox{NP}\subseteq\mbox{PSPACE}\subseteq\mbox{EXP}\subseteq\mbox{NEXP}$
and $\mbox{P}\not =\mbox{EXP}$.

Intuitively, a problem $A$ is hard $\wrt$ a complexity class ${\cal
  C}\in\{\mbox{P,NP,PSPACE,EXP,NEXP}\}$, or ${\cal C}$-hard for short, if
  every problem $B$ in ${\cal C}$ can be polynomially reduced to
  $A$, so that an algorithm for $B$ can be ``easily'' obtained from an 
  algorithm for $A$. A problem is complete $\wrt$ a complexity class
  ${\cal C}$, or ${\cal C}$-complete for short, if it is in ${\cal C}$ and is ${\cal C}$-hard.
The reader is referred to \cite{Garey79a,Hopcroft79a} for details.
\subsection{Assumptions on the structure of time}
We make the following assumptions on the structure of time:
\begin{enumerate}
  \item time is discrete;
  \item it has an initial moment with no predecessors; and
  \item it is branching and infinite into the future, and all moments have the same number of
    immediate-successor moments.
\end{enumerate}
Temporal formulas will be interpreted over temporal structures consisting of infinite
$k$-ary $\Sigma$-trees, with $k\geq 1$ and $\Sigma =2^{\cal P}$, ${\cal P}$ being a
countably infinite set of atomic propositions. Such a structure is of the form
$t=<K^*,\pi ,R^*>$:
\begin{enumerate}
  \item $K=\{d_1,\ldots ,d_k\}$ is a set of $k$ directions ---$K^*$ is thus the set
    of finite words over $K$, representing the set of nodes of $t$;
  \item $\pi :\Sigma ^*\rightarrow 2^{\cal P}$ is a truth assignment function, mapping each
    node $x$ of $t$ into the set of atomic propositions true at $x$; and
  \item $R$ is a serial, irreflexive and antisymmetric $k$-ary accessibility
    relation, mapping each node to its $k$ immediate successors ---$R^*$ is
    thus the reflexive-transitive closure of $R$.
\end{enumerate}
A simpler way of representing $t$ is to remove the reflexive-transitive closure symbol,
$*$, from $R^*$; i.e., as $t=<K^*,\pi ,R>$. But there is even simpler: as a
mapping $t:K^*\rightarrow 2^{\cal P}$.

The structure of time so described should be augmented with functions (concrete
features), each of which is associated with an object of the spatial concrete domain
of interest, and records, at each node (time point) of the structure, the position of that
associated object.
\section{A quick overview of the spatial relations to be used as the predicates
of the concrete domain}
We provide a quick overview of the spatial relations to be used, in the family
$\xdl$ of qualitative theories for spatial change, as the predicates of the concrete
domain. But we first remind some general points we discussed in the previous sections:
\begin{enumerate}
  \item $\xdl$ is a spatio-temporalisation of $\alcd$
    \cite{Baader91a}:
    \begin{enumerate}
      \item the abstract objects of $\xdl$ (tree-like) interpretations are time points;
      \item $\xdl$ roles consist of $m+n$ immediate-successor accessibility relations,
        which are antisymmetric and serial, and of which $m$ are general, not necessarily
        functional, the remaining $n$ functional;
      \item the concrete domain ${\cal D}_x$ is generated by a spatial RA $x$ chosen
        as a tool for representing knowledge on $n$-tuples of objects of the spatial
        domain at hand, where $n$ is the arity of the $x$ relations ---stated otherwise,
        the $x$ relations will be used as the predicates of ${\cal D}_x$.
    \end{enumerate}
  \item For clarity of presentation, we focus in this work on $x$ being either of three
    RAs we have already mentioned: either of the two binary RAs $\rcc8$
    \cite{Randell92a,Egenhofer91b} and $\cdalg$ \cite{Frank92b}, or
    the ternary RA $\atra$ of 2-d orientations in \cite{Isli98a,Isli00b}.
  \item The work generalises, in an obvious way, to all spatial RAs $x$ for
    which the atoms are Jointly Exhaustive and Pairwise Disjoint (henceforth JEPD), and
    such that the atomic relations form a decidable subclass: these include
    the binary rectangle algebra in \cite{Balbiani98a,Guesgen89a,Mukerjee90a}, whose atomic relations form a tractable
    subset \cite{Balbiani98a}.
\end{enumerate}
\begin{figure}[t]
\begin{center}
\epsffile{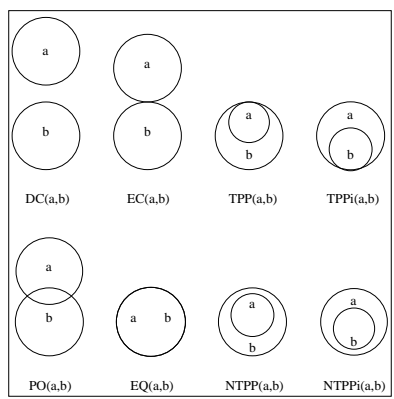}
\caption{An illustration of the RCC-8 atoms.}\label{rcc-cda-atoms}
\end{center}
\end{figure}
\begin{figure}[t]
\begin{center}
\epsffile{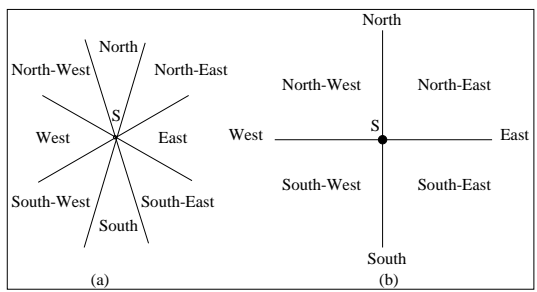}
\caption{Frank's cone-shaped (a) and projection-based (b) models of
cardinal directions.}\label{rcc-cda-atomstwo}
\end{center}
\end{figure}
\subsection{The RA $\rcc8$}
Topology is one of the most developed aspects within the QSR
Community. This is illustrated by the well-known RCC theory
\cite{Randell92a}, from which derives the already mentioned RCC-8 calculus
\cite{Randell92a,Egenhofer91b}. The RCC theory, on the other hand, stems from
Clarke's ``calculus of individuals'' \cite{Clarke81a}, based on a
binary ``connected with'' relation ---sharing of a point of the arguments.
Clarke's work, in turn, was developed from classical mereology
\cite{Leonard40a,Lesniewski28a} and Whitehead's
``extensionally connected with'' relation \cite{Whitehead29a}.
The RCC-8 calculus \cite{Randell92a,Egenhofer91b} consists of a set of eight
JEPD atoms, $\rccdc$ (DisConnected), $\rccec$ (Externally Connected),
$\rcctpp$ (Tangential Proper Part), $\rccpo$ (Partial Overlap), $\rcceq$
(EQual),
$\rccntpp$ (Non Tangential Proper Part), and the converses, $\rcctppi$ and
$\rccntppi$, of $\rcctpp$ and $\rccntpp$, respectively. The reader is
referred to Figure \ref{rcc-cda-atoms} for an illustration of the atoms.
\subsection{The RA $\cdalg$}
Frank's models of cardinal directions in 2D \cite{Frank92b} are
illustrated in Figure \ref{rcc-cda-atomstwo}. They use a partition of the plane
into regions determined by lines
passing through a reference object, say $S$. Depending on the region
a point $P$ belongs to, we have $\north (P,S)$, $\northeast (P,S)$, $\east (P,S)$, 
$\southeast (P,S)$, $\south (P,S)$, $\southwest (P,S)$, $\west (P,S)$, $\northwest (P,S)$, or
$\equal (P,S)$, corresponding, respectively, to the position of $P$
relative to $S$ being $\mbox{\em north}$, $\mbox{\em north-east}$,
$\mbox{\em east}$, $\mbox{\em south-east}$, $\mbox{\em south}$,
$\mbox{\em south-west}$, $\mbox{\em west}$, $\mbox{\em north-west}$,
or $\mbox{\em equal}$. Each of the two models can thus be seen as a
binary RA, with nine atoms.
Both use a global, {\em west-east/south-north}, reference frame.
We focus our attention on the projection-based model (part (b) in Figure \ref{rcc-cda-atomstwo}), which has been
assessed as being cognitively more adequate \cite{Frank92b}.
\subsection{The RA $\atra$}
The set $\deuxdo$ of 2-d orientations is defined in the usual way, and is isomorphic to the set of directed
lines incident with a fixed point, say $O$. Let $h$ be the natural isomorphism, associating with each
orientation $x$ the directed line (incident with $O$) of orientation $x$. The angle $\langle x,y\rangle$
between two orientations $x$ and $y$ is the anticlockwise angle $\langle h(x),h(y)\rangle$.
Isli and Cohn \cite{Isli98a,Isli00b} have defined
a binary RA of 2D orientations, $\apra$, that contains four atoms:
$e$ (equal), $l$ (left), $o$ (opposite) and $r$ (right). For all $x,y\in\deuxdo$:
\begin{eqnarray}
e(y,x) &\Leftrightarrow &\langle x,y\rangle =0  \nonumber  \\
l(y,x) &\Leftrightarrow &\langle x,y\rangle\in (0,\pi )  \nonumber  \\
o(y,x) &\Leftrightarrow &\langle x,y\rangle =\pi  \nonumber  \\
r(y,x) &\Leftrightarrow &\langle x,y\rangle\in (\pi ,2\pi )  \nonumber
\end{eqnarray}
Based on $\apra$, Isli and Cohn \cite{Isli98a,Isli00b} have
built a ternary RA, $\atra$, for cyclic ordering of 2D
orientations: $\atra$ has $24$ atoms, thus $2^{24}$
relations.
The atoms of $\atra$ are written as $b_1b_2b_3$, where
$b_1,b_2,b_3$ are atoms of $\apra$, and such an atom
is interpreted as follows:
$(\forall x,y,z\in\deuxdo )(b_1b_2b_3(x,y,z)\Leftrightarrow b_1(y,x)
 \wedge b_2(z,y)\wedge b_3(z,x))$.
The reader is referred to \cite{Isli98a,Isli00b} for more details.
\section{The $\xdl$ description logics, with $x\in\{\rcc8 ,\cdalg ,\atra\}$}
Description Logics (DLs) constitute a knowledge representation family with a
well-defined semantics, contrary to their ancestors, such as semantic networks
\cite{Quillian68a} or frame systems \cite{Minsky75a}. Their main advantage is
that they are highly expressive while still remaining decidable, or offering
interesting decidable restrictions. One of the most important DLs in the
literature is Schmidt-Schauss and Smolka's $\alc$
\cite{Schmidt-Schauss91a}. $\alc$ includes concepts and roles, which
are, respectively, unary relations and binary relations on a set of (abstract)
objects. One drawback of $\alc$ is that it does not offer a way of referring
to objects of a specific domain of interest, such as a spatial domain, where
the objects could be regions of a topological space, orientations of the
2-dimensional space, or points. To get rid of this insufficiency, $\alc$ has
been extended to what is known as $\alcd$, which augments $\alc$ which a
concrete domain $D$, consisting of a universe of objects, and of predicates
for representing knowledge on these objects \cite{Baader91a}. The roles in
$\alc$ are binary relations in the general meaning of the term, in the sense
that they are not necessarily functional; in $\alcd$, however, they split
into general, not necessarily functional
roles (referred to simply as roles), and functional roles (also referred to as
abstract features).

Temporalisations of DLs are known in the literature (see, e.g.,
\cite{Artale00a,Bettini97a}); as well as spatialisations of DLs  (see, e.g.,
\cite{Haarslev99a}). The present work considers a spatio-temporalisation of
the well-known family $\alcd$ of DLs with a concrete domain \cite{Baader91a}.
Specifically, we consider a temporalisation of the roles of the family,
together with a spatialisation of its concrete domain.
\subsection{Concrete domain}
The role of a concrete domain in so-called DLs with a concrete domain \cite{Baader91a}, is to give
the user of the DL the opportunity to ``touch'' at, and directly refer to, the application domain
at hand. Specifically, the opportunity to represent, thanks to predicates, knowledge on
objects of the application domain, as constraints on tuples of these objects. The set of objects
of the application domain on which a concrete domain ${\cal D}$ represents knowledge, is referred
to as $\Delta _{{\cal D}}$; and the set of predicates used by the concrete domain as a tool for
representing knowledge on tuples of objects from $\Delta _{{\cal D}}$, constitutes the set
$\Phi _{{\cal D}}$ of predicates of ${\cal D}$. As such, a concrete domain can be seen as opening
a DL a window to the application domain. The difference between a DL without, and a DL with, a
concrete domain, is similar to the difference between pure LP, on the one hand,
and, on the other hand, CLP
\cite{BenhamouF93a,Colmerauer90a,Jaffar87a,Jaffar94a,vanHentenryck89a} with a specific domain such
as the rationals or the integers. CLP, as already discussed, additionally incorporates the idea of
specifying a domain for a variable, or for a pair of variables ---i.e., restricting a variable's
range to the elements of a unary relation, or a pair of variables' range to the elements of a
binary relation. Formally, a concrete domain is defined as follows:
\begin{definition}[concrete domain \cite{Baader91a}]\label{cddefinition}
A concrete domain ${\cal D}$ consists of a pair
$(\Delta _{{\cal D}},\Phi _{{\cal D}})$, where
$\Delta _{{\cal D}}$ is a set of (concrete) objects, and
$\Phi _{{\cal D}}$ is a set of predicates over the objects in $\Delta _{{\cal D}}$.
Each predicate $P\in\Phi _{{\cal D}}$ is associated
with an arity $n$ and we have
$P\subseteq (\Delta _{{\cal D}})^n$.
\end{definition}
\begin{definition}[admissibility \cite{Baader91a}]\label{cdadmissibility}
A concrete domain ${\cal D}$ is admissible if:
\begin{enumerate}
  \item the set of its predicates is closed under negation and
    contains a predicate for $\Delta _{{\cal D}}$; and
  \item the satisfiability problem for finite conjunctions of
    predicates is decidable.
\end{enumerate}
\end{definition}
\subsection{The concrete domains ${\cal D}_x$,
            with $x\in\{\rcc8 ,\cdalg ,\atra\}$}
The set $\xat$ of $x$ atoms is the $x$ universal relation, which we also refer
to as the $\xdl$ top predicate $\top ^x$: $\top ^x=\xat$. The set
$\npx$ of $\xdl$ predicate names, which constitutes the set of $\xdl$ atomic
predicates, is the set of $x$ atomic relations:
$\npx =\{\{r\}:r\in\xat\}$.
(Possibly complex) $\xdl$ predicates are obtained by considering
the closure of $\npx$ under the set-theoretic operations of complement with
respect to $\top ^x$, union and intersection. Formally, if the complement
$\top ^x\setminus P$ of $P$ with respect to $\top ^x$ is represented as
$\overline{P}$, then we have the following:
\begin{definition}[$\xdl$ predicates]
The set of $\xdl$ predicates is the smallest set such that:
\begin{enumerate}
  \item every predicate name $P\in\npx$ is an $\xdl$ predicate; and,
  \item if $P_1$ and $P_2$ are $\xdl$ predicates,
    then so are:
    $\overline{P_1}$,
    $P_1\cap P_2$, and
    $P_1\cup P_2$.
\end{enumerate}
\end{definition}
Because the $x$ atoms are JEPD,
the set of $\xdl$ predicates reduces to the set $2^{\xat}$ of all $x$
relations:
each $\xdl$ predicate can be written as $\{P_1,\ldots ,P_n\}$, where
the $P_i$'s are $x$ atoms. The empty relation, $\emptyset$, represents
the bottom predicate, which we also refer to as $\bot _x$.
\begin{remark}
We could use only the $x$ atoms as $\xdl$ predicates, since a constraint
of the form $\{P_1,\ldots ,P_n\}(x_1,\cdots ,x_m)$, where $m$ is the
arity of the $x$ relations, can be equivalently written as the
disjunction
$P_1(x_1,\cdots ,x_m)\vee\ldots\vee P_n(x_1,\cdots ,x_m)$; i.e., as a
disjunction of constraints involving only the predicate names. However,
the first form is preferred to the second, for at least two reasons:
\begin{enumerate}
  \item {\bf flexibility:} in the constraint community, and in particular in
the constraint-based spatial reasoning community, the first form is
preferred to the second -a CSP is nothing else than a conjunction of
such constraints.
  \item {\bf efficiency of a priori pruning with a constraint engine
embedded in the tableaux search space:} the \mbox{spatial-CSP} part of the
label of a node of the search space, as already discussed, is nothing else than a conjunction
of constraints expressed in the RA $x$ which has given birth to the
$\xdl$ concrete domain ---in other words, a CSP expressed in $x$. If
we use the second form, the constraints are instantiated, in a passive
generate-and-test manner, with one of the disjuncts before they are
submitted to the a priori pruning with the constraint engine -they are
submitted as they are, if we use the first form, which gives the
pruning process more potential in preventing failures.
\end{enumerate}
\end{remark}
Let $x\in\{\rcc8 ,\cdalg ,\atra\}$. Thanks to the above discussion,
the concrete domain generated by $x$, ${\cal D}_x$, can be written as
${\cal D}_x=(\Delta _{{\cal D}_x},\Phi _{{\cal D}_x})$, with:
                   \begin{eqnarray}
                         {\cal D}_{\rcc8}   &=&(\rtopspace ,2^{\rccats})  \nonumber  \\
                         {\cal D}_{\cdalg}  &=&(\deuxdp ,2^{\cdalgats})  \nonumber  \\
                         {\cal D}_{\atra}  &=&(\deuxdo ,2^{\atraats})  \nonumber
                   \end{eqnarray}
where:
\begin{enumerate}
  \item $\rtopspace$ is the set of regions of a topological space $\topspace$;
    $\deuxdp$ is the set of 2D points;
    $\deuxdo$ is the set of 2D orientations; and
  \item $\xat$, as we have seen, is the set of $x$ atoms
    ---$2^{\xat}$ is thus the set of all $x$ relations.
\end{enumerate}
\subsection{Admissibility of the concrete domains ${\cal D}_x$,
            with $x\in\{\rcc8 ,\cdalg ,\atra\}$}\label{admissibility}
Let $x$ be an RA from the set $\{\rcc8 ,\cdalg ,\atra\}$.

Closure of the set of predicates, $\Phi _{{\cal D}_x}=2^{\xat}$, of the concrete domain ${\cal D}_x$ is already implicit in
what has been said so far. Given a predicate $P$ of ${\cal D}_x$, corresponding to the $x$
relation $\{r_1,\ldots ,r_n\}$, its negation, $\overline{P}$, is the complement of
$\{r_1,\ldots ,r_n\}$ $\wrt$ the set $\xat$, which represents the $x$ universal relation:
\begin{eqnarray}
\overline{P}=\xat\setminus\{r_1,\ldots ,r_n\}
\end{eqnarray}
The reader should have no difficulty to see that $\overline{P}$ is an element of $\Phi _{{\cal D}_x}$.

A unary relation $R$ can always be written as a particular $n$-ary relation, with
$n\geq 2$. For instance, as the relation $\{(x,\ldots ,x)\in\{x\}^n|x\in R\}$. The
domain $\Delta _{{\cal D}_x}$, which
is a unary relation, can be written as a particular predicate of $\Phi _{{\cal D}_x}$, as follows:
\begin{enumerate}
  \item as the predicate $\rcceq$ if $x=\rcc8$;
  \item as the predicate $\equal$ if $x=\cdalg$; and
  \item as the predicate $\eee$ if $x=\atra$.
\end{enumerate}
In order to establish admissibility of the concrete domains ${\cal D}_x$, it
remains to convince the reader of the decidability of the satisfiability
problem for finite conjunctions of predicates of $\Phi _{{\cal D}_x}$. This
derives from (decidability and) tractability of the subset
$\{\{r\}|r\in\xat\}$ of $x$ atomic relations, for each
$x\in\{\rcc8 ,\cdalg ,\atra\}$:
\begin{enumerate}
  \item The $\rcc8$ atomic relations have been shown to form a tractable subset of $\rcc8$
    by Renz and Nebel \cite{Renz99b}. A problem expressed  in the subset can be checked for
    consistency using Allen's constraint propagation algorithm \cite {Allen83b}.
  \item  The $\cdalg$ atomic relations have been shown to form a tractable subset of $\cdalg$
    by Ligozat \cite{Ligozat98a}. A problem expressed  in the subset can be checked for
    consistency by applying Allen's constraint propagation algorithm \cite{Allen83b} to each
    of the projections on the axes of an orthogonal system of coordinates, chosen in such a
    way that the x- and y-axes are, respectively, a west-east horizontal directed line
    ($\dligne$) and a south-north vertical $\dligne$ ---each of the projections is a problem
    expressed in Vilain and Kautz's temporal point algebra \cite{Vilain86a}.
  \item Isli and Cohn \cite{Isli98a,Isli00b} have provided a propagation algorithm achieving
    4-consistency for CSPs expressed in their RA $\atra$, and shown that the propagation is
    complete for the subset of atomic relations. Indeed, the propagation does even better
    than just being complete: given a CSP of $\atra$ atomic relations, the algorithm either
    detects its inconsistency, if it is inconsistent, or transforms it into a CSP which is
    globally consistent ---the property of global consistency, also called strong
    $n$-consistency in \cite{Freuder82a}, where $n$ is the size of the input CSP, is
    computationally important, for it implies that a solution can searched for in a
    backtack-free manner \cite{Freuder82a}.
\end{enumerate}
The situation is summarised by the following theorem:
\begin{theorem}\label{admissibilitythm}
Let $x\in\{\rcc8 ,\cdalg ,\atra\}$. The concrete domain ${\cal D}_x$ is admissible.
\end{theorem}
\begin{remark}
The concrete domains ${\cal D}_x$, $x\in\{\rcc8 ,\cdalg ,\atra\}$, we consider in this work
behave better than just being admissible. Solving the consistency problem of a conjunction
of constraints expressed in the set of $x$ atomic relations is not only decidable but
tractable as well. Specifically, as already explained, such a conjunction can be solved
with a path consistency algorithm such as Allen's \cite{Allen83b}, in case $x$ is binary,
and with a 4-consistency algorithm such as Isli and Cohn's \cite{Isli98a,Isli00b}, in case
$x$ is ternary. Freksa's point-based calculus of relative orientation
\cite{Freksa92b,Zimmermann96a}, for instance, can generate an admissible concrete domain,
for consistency of a conjunction of constraints expressed in the calculus is
decidable \cite{Scivos01a}; however, the calculus does not verify the tractability property
above (again, the reader is referred to \cite{Scivos01a}). The reason for considering only
``nicely'' admissible concrete domains is that we want to use $\csp$ techniques for the
solving of a conjunction of $x$ constraints; namely:
\begin{enumerate}
  \item a solution search algorithm such as Ladkin and Reinefeld's \cite{Ladkin92a}, which
    uses Allen's path-consistency algorithm \cite{Allen83b} as the filtering procedure
    during the search; and
  \item a solution search algorithm such as Isli and Cohn's \cite{Isli98a,Isli00b}, which
    uses the $4$-consistency algorithm in \cite{Isli98a,Isli00b} as the filtering procedure
    during the search.
\end{enumerate}
\end{remark}
\subsection{Syntax of $\xdl$ concepts,
            with $x\in\{\rcc8 ,\cdalg ,\atra\}$}
Let $x$ be an RA from the set $\{\rcc8 ,\cdalg ,\atra\}$. $\xdl$, as already explained,
is obtained from $\alcd$ by temporalisng the roles, and spatialising the concrete domain.
The roles in $\alc$, as well as the roles other than the abstract features in $\alcd$, are
interpreted in a similar way as the modal operators of the multi-modal logic
${\cal K}_{(m)}$ \cite{Halpern85a} (${\cal K}_{(m)}$ is a multi-modal version of the
minimal normal modal system ${\cal K}$), which explains Schild's \cite{Schild91a} correspondence
between $\alc$ and ${\cal K}_{(m)}$. As in $\alcd$, we will suppose a countably infinite set
$N_R$ of role names (or just roles), and a countably infinite subset $N_{aF}$ of $N_R$ whose
elements consist of abstract feature names (or just abstract features). Additionally, however,
we suppose that the roles (including the abstract features) are antisymmetric and serial ---the
abstract features are also linear.
\begin{definition}[$\xdl$ concepts]\label{defxdlconcepts}
Let $x$ be an RA from the set $\{\rcc8 ,\cdalg ,\atra\}$. Let $N_C$, $N_R$
and $N_{cF}$ be mutually disjoint and countably infinite sets of concept
names, role names, and concrete features, respectively; and $N_{aF}$ a countably infinite subset
of $N_R$ whose elements are abstract features. A (concrete)
feature chain is any finite composition $f_1\ldots f_ng$ of $n\geq 0$ abstract
features $f_1,\ldots ,f_n$ and one concrete feature $g$. The set of $\xdl$
concepts is the smallest set such that:
\begin{enumerate}
  \item\label{defxdlconceptsone} $\top$ and $\bot$ are $\xdl$ concepts
  \item\label{defxdlconceptstwo} an $\xdl$ concept name is an $\xdl$
    (atomic) concept
  \item\label{defxdlconceptsthree} if
    $C$ and $D$ are $\xdl$ concepts;
    $R$ is a role (in general, and an abstract feature in particular);
    $g$ is a concrete feature;
    $u_1$, $u_2$ and $u_3$ are feature chains; and
    $P$ is an $\xdl$ predicate,
    then the following expressions are also $\xdl$ concepts:
    \begin{enumerate}
      \item\label{defxdlconceptsthreea} $\neg C$,
            $C\sqcap D$,
            $C\sqcup D$,
            $\exists R.C$,
            $\forall R.C$; and
      \item\label{defxdlconceptsthreeb}
            $\exists (u_1)(u_2).P$ if $x$ binary,
            $\exists (u_1)(u_2)(u_3).P$ if $x$ ternary.
    \end{enumerate}
\end{enumerate}
\end{definition}
We denote by $\mtalc$ the sublanguage of $\xdl$ given by rules
\ref{defxdlconceptsone}, \ref{defxdlconceptstwo} and
\ref{defxdlconceptsthreea} in Definition \ref{defxdlconcepts},
which is the temporal component of $\xdl$. It is worth noting that $\mtalc$
does not consist of a mere temporalisation of $\alc$
\cite{Schmidt-Schauss91a}. Indeed, $\alc$ contains only general, not
necessarily functional roles,
whereas $\mtalc$ contains abstract features as well. As it will become
clear shortly, a mere temporalisation of $\alc$ (i.e., $\mtalc$
without abstract features) cannot capture the expressiveness of two
well-known modal temporal logics: Propositional Linear Temporal Logic
$\pltl$, and the $\ctl$ version of the full branching modal temporal logic $\ctlstar$
\cite{Emerson90a}.
Given two integers $p\geq 0$ and $q\geq 0$, the sublanguage of
$\xdl$ (resp. $\mtalc$) whose concepts involve at most $p$
general, not necessarily functional roles, and $q$ abstract features
will be referred to as $\xdlpq$ (resp. $\mtalcpq$). We discuss shortly
the cases $(p,q)=(0,0)$, $(p,q)=(0,1)$, and $(0,q)$ with $q\geq 0$. We first
define weakly cyclic TBoxes.
\subsection{Weakly cyclic TBoxes}
An ($\xdl$ terminological) axiom is an expression of the form $A\doteq C$, $A$ being
a (defined) concept name and $C$ a concept. A TBox is a finite set of axioms, with
the condition that no concept name appears more than once as the left hand side of
an axiom.

Let $T$ be a TBox. $T$ contains two kinds of concept names: concept names appearing
as the left hand side of an axiom of $T$ are defined concepts; the others are
primitive concepts. A defined concept $A$ ``{\em directly uses}'' a defined
concept $B$ $\iff$ $B$ appears in the right hand side of the axiom defining $A$. If
``{\em uses}'' is the transitive closure of ``{\em directly uses}'' then $T$
contains a cycle $\iff$ there is a defined concept $A$ that ``{\em uses}'' itself.
$T$ is cyclic if it contains a cycle; it is acyclic otherwise. $T$ is
weakly cyclic if it satisfies the following two conditions:
\begin{enumerate}
  \item Whenever $A$ uses $B$ and $B$ uses $A$, we have $B=A$ ---the only possibility for a defined concept to get
involved in a cycle is to appear in the right hand side of the axiom
defining it.
  \item All possible occurrences of a defined concept $B$ in the right
    hand side of the axiom defining $B$ itself, are within the scope of an
    existential or a universal quantifier; i.e., in subconcepts of 
    $C$ of the form $\exists R.D$ or $\forall R.D$, $C$ being the right 
    hand side of the axiom, $B\doteq C$, defining $B$.
\end{enumerate}
We suppose that the defined concepts of a TBox split into {\em eventuality}
defined concepts and {\em noneventuality} defined concepts.

In the rest of the paper, unless explicitly stated otherwise, we denote
concepts reducing to concept names by the letters $A$ and $B$,
possibly complex concepts by the letters $C$, $D$, $E$,
general (possibly functional) roles by the letter $R$,
abstract features by the letter $f$, concrete features by the letters $g$ and $h$,
feature chains by the letter $u$,
(possibly complex) predicates by the letter $P$.
\subsection{$\xdlzz$: domain-specific Qualitative Spatial $\clp$}
$\xdlzz$ involves no roles and no abstract features. What differentiates it from
the propositional calculus, is the possibility to refer to spatial variables,
thanks to the concrete features, and to ``qualitatively'' restrict, in 
the case $x$ binary, for instance, the domains
of pairs of such variables, thanks to the predicates of the concrete domain. In
other words, $\xdlzz$ can also express constraints of the form
$\exists (g_1)(g_2).P$, where $g_1$ and $g_2$ are concrete features and $P$ is a
predicate of the concrete domain (a qualitative spatial relation of the RA $x$).
$\xdlzz$ can thus bee seen as domain-specific Qualitative Spatial $\clp$ (the
case $x=\rcc8$, for instance, corresponds to the specific domain with the set of
regions of a topological space, as the variables' domain, and with $\rcc8$
relations as constraints for restricting the variation of pairs of these
variables). Item \ref{defxdlconceptsthree} in Definition \ref{defxdlconcepts}
becomes as follows:
\begin{enumerate}
  \item[(3)] if
    $C$ and $D$ are concepts;
    $g_1$, $g_2$ and $g_3$ are concrete features; and
    $P$ is a predicate,
    then the following expressions are also concepts:
    \begin{enumerate}
      \item $\neg C$,
            $C\sqcap D$,
            $C\sqcup D$; and
      \item
            $\exists (g_1)(g_2).P$ if $x$ binary,
            $\exists (g_1)(g_2)(g_3).P$ if $x$ ternary.
    \end{enumerate}
\end{enumerate}
\subsection{$\xdlzo$}
$\xdlzo$ is the sublanguage of $\xdl$, with no nonfunctional roles,
and one abstract feature which we refer to as $f$. $\xdlzo$ with
weakly cyclic TBoxes subsumes the Propositional Linear Temporal
Logic, $\pltl$ (see, for instance, \cite{Emerson90a}). The feature
chains of $\xdlzo$ are of the form $f\ldots fg$ (a finite chain of
the $f$ symbol, followed by a concrete feature). Item
\ref{defxdlconceptsthree} in Definition \ref{defxdlconcepts}
becomes as follows:
\begin{enumerate}
  \item[(3)] if
    $C$ and $D$ are concepts;
    $u_1$, $u_2$ and $u_3$ are feature chains; and
    $P$ is a predicate,
    then the following expressions are also concepts:
    \begin{enumerate}
      \item $\neg C$,
            $C\sqcap D$,
            $C\sqcup D$,
            $\exists f.C$,
            $\forall f.C$; and
      \item
            $\exists (u_1)(u_2).P$ if $x$ binary,
            $\exists (u_1)(u_2)(u_3).P$ if $x$ ternary.
    \end{enumerate}
\end{enumerate}
Well-formed formulas (WFFs) of $\pltl$, over an alphabet ${\cal P}$ of atomic
propositions, are defined as follows, where $\tnext$, $\Box$, $\diamondsuit$
and $U$ are the standard temporal operators {\em next},
{\em necessarily}, {\em eventually} and {\em Until}:
\begin{enumerate}
  \item {\bf true} and {\bf false} are WFFs
  \item an atomic proposition is a WFF
  \item if $\phi$ and $\psi$ are WFFs then so are $\neg\phi$, $\phi\wedge\psi$, $\phi\vee\psi$,
    $\tnext\phi$, $\Box\phi$, $\diamondsuit\phi$ and $\phi U\psi$
\end{enumerate}
A state $s$ of a linear temporal structure $t=<S,\pi ,R^*>$ satisfies a $\pltl$ formula $\phi$, denoted by $t,s\models\phi$, is defined inductively as follows:
\begin{enumerate}
  \item $t,s\models p$ $\iff$ $p\in\pi (s)$, for all atomic propositions $p\in{\cal P}$
  \item $t,s\models\neg\phi$ $\iff$ it is not the case that $t,s\models\phi$
  \item $t,s\models\phi\wedge\psi$ $\iff$ $t,s\models\phi$ and $t,s\models\psi$
  \item $t,s\models\phi\vee\psi$ $\iff$ $t,s\models\phi$ or $t,s\models\psi$
  \item $t,s\models\tnext\phi$ $\iff$ $t,s'\models\phi$, where $s'$ is the immediate successor of $s$ in $t$ ---i.e., $s'$ is such that $f(s,s')$
  \item $t,s\models\Box\phi$ $\iff$ $t,s'\models\phi$, for all $s'$ such that $f^*(s,s')$
  \item $t,s\models\diamondsuit\phi$ $\iff$ $t,s'\models\phi$, for some $s'$ such that $f^*(s,s')$
  \item $t,s\models\phi U\psi$ $\iff$ for some $s'$ such that $f^*(s,s')$:
    \begin{enumerate}
      \item $t,s'\models\psi$; and
      \item $t,s''\models\phi$, for all $s''$ such that $f^*(s,s'')$ and $f^+(s'',s')$
    \end{enumerate}
\end{enumerate}
Formulas of the form $\diamondsuit\phi$ or $\phi U\psi$ are eventuality formulas. A state $s$
of a structure satisfies $\diamondsuit\phi$ (resp. $\phi U\psi$) $\iff$, there exists a
successor state $s'$ of $s$ such that $s'$ satisfies $\phi$ (resp. $s'$ satisfies $\psi$ and
all states between $s$ and $s'$, not necessarily including $s'$, satisfy $\phi$). The Boolean
operators $\wedge$ and $\vee$ are associated with the operators $\sqcap$ and $\sqcup$,
respectively. Each atomic proposition $p$ from ${\cal P}$ is associated with a primitive
concept $A_{p}$. With each $\pltl$ WFF, $\phi$, we associate the $\xdlzo$ defined concept
$B_{\phi}$, defined inductively as follows:
\begin{enumerate}
  \item $B_p\doteq A_p$, for all formulas reducing to an atomic proposition $p$
  \item $B_{true}\doteq\top$
  \item $B_{false}\doteq\bot$
  \item $B_{\neg\phi}\doteq\neg B_{\phi}$
  \item $B_{\phi\wedge\psi}\doteq B_{\phi}\sqcap B_{\psi}$
  \item $B_{\phi\vee\psi}\doteq B_{\phi}\sqcup B_{\psi}$
  \item $B_{\tnext\phi}\doteq\exists f.B_{\phi}$
  \item $B_{\Box\phi}\doteq B_{\phi}\sqcap\exists f.B_{\Box\phi}$
  \item $B_{\diamondsuit\phi}\doteq B_{\phi}\sqcup\exists f.B_{\diamondsuit\phi}$
  \item $B_{\phi U\psi}\doteq B_{\psi}\sqcup(B_{\phi}\sqcap\exists f.B_{\phi U\psi})$
\end{enumerate}
Each of the defined concepts $B_{\Box\phi}$, $B_{\diamondsuit\phi}$ and
$B_{\phi U\psi}$ ``directly uses'' itself. More generally, the procedure is such
that, whenever a defined concept $B_{\phi}$ "directly uses" a defined
concept $B_{\psi}$, $\psi$ is either $\phi$ ($B_{\psi}$ is then
enclosed within the scope of an existential or universal quantifier), or a strict subformula of 
$\phi$. This ensures that the TBox is weakly cyclic.

The axioms defining
$B_{\diamondsuit\phi}$ and $B_{\phi U\psi}$ do not correspond to equivalences. The
intuitive reason behind it is that, they may raise the illusion that,
for instance, a temporal structure satisfies a concept of the form $B_{\diamondsuit\phi}$, even if we report
indefinitely its satisfiability from the current state of the structure to the next,
without satisfying $\phi$. Such defined concepts will be referred to as
{\em eventuality} concepts; these will be used in the determination of the accepting
states of the weak alternating automaton to be associated with the satisfiability of a concept
$\wrt$ a weakly cyclic TBox.
\subsection{$\xdlzq$, with $q\geq 0$}\label{xdlozappendix}
$\xdlzq$, with $q\geq 0$, has no nonfunctional role and $q$ abstract features.
Item \ref{defxdlconceptsthree} in Definition \ref{defxdlconcepts} becomes as follows:
\begin{enumerate}
  \item[(3)] if
    $C$ and $D$ are concepts;
    $f$ is an abstract feature;
    $u_1$, $u_2$ and $u_3$ are feature chains; and
    $P$ is a predicate,
    then the following expressions are also concepts:
    \begin{enumerate}
      \item $\neg C$,
            $C\sqcap D$,
            $C\sqcup D$,
            $\exists f.C$,
            $\forall f.C$; and
      \item
            $\exists (u_1)(u_2).P$ if $x$ binary,
            $\exists (u_1)(u_2)(u_3).P$ if $x$ ternary.
    \end{enumerate}
\end{enumerate}
We now consider the restricted version, $\ctl$, of the full branching modal temporal logic, $\ctlstar$ \cite{Emerson90a}.
{\em State} formulas (true or false of states) and {\em path} formulas
(true or false of paths) of $\ctl$, over an alphabet ${\cal P}$ of
atomic propositions, are defined by rules S1-S2-S3-S4-P0 below,
where the symbols ${\cal A}$ and ${\cal E}$ denote, respectively, the path quantifiers
``for all futures'' (along all paths) and ``for some future'' (along some path):
\begin{enumerate}
  \item[S1]\label{defctlstars1} {\bf true} and {\bf false} are state formulas
  \item[S2]\label{defctlstars2} an atomic proposition is a state formula
  \item[S3]\label{defctlstars3} if $\phi$ and $\psi$ are state formulas then so are $\neg\phi$,
    $\phi\wedge\psi$ and $\phi\vee\psi$
  \item[S4]\label{defctlstars4} if $\phi$ is a path formula then
    ${\cal A}\phi$ and ${\cal E}\phi$ are state formulas
  \item[P0]\label{defctlstarp0} if $\phi$ and $\psi$ are state formulas then $\tnext\phi$,
    $\Box\phi$, $\diamondsuit\phi$ and $\phi U\psi$ are path formulas
\end{enumerate}
The language of $\ctl$, i.e., the set of well-formed formulas (WFFs)
of $\ctl$, is the set of all $\ctl$ state formulas.
Given a branching temporal structure $t=<S,\pi ,R^*>$, a full path of $t$ is an infinite
sequence $s_0,s_1,s_2,\ldots$ such that, for all $i\geq 0$, $R(s_i,s_{i+1})$. As in
\cite{Emerson90a}, we use the convention that $x=(s_0,s_1,s_2,\ldots )$ denotes a full path, and
that $x^i$ denotes the suffix path $(s_i,s_{i+1},\ldots )$. We denote by $t,s\models\phi$
(resp. $t,x\models\phi$) the fact that state formula (resp. path formula) $\phi$ is true in
structure $t$ at state $s_0$ (resp. of path $x$). $t,s\models\phi$ and $t,x\models\phi$ are
defined inductively as follows:
\begin{enumerate}
  \item[S1a] $t,s\models \mbox{\bf true}$
  \item[S1b] $t,s\not\models \mbox{\bf false}$
  \item[S2a] $t,s\models p$ $\iff$ $p\in\pi (s)$, for all atomic propositions $p\in{\cal P}$
  \item[S3a] $t,s\models\neg\phi$ $\iff$ $t,s\not\models\phi$
  \item[S3b] $t,s\models\phi\wedge\psi$ $\iff$ $t,s\models\phi$ and $t,s\models\psi$
  \item[S3c] $t,s\models\phi\vee\psi$ $\iff$ $t,s\models\phi$ or $t,s\models\psi$
  \item[S4a1] $t,s\models {\cal A}\tnext\phi$ $\iff$ for all $s'$ such that $R(s,s')$,
    $t,s'\models\phi$
  \item[S4a2] $t,s\models {\cal A}\Box\phi$ $\iff$ $t,s\models\phi$ and, for all $s'$
    such that $R(s,s')$, $t,s'\models{\cal A}\Box\phi$
  \item[S4a3] $t,s\models {\cal A}\diamondsuit\phi$ $\iff$ $t,s\models\phi$ or, for all
    $s'$ such that $R(s,s')$, $t,s'\models{\cal A}\diamondsuit\phi$
  \item[S4a4] $t,s\models {\cal A}(\phi U\psi )$ $\iff$ $t,s\models\psi$; or
    $t,s\models\phi$ and, for all $s'$ such that $R(s,s')$,
    $t,s\models {\cal A}(\phi U\psi )$
  \item[S4b1] $t,s\models {\cal E}\tnext\phi$ $\iff$ for some $s'$ such that $R(s,s')$,
    $t,s'\models\phi$
  \item[S4b2] $t,s\models {\cal E}\Box\phi$ $\iff$ $t,s\models\phi$ and, for some $s'$
    such that $R(s,s')$, $t,s'\models{\cal E}\Box\phi$
  \item[S4b3] $t,s\models {\cal E}\diamondsuit\phi$ $\iff$ $t,s\models\phi$ or, for some
    $s'$ such that $R(s,s')$, $t,s'\models{\cal E}\diamondsuit\phi$
  \item[S4b4] $t,s\models {\cal E}(\phi U\psi )$ $\iff$ $t,s\models\psi$; or
    $t,s\models\phi$ and, for some $s'$ such that $R(s,s')$,
    $t,s'\models {\cal E}(\phi U\psi )$
\end{enumerate}
It is worth noting that $\ctl$ WFFs prefixed by a quantifier are of the form ${\cal A}\chi$ or
${\cal E}\chi$, where $\chi$ is a path formula of the form $\tnext\phi$,
$\Box\phi$, $\diamondsuit\phi$ or $\phi U\psi$, $\phi$ and $\psi$
being state formulas.

Similarly to the case of $\pltl$, this leads us to the following. The Boolean
operators $\wedge$ and $\vee$ are associated with the operators $\sqcap$ and $\sqcup$,
respectively. Each atomic proposition $p$ from ${\cal P}$ is associated with a primitive
concept $A_{p}$. With each $\ctl$ WFF, $\phi$, we associate the $\mtalczq$ defined concept
$B_{\phi}$, defined recursively by the steps below. Initially, there
is no abstract feature, and no general role. The abstract features are 
created by the procedure as needed. We make use of a general role $R$
which we initialise to the empty role. Whenever a fresh abstract
feature, say $f$, is created, it is added to $R$. So doing, the
general role $R$, by the time the procedure will have completed, will be the union of
all the abstract features in the TBox created for the input
formula. If the created abstract features are $f_1,\ldots ,f_n$, then $R=f_1\cup\ldots\cup f_n$. If $C$ is a concept, then
$\exists R.C$ is synonymous with $\exists
f_1.C\sqcup\ldots\sqcup\exists f_n.C$, and $\forall R.C$ with $\forall
f_1.C\sqcap\ldots\sqcap\forall f_n.C$:
\begin{enumerate}
  \item $B_p\doteq A_p$, for all formulas consisting of an atomic proposition $p$
  \item $B_{true}\doteq\top$
  \item $B_{false}\doteq\bot$
  \item $B_{\neg\phi}\doteq\neg B_{\phi}$
  \item $B_{\phi\wedge\psi}\doteq B_{\phi}\sqcap B_{\psi}$
  \item $B_{\phi\vee\psi}\doteq B_{\phi}\sqcup B_{\psi}$
  \item $B_{{\cal A}\tnext\phi}\doteq\forall R.B_{\phi}$
  \item $B_{{\cal A}\Box\phi}\doteq B_{\phi}\sqcap\forall R.B_{{\cal A}\Box\phi}$
  \item $B_{{\cal A}\diamondsuit\phi}\doteq B_{\phi}\sqcup\forall R.B_{{\cal A}\diamondsuit\phi}$
  \item $B_{{\cal A}(\phi U\psi )}\doteq B_{\psi}\sqcup(B_{\phi}\sqcap\forall R.B_{{\cal A}(\phi U\psi )})$
  \item $B_{{\cal E}\tnext\phi}\doteq\exists f.B_{\phi}$, where $f$ is 
    a fresh abstract feature which we add to $R$ ($R\leftarrow R\cup f$)
  \item $B_{{\cal E}\Box\phi}\doteq B_{\phi}\sqcap\exists f.B_{{\cal E}\Box\phi}$, where $f$ is 
    a fresh abstract feature ($R\leftarrow R\cup f$)
  \item $B_{{\cal E}\diamondsuit\phi}\doteq B_{\phi}\sqcup\exists f.B_{{\cal E}\diamondsuit\phi}$, where $f$ is 
    a fresh abstract feature ($R\leftarrow R\cup f$)
  \item $B_{{\cal E}(\phi U\psi )}\doteq B_{\psi}\sqcup(B_{\phi}\sqcap\exists f.B_{{\cal E}(\phi U\psi )})$, where $f$ is 
    a fresh abstract feature ($R\leftarrow R\cup f$)
\end{enumerate}
The decreasing property explained for the case $\xdlzo$ ensures that,
given an input formula, the procedure outputs a TBox which is weakly
cyclic.

We define the set of subformulas of a formula $\phi$, $\subfset (\phi
)$, inductively in the following obvious way:
\begin{enumerate}
  \item $\subfset (p)=\{p\}$, for all formulas consisting of an atomic proposition $p$
  \item $\subfset (true)=\{true\}$
  \item $\subfset (false)=\{false\}$
  \item $\subfset (\neg\phi )=\{\neg\phi\}\cup\subfset (\phi )$
  \item $\subfset (\phi\wedge\psi )=\{\phi\wedge\psi\}\cup\subfset (\phi )\cup\subfset (\psi )$
  \item $\subfset (\phi\vee\psi )=\{\phi\vee\psi\}\cup\subfset (\phi)\cup\subfset (\psi )$
  \item $\subfset ({\cal A}\tnext\phi )=\{{\cal A}\tnext\phi\}\cup\subfset (\phi )$
  \item $\subfset ({\cal A}\Box\phi )=\{{\cal A}\Box\phi\}\cup\subfset (\phi )$
  \item $\subfset ({\cal A}\diamondsuit\phi )=\{{\cal A}\diamondsuit\phi\}\cup\subfset (\phi )$
  \item $\subfset ({\cal A}(\phi U\psi  ))=\{{\cal A}(\phi U\psi\}\cup\subfset (\psi)\cup\subfset (\phi )$
  \item $\subfset ({\cal E}\tnext\phi )=\{{\cal E}\tnext\phi\}\cup\subfset (\phi )$
  \item $\subfset ({\cal E}\Box\phi )=\{{\cal E}\Box\phi\}\cup\subfset (\phi )$
  \item $\subfset ({\cal E}\diamondsuit\phi )=\{{\cal E}\diamondsuit\phi\}\cup\subfset (\phi )$
\end{enumerate}
Given a formula $\phi$, the defined concept $B_{\phi}$ associated 
with $\phi$ by the procedure described above, is so that all defined
concepts $B_{\phi _1},\ldots ,B_{\phi _n}$ which $B_{\phi}$ ``directly
uses'', and different from $B_{\phi}$ itself, verify the decreasing
property $\mbox{size}(\phi _1)+\ldots +\mbox{size}(\phi
_n)<\mbox{size}(\phi )$, where $\mbox{size}(\psi )$, for a formula
$\psi$, is the size of $\psi$ in terms of number of symbols. This
ensures that the number of defined concepts in the TBox associated
with a formula $\phi$ is linear, and bounded by $\mbox{size}(\phi )$.

It is important to note that, given the fact that formulas of the form
${\cal A}\diamondsuit\phi$, ${\cal A}(\phi U\psi )$,
${\cal E}\diamondsuit\phi$ or ${\cal E}(\phi U\psi )$ are
eventualities, the defined concepts of the form
$B_{{\cal A}\diamondsuit\phi}$, $B_{{\cal A}(\phi U\psi )}$,
$B_{{\cal E}\diamondsuit\phi}$ or $B_{{\cal E}(\phi U\psi )}$,
created by the procedure above, should be marked as eventuality
defined concepts.

Before giving the formal semantics of $\xdl$, we provide some examples.
\section{Examples}
\begin{figure*}[t]
\begin{center}
\epsffile{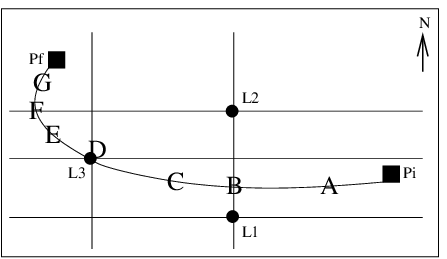}
\caption{Illustration of $\xdlzocda$: the
upward arrow pointing at N indicates North.}\label{sectors}
\end{center}
\end{figure*}
We now provide illustrating examples. Each of Examples \ref{xdlzocda} and
\ref{xdlzoatra} uses an acyclic TBox, which includes feature chains other than concrete
features; whereas each of Examples \ref{xdlztrcc} and \ref{xdlzorcc}
represents a \mbox{non-terminating} physical system, and uses a weakly
cyclic TBox. The TBox of Example \ref{xdlzorcc} uses an eventuality concept.
\begin{example}[illustration of $\xdlzocda$]\label{xdlzocda}
Consider a satellite-like high-level surveillance system, aimed at the 
surveillance of flying aeroplanes within a three-landmark environment.
The basic task of the system is to situate qualitatively an aeroplane
relative to the different landmarks, as well as to relate
qualitatively the different positions of an aeroplane while in flight.
If the system is used for the surveillance of the European sky, the
landmarks could be capitals of European countries, such as Berlin,
London and Paris. For the purpose, the system uses a high-level
spatial description language, such as a QSR language, which we
suppose in this example to be the Cardinal Directions Algebra $\cdalg$
\cite{Frank92b}.
The example is illustrated in Figure \ref{sectors}. The horizontal and
vertical lines through the three
landmarks partition the plane into 0-, 1- and 2-dimensional regions, as shown
in Figure \ref{sectors}. The flight of an aeroplane within the
environment, as tracked by the surveillance system, starts from some point
$P_i$ in Region $A$ (initial region), and ends at some point $P_f$ in Region
$G$ (final, or goal region). Immediately after the initial region, the flight
``moves'' to Region $B$, then to Region $C$, $\ldots$, then to Region $F$,
and finally to the goal region $G$. The tracking of the system consists of
qualitative knowledge on how it ``sees'' the aeroplane at each moment of
the flight being tracked ---within the same region, the knowledge is
constant. The tracking consists thus of recording successive snapshots of the
flight, one per region. A snapshot is a conjunction of constraints giving the
$\cdalg$ relation relating the aeroplane to each of the three landmarks,
situating thus the aeroplane at the corresponding moment. The entire flight
consists of a succession of subflights, $f_A,f_B,\ldots ,f_G$,
such that $f_B$ immediately follows $f_A$, $f_C$ immediately follows $f_B$,
$\ldots$, and $f_G$ immediately follows $f_F$. Subflight $f_X$,
$X\in\{A,\ldots ,G\}$, takes place in Region $X$, and gives rise to a
defined concept $B_X$ describing the panorama of the aeroplane O while in Region $X$, and
saying which subflight takes place next, i.e., which Region is
flied over next. We make use of the concrete features $g_{l1}$, $g_{l2}$,
$g_{l3}$ and $g_{o}$, which have the task of ``referring'', respectively,  to
the actual positions of landmarks $l_1$, $l_2$, $l_3$, and of the aeroplane
$O$. As roles, only one abstract feature is needed, which we refer to as $f$,
which is the linear-time immediate successor function. The acyclic TBox
composed of the following axioms describes the flight:
\begin{footnotesize}
\begin{eqnarray}
B_A&\doteq&
           \exists (g_{o})(g_{l1}).\northeast\sqcap
           \exists (g_{o})(g_{l2}).\southeast\sqcap
           \exists (g_{o})(g_{l3}).\southeast\sqcap
           \exists f.B_B  \nonumber\\
B_B&\doteq&
           \exists (g_{o})(g_{l1}).\north\sqcap
           \exists (g_{o})(g_{l2}).\south\sqcap
           \exists (g_{o})(g_{l3}).\southeast\sqcap
           \exists f.B_C  \nonumber\\
B_C&\doteq&
           \exists (g_{o})(g_{l1}).\northwest\sqcap
           \exists (g_{o})(g_{l2}).\southwest\sqcap
           \exists (g_{o})(g_{l3}).\southeast\sqcap
           \exists f.B_D  \nonumber\\
B_D&\doteq&
           \exists (g_{o})(g_{l1}).\northwest\sqcap
           \exists (g_{o})(g_{l2}).\southwest\sqcap
           \exists (g_{o})(g_{l3}).\equal\sqcap
           \exists f.B_E  \nonumber\\
B_E&\doteq&
           \exists (g_{o})(g_{l1}).\northwest\sqcap
           \exists (g_{o})(g_{l2}).\southwest\sqcap
           \exists (g_{o})(g_{l3}).\northwest\sqcap
           \exists f.B_F  \nonumber\\
B_F&\doteq&
           \exists (g_{o})(g_{l1}).\northwest\sqcap
           \exists (g_{o})(g_{l2}).\west\sqcap
           \exists (g_{o})(g_{l3}).\northwest\sqcap
           \exists f.B_G  \nonumber\\
B_G&\doteq&
           \exists (g_{o})(g_{l1}).\northwest\sqcap
           \exists (g_{o})(g_{l2}).\northwest\sqcap
           \exists (g_{o})(g_{l3}).\northwest  \nonumber
\end{eqnarray}
\end{footnotesize}
The concept $B_A$, for instance, describes the snapshot of the plane while in Region $A$.
It says that the aeroplane is
                           northeast landmark $L_1$ ($\exists (g_{o})(g_{l1}).\northeast$);
                           southeast landmark $L_2$ ($\exists (g_{o})(g_{l2}).\southeast$); and
                           southeast landmark $L_3$ ($\exists (g_{o})(g_{l3}).\southeast$).
The concept also says that the subflight to take place next is $f_B$ ($\exists f.B_B$).

One might want as well the system to track how the aeroplane's different positions during the
flight relate to each other. For example, that the aeroplane, while in region C, remains northwest of its position
while in region B; or, that the position, while in the goal region G, remains northwest of the position
while in region E. These two constraints can be injected into the TBox by modifying
the axioms $B_B$ and $B_E$ as follows:
\begin{footnotesize}
\begin{eqnarray}
B_B&\doteq&
           \exists (g_{o})(g_{l1}).\north\sqcap
           \exists (g_{o})(g_{l2}).\south\sqcap
           \exists (g_{o})(g_{l3}).\southeast\sqcap
           \exists (g_{o})(fg_{o}).\southeast\sqcap
           \exists f.B_C  \nonumber\\
B_E&\doteq&
           \exists (g_{o})(g_{l1}).\northwest\sqcap
           \exists (g_{o})(g_{l2}).\southwest\sqcap
           \exists (g_{o})(g_{l3}).\northwest\sqcap
           \exists (g_{o})(ffg_{o}).\southeast\sqcap
           \exists f.B_F  \nonumber
\end{eqnarray}
\end{footnotesize}
\end{example}
\begin{figure*}[t]
\begin{center}
\epsffile{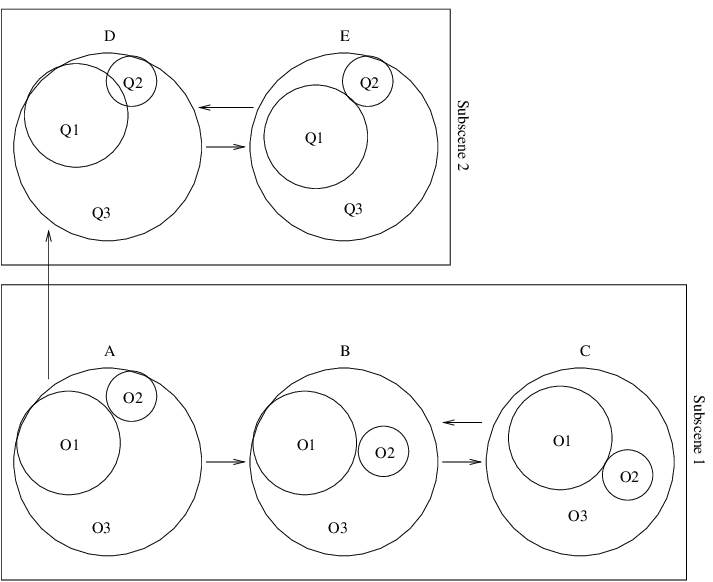}
\caption{Illustration of $\xdlztrcc$.}\label{sectorstwo}
\end{center}
\end{figure*}
\begin{example}[illustration of $\xdlztrcc$]\label{xdlztrcc}Consider the moving spatial scene depicted in Figure
\ref{sectorstwo}, consisting of two subscenes: a subscene 1, composed of three objects o1, o2 and
o3; and a subscene 2, also composed of three objects, q1, q2 and q3:
\begin{enumerate}
  \item For subscene 1, three snapshots of three submotions are presented, and labelled A, B and C;
    the arrows show the transitions from the current submotion to the next. The motion is cyclic. It
    starts with the configurtion A, with o1 touching o2 and tangential proper part of o3, and o2
    tangential proper part of o3. The scene's configuration then ``moves'' to configuration B, involving
    the change of the $\rcc8$ relation on the pair (o2,03) from $\rcctpp$ to its conceptual
    neighbour $\rccntpp$. The next submotion is given by configuration C, involving the object o1 to
    move completely inside o3, becoming thus $\rccntpp$ to it. From C, the motion
    ``moves'' back to the submotion B, and repeats the
    submotions B and C in a \mbox{non-terminating} loop.
  \item For subscene 2, two snapshots of two submotions are presented and labelled D and E; the
    arrows show the transitions from the current submotion to the next. The motion is cyclic. It
    starts with the configurtion D, with q1 partially overlapping q2 and tangential proper part of q3, and q2
    tangential proper part of q3. The scene's configuration then
    ``moves'' to configuration E, involving the change of the $\rcc8$ relation on the pair (q1,q2) from $\rccpo$ to its
    conceptual neighbour $\rccec$, as well as
    the change of the $\rcc8$ relation on the pair (q1,q3) from $\rcctpp$ to its conceptual
    neighbour $\rccntpp$. From E, the motion ``moves'' back to D, and repeats the steps in a
    \mbox{non-terminating} loop.
  \item The scene's motion starts with submotion A. The immediate successors of submotion A are
    submotions B and D, in an incomparable order (the branching from A to B and D is thus an
    and-branching, and not an or-branching). We make use of the concrete features $g_1$, $g_2$ and
    $g_3$ to refer to the actual regions corresponding to objects o1, o2 and o3 in Subscene 1, and
    of the concrete features $h_1$, $h_2$ and $h_3$ to refer to the actual regions corresponding to
    objects q1, q2 and q3 in Subscene 2.
\end{enumerate}
We make use of two abstract features, $f_1$ for the infinite path recording Subscene 1 and starting
at A, and $f_2$ for the infinite path recording Subscene 2 and also starting at A. The weakly
cyclic TBox composed of the following axioms represents the described moving spatial scene:
\begin{footnotesize}
\begin{eqnarray}
B_i&\doteq&B_A\sqcap\exists f_1.B_{BC}\sqcap\exists f_2.B_{DE}
           \nonumber\\
B_{BC}&\doteq&B_B\sqcap
           \exists f_1.(B_C\sqcap\exists f_1.B_{BC})
           \nonumber\\
B_{DE}&\doteq&B_D\sqcap
           \exists f_2.(B_E\sqcap\exists f_2.B_{DE})
           \nonumber\\
B_A&\doteq&\exists (g_{1})(g_{2}).\rccec\sqcap
           \exists (g_{1})(g_{3}).\rcctpp\sqcap
           \exists (g_{2})(g_{3}).\rcctpp
           \nonumber\\
B_B&\doteq&\exists (g_{1})(g_{2}).\rccec\sqcap
           \exists (g_{1})(g_{3}).\rcctpp\sqcap
           \exists (g_{2})(g_{3}).\rccntpp
           \nonumber\\
B_C&\doteq&\exists (g_{1})(g_{2}).\rccec\sqcap
           \exists (g_{1})(g_{3}).\rccntpp\sqcap
           \exists (g_{2})(g_{3}).\rccntpp
           \nonumber\\
B_D&\doteq&\exists (h_{1})(h_{2}).\rccpo\sqcap
           \exists (h_{1})(h_{3}).\rcctpp\sqcap
           \exists (h_{2})(h_{3}).\rcctpp
           \nonumber\\
B_E&\doteq&\exists (h_{1})(h_{2}).\rccec\sqcap
           \exists (h_{1})(h_{3}).\rccntpp\sqcap
           \exists (h_{2})(h_{3}).\rcctpp
           \nonumber
\end{eqnarray}
\end{footnotesize}
The defined concepts $B_A,B_B,B_C$ (resp. $B_D,B_E$) describe the snapshot of Subscene 1 (resp. Subscene
2) during Submotions $A,B,C$ (resp. $D,E$). The concept $B_A$, for instance, says
     that o1 and o2 are related by the $\rccec$ relation ($\exists (g_{1})(g_{2}).\rccec$);
     that o1 and o3 are related by the $\rcctpp$ relation ($\exists (g_{1})(g_{3}).\rcctpp$); and
     that o2 and o3 are also related by the $\rcctpp$ relation ($\exists (g_{2})(g_{3}).\rcctpp$).
The concept $B_{BC}$ describes the cyclic part of Subscene 1, consisting in repeating indefinitely
Submotions $B$ and $C$. Similarly, the concept $B_{DE}$ describes the cyclic part of Subscene 2,
consisting in repeating indefinitely Submotions $D$ and $E$. The
defined concept $B_i$ describes the
initial state of the physical system, which starts with Submotion $A$ and then ``moves'' to the cyclic
submotion $B_{BC}$ along the path $f_1$, and to the other cyclic submotion, $B_{DE}$, along the
path $f_2$.
\end{example}
\begin{figure*}[t]
\begin{center}
\epsffile{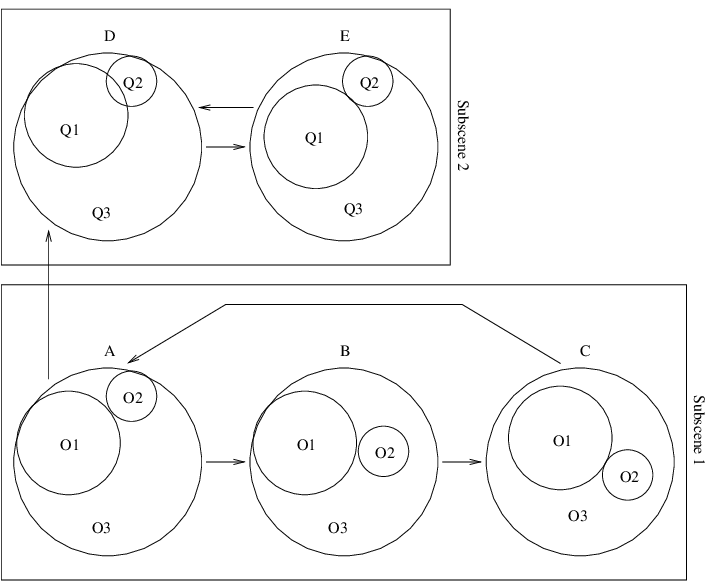}
\caption{Illustration of $\xdlzorcc$.}\label{ill-xdlzorcc}
\end{center}
\end{figure*}
\begin{example}[illustration of $\xdlzorcc$]\label{xdlzorcc}Consider a physical system similar to
the one of the previous example, except that (see Figure \ref{ill-xdlzorcc}):
\begin{enumerate}
  \item in Subscene 1, from Configuration C, the motion ``moves'' back, not to configuration B, but
    to the very first configuration, A; and
  \item the branching from the initial configuration A to the two immediate successors, B and D,
    is not an and-branching, rather an or-branching: from A, the system nondeterministically
    chooses configuration B or configuration D as the next configuration.
\end{enumerate}
We suppose that the configuration of Subscene 2 is reachable, in the sense that the system will
at some point enter configuration D, and then go forever in the repeating of Subscene 2. The
system can thus be seen as ``repeat Subscene 1 until Subscene 2 is reached''.
We make use of one abstract feature, which we denote by $f$. The
defined concepts $B_A$, $B_B$, $B_C$, $B_D$
and $B_E$ remain the same as in the previous example.
The weakly cyclic TBox composed of the
following axioms represents the
described moving spatial scene:
\begin{footnotesize}
\begin{eqnarray}
B_i&\doteq&B_A\sqcap\exists f.(B_B\sqcap\exists f.(B_C\sqcap\exists f.B_i)\sqcup B_{DE})
           \nonumber\\
B_{DE}&\doteq&B_D\sqcap
           \exists f.(B_E\sqcap\exists f.B_{DE})
           \nonumber\\
B_A&\doteq&\exists (g_{1})(g_{2}).\rccec\sqcap
           \exists (g_{1})(g_{3}).\rcctpp\sqcap
           \exists (g_{2})(g_{3}).\rcctpp
           \nonumber\\
B_B&\doteq&\exists (g_{1})(g_{2}).\rccec\sqcap
           \exists (g_{1})(g_{3}).\rcctpp\sqcap
           \exists (g_{2})(g_{3}).\rccntpp
           \nonumber\\
B_C&\doteq&\exists (g_{1})(g_{2}).\rccec\sqcap
           \exists (g_{1})(g_{3}).\rccntpp\sqcap
           \exists (g_{2})(g_{3}).\rccntpp
           \nonumber\\
B_D&\doteq&\exists (h_{1})(h_{2}).\rccpo\sqcap
           \exists (h_{1})(h_{3}).\rcctpp\sqcap
           \exists (h_{2})(h_{3}).\rcctpp
           \nonumber\\
B_E&\doteq&\exists (h_{1})(h_{2}).\rccec\sqcap
           \exists (h_{1})(h_{3}).\rccntpp\sqcap
           \exists (h_{2})(h_{3}).\rcctpp
           \nonumber
\end{eqnarray}
\end{footnotesize}
A (reflexive and transitive) partial order, $\geq$, on the defined concepts in the above TBox
can be defined, which verifies
$B_i\geq B_A$, $B_i\geq B_B$, $B_i\geq B_C$, $B_i\geq B_{DE}$, $B_{DE}\geq B_D$ and
$B_{DE}\geq B_E$ (see Figure \ref{ill-xdlzorcctwo}). The TBox verifies the property that,
given any two defined concepts, $C$ and $D$, if $C$ ``uses'' $D$ then $C\geq D$. The TBox is
thus weakly cyclic.
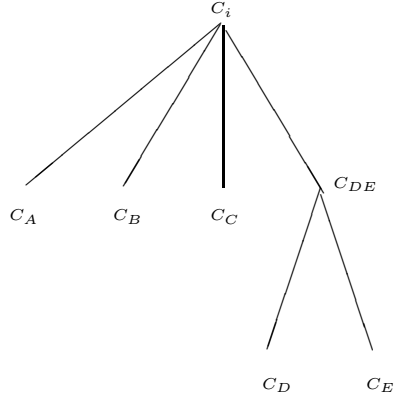
\begin{figure*}[t]
\begin{center}
\setlength{\unitlength}{3572sp}%
\begingroup\makeatletter\ifx\SetFigFont\undefined%
\gdef\SetFigFont#1#2#3#4#5{%
  \reset@font\fontsize{#1}{#2pt}%
  \fontfamily{#3}\fontseries{#4}\fontshape{#5}%
  \selectfont}%
\fi\endgroup%
\begin{picture}(2532,2745)(4366,-2581)
\thinlines
\special{ps: gsave 0 0 0 setrgbcolor}\put(5851,-61){\line( 0,-1){1125}}
\special{ps: grestore}\special{ps: gsave 0 0 0 setrgbcolor}\put(5833,-39){\line(-6,-5){1350}}
\special{ps: grestore}\special{ps: gsave 0 0 0 setrgbcolor}\put(5831,-49){\line(-3,-5){675}}
\special{ps: grestore}\special{ps: gsave 0 0 0 setrgbcolor}\put(5871,-94){\line( 3,-5){675}}
\special{ps: grestore}\special{ps: gsave 0 0 0 setrgbcolor}\put(6526,-1186){\line(-1,-3){373.500}}
\special{ps: grestore}\special{ps: gsave 0 0 0 setrgbcolor}\put(6526,-1231){\line( 1,-3){360}}
\special{ps: grestore}\put(5761, 29){\makebox(0,0)[lb]{\smash{\SetFigFont{6}{7.2}{\rmdefault}{\mddefault}{\updefault}\special{ps: gsave 0 0 0 setrgbcolor}$C_i$\special{ps: grestore}}}}
\put(6616,-1186){\makebox(0,0)[lb]{\smash{\SetFigFont{6}{7.2}{\rmdefault}{\mddefault}{\updefault}\special{ps: gsave 0 0 0 setrgbcolor}$C_{DE}$\special{ps: grestore}}}}
\put(4366,-1411){\makebox(0,0)[lb]{\smash{\SetFigFont{6}{7.2}{\rmdefault}{\mddefault}{\updefault}\special{ps: gsave 0 0 0 setrgbcolor}$C_A$\special{ps: grestore}}}}
\put(5086,-1411){\makebox(0,0)[lb]{\smash{\SetFigFont{6}{7.2}{\rmdefault}{\mddefault}{\updefault}\special{ps: gsave 0 0 0 setrgbcolor}$C_B$\special{ps: grestore}}}}
\put(5761,-1411){\makebox(0,0)[lb]{\smash{\SetFigFont{6}{7.2}{\rmdefault}{\mddefault}{\updefault}\special{ps: gsave 0 0 0 setrgbcolor}$C_C$\special{ps: grestore}}}}
\put(6121,-2581){\makebox(0,0)[lb]{\smash{\SetFigFont{6}{7.2}{\rmdefault}{\mddefault}{\updefault}\special{ps: gsave 0 0 0 setrgbcolor}$C_D$\special{ps: grestore}}}}
\put(6841,-2581){\makebox(0,0)[lb]{\smash{\SetFigFont{6}{7.2}{\rmdefault}{\mddefault}{\updefault}\special{ps: gsave 0 0 0 setrgbcolor}$C_E$\special{ps: grestore}}}}
\end{picture}
\caption{The partial order on the defined concepts of
  the TBox in Example \ref{xdlzorcc}.}\label{ill-xdlzorcctwo}
\end{center}
\end{figure*}

The concept $B_i$ describes the
initial state of the physical system, which either performs the submotion of Subscene 1 before
repeating itself, or skips to Subscene 2 which it repeats indefinitely. Again, because we want
Subscene 2 to be reachable, the concept $B_i$ describes an eventuality, and should be marked as
an eventuality concept -this allows rejecting those potential models which repeat indefinitely
Subscene 1 without reaching Subscene 2.
\end{example}
\begin{example}[illustration of $\xdlzoatra$]\label{xdlzoatra}
We consider an environment with four landmarks, L1, L2, L3 and L4, as depicted in Figure
\ref{tessellation}(Left). The lines through the different pairs of landmarks partition the plane into a
tessellation of two-, one- and zero-dimensional convex regions. Nine of these regions are numbered
R1, ..., R9 in Figure \ref{tessellation}(Left). A robot R has to navigate all the way through from some
point Pi in Region R1 to some point Pf in Rgion R9, traversing in between Regions R2, ..., R8, in that
order. With each region R$i$, $i=1\ldots 9$, we associate a concept $B_i$ describing the panorama of
the robot while in Region R$i$, and giving the
region the robot will be in next. We make use of four concrete features $g_1,\ldots ,g_4$, which
``perceive'' at each time instant the orientations $o_1$, $o_2$, $o_3$ and $o_4$ of the directed lines
joining the robot to Landmarks L1, L2, L3 and L4, respectively ---Figure \ref{tessellation}(Right);
and of one abstract feature $f$ representing the linear time immediate-successor function. The
panorama of the robot at a specific time point consists in the conjunction of $\atra$ constraints
associating with each triple of the four orientations the $\atra$ relation it satisfies. Within the
same region, the panorama is constant. The navigation of the robot can thus be seen as a chronological
evolution of the changing panorama. The TBox with the following axioms provides a plan describing a
path the robot has to follow to reach the goal.
\begin{figure*}[t]
\epsffile{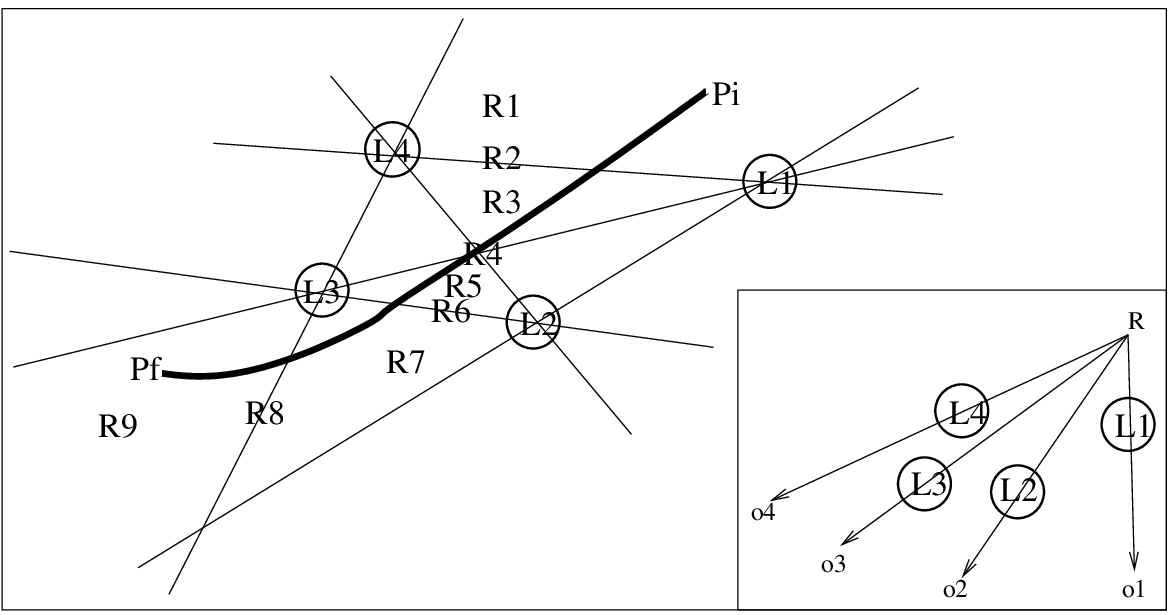}
\caption{Illustration of $\xdlzoatra$.}\label{tessellation}
\end{figure*}
\begin{footnotesize}
\begin{eqnarray}
B_1&\doteq&
           \exists (g_1)(g_2)(g_3).\rrr\sqcap
           \exists (g_1)(g_2)(g_4).\rrr\sqcap
           \exists (g_1)(g_3)(g_4).\rrr\sqcap
           \exists (g_2)(g_3)(g_4).\rrr\sqcap
           \exists f.B_2  \nonumber\\
B_2&\doteq&
           \exists (g_1)(g_2)(g_3).\rrr\sqcap
           \exists (g_1)(g_2)(g_4).\rro\sqcap
           \exists (g_1)(g_3)(g_4).\rro\sqcap
           \exists (g_2)(g_3)(g_4).\rrr\sqcap
           \exists f.B_3  \nonumber\\
B_3&\doteq&
           \exists (g_1)(g_2)(g_3).\rrr\sqcap
           \exists (g_1)(g_2)(g_4).\rrl\sqcap
           \exists (g_1)(g_3)(g_4).\rrl\sqcap
           \exists (g_2)(g_3)(g_4).\rrr\sqcap
           \exists f.B_4  \nonumber\\
B_4&\doteq&
           \exists (g_1)(g_2)(g_3).\rro\sqcap
           \exists (g_1)(g_2)(g_4).\rol\sqcap
           \exists (g_1)(g_3)(g_4).\orl\sqcap
           \exists (g_2)(g_3)(g_4).\rro\sqcap
           \exists f.B_5  \nonumber\\
B_5&\doteq&
           \exists (g_1)(g_2)(g_3).\rrl\sqcap
           \exists (g_1)(g_2)(g_4).\rll\sqcap
           \exists (g_1)(g_3)(g_4).\lrl\sqcap
           \exists (g_2)(g_3)(g_4).\rrl\sqcap
           \exists f.B_6  \nonumber\\
B_6&\doteq&
           \exists (g_1)(g_2)(g_3).\rol\sqcap
           \exists (g_1)(g_2)(g_4).\rll\sqcap
           \exists (g_1)(g_3)(g_4).\lrl\sqcap
           \exists (g_2)(g_3)(g_4).\orl\sqcap
           \exists f.B_7  \nonumber\\
B_7&\doteq&
           \exists (g_1)(g_2)(g_3).\rll\sqcap
           \exists (g_1)(g_2)(g_4).\rll\sqcap
           \exists (g_1)(g_3)(g_4).\lrl\sqcap
           \exists (g_2)(g_3)(g_4).\lrl\sqcap
           \exists f.B_8  \nonumber\\
B_8&\doteq&
           \exists (g_1)(g_2)(g_3).\rll\sqcap
           \exists (g_1)(g_2)(g_4).\rll\sqcap
           \exists (g_1)(g_3)(g_4).\lel\sqcap
           \exists (g_2)(g_3)(g_4).\lel\sqcap
           \exists f.B_9  \nonumber\\
B_9&\doteq&
           \exists (g_1)(g_2)(g_3).\rll\sqcap
           \exists (g_1)(g_2)(g_4).\rll\sqcap
           \exists (g_1)(g_3)(g_4).\lll\sqcap
           \exists (g_2)(g_3)(g_4).\lll  \nonumber
\end{eqnarray}
\end{footnotesize}
The defined concept $B_4$, for instance, provides the information that the
orientations $o_1,\ldots ,o_4$ should satisfy the constraints
that
the $\atra$ relation on the triple $(o_1,o_2,o_3)$ is $\rro$,
the one on the triple $(o_1,o_2,o_4)$ is $\rol$,
the one on the triple $(o_1,o_3,o_4)$ is $\orl$, and
the one on the triple $(o_2,o_3,o_4)$ is $\rro$
---which is a description of the panorama of the robot while
in Region $R_4$. Concept $B_4$ also tells which submotion
should take place next ($\exists f.B_5$).

As with Example \ref{xdlzocda}, we can use feature chains of length greater than one (i.e., not
reducing to concrete features) to relate, for instance, the value of the line joining the robot to
Landmark $L_3$ while the robot is in Region $R_1$, to the value of the same line while the robot
will be in Region $R_9$. We might want to constrain the motion of the robot, so that it does not expand
beyond the part of Region $R_9$ which, from Region $R_1$, appears to the robot's visual system to be
to the left hand side of Landmark $L_3$. The reason for forcing such a constraint could be that,
from Region $R_1$, the part of Region $R_9$ within the right hand side of Landmark $R_3$ is hidden
to the robot's vision system, which makes its reaching a potential danger. This can be done by
forcing the value of orientation $0_3$
while the robot is in Region $R_1$, to be to the left of the value of the same orientation while the
robot is in Region $R_9$. This constraint can be injected into the TBox by modifying the concept
$B_1$ as follows, where $f^8g_3$ stands for the feature chain $ffffffffg_3$:
\begin{footnotesize}
\begin{eqnarray}
B_1&\doteq&
           \exists (g_1)(g_2)(g_3).\rrr\sqcap
           \exists (g_1)(g_2)(g_4).\rrr\sqcap
           \exists (g_1)(g_3)(g_4).\rrr\sqcap
           \exists (g_2)(g_3)(g_4).\rrr\sqcap\nonumber\\
   &      &\exists (g_3)(g_3)(f^8g_3).\err\sqcap
           \exists f.B_2  \nonumber
\end{eqnarray}
\end{footnotesize}
\end{example}
\begin{example}[another illustration of $\rccdl$]\label{ciarccdltwo}In \cite{Bennett98a}, the authors
describe a
system answering queries on the RCC-8 \cite{Randell92a} relation between two input (polygonal) regions
of a (quantitative) geographic database. The system also includes the computation of the qualitative
abstraction of a quantitative geographic database, which is done by transforming the quantitative
database into a qualitative one, which records the $\rcc8$ relation on each pair of the regions in the
quantitative database. The importance of the system is obvious: most applications querying geographic
databases only need the query-answering system to provide them with the topological relation on pairs of
regions in the database. If we think of the database as representing the World's geographic map, then
the queries could be of the form ``Is Hamburg a German city?'', ``What are the Mediterranean countries of
Africa?'', ``Are France and Germany neighbouring countries?'', or ``Does the Sahara Desert just partially
overlap, or is it part of, Algeria?''. Because computing such a relation directly from a quantitative
database is time-consuming, it is worth, especially in situations of repetition of such queries, to compute
once and for all the topological relation between every pair of regions in the database, and to store them
in a qualitative
database; the next time a similar topological query reaches the system, the latter would then only need
to access (in constant time) the qualitative database, and to retrieve the relation from there. Because
of phenomena such as erosion and (unfortunately) wars, the boundaries of the regions in a geographic
database may change with time. It should be clear that $\rccdl$ can be used to represent the history of
the qualitative abstraction of such a geographic database.
\end{example}
\section{Semantics of $\xdl$,
            with $x\in\{\rcc8 ,\cdalg ,\atra\}$}
\begin{definition}[$k$-ary $\Sigma$-tree]\label{karymtree}
Let $\Sigma$ and $K=\{d_1,\ldots ,d_k\}$, $k\geq 1$, be two disjoint alphabets: $\Sigma$ is a labelling alphabet and $K$ an
alphabet of directions. A (full) $k$-ary tree is an infinite tree
whose nodes $\alpha\in K^*$ have exactly $k$ immediate successors each,
$\alpha d_1,\ldots ,\alpha d_k$. A $\Sigma$-tree is a tree whose nodes are
labelled with elements of $\Sigma$. A (full) $k$-ary $\Sigma$-tree is
a $k$-ary tree $t$ which is also a $\Sigma$-tree, which we consider as
a mapping $t:K^*\rightarrow\Sigma$ associating with each node
$\alpha\in K^*$ an element $t(\alpha )\in\Sigma$. The empty word, $\epsilon$, denotes
the root of $t$. Given a node $\alpha\in K^*$ and a direction $d\in K$,
the concatenation of $\alpha$ and $d$, $\alpha d$, denotes the $d$-successor of
$\alpha$. The level $|\alpha |$ of a node $\alpha$ is the length of $\alpha$ as a word. We can
thus think of the edges of $t$ as being labelled with directions from $K$, and of the nodes of
$t$ as being labelled with letters from $\Sigma$. A partial $k$-ary
$\Sigma$-tree (over the set $K$ of directions) is a $\Sigma$-tree with
the property that a node may not have a $d$-successor for each
direction $d$; in other terms, a partial $k$-ary $\Sigma$-tree is a
$\Sigma$-tree which is a prefix-closed\footnote{$t$ is prefix-closed
  if, for all nodes $\alpha$, if $t$ is defined for $\alpha$ then it
  defined for all nodes $\alpha '$ consisting of prefixes of
  $\alpha$.} partial function $t:K^*\rightarrow\Sigma$.
\end{definition}
$\xdl$ is equipped with a Tarski-style, possible worlds
semantics. $\xdl$ interpretations are spatio-temporal structures
consisting of $k$-ary trees $t$, representing
\mbox{$k$-immediate-successor} branching time, together with an interpretation
function associating with each primitive concept $A$ the nodes of $t$
at which $A$ is true, and, additionally, associating with each
concrete feature $g$ and each node $u$ of $t$, the value at $u$ (seen as
a time instant) of the spatial concrete object referred to by $g$. Formally:
\begin{definition}[interpretation]
Let $x\in\{\rcc8 ,\cdalg ,\atra\}$ and $K=\{d_1,\ldots ,d_k\}$ a set
of $k$ directions.
An interpretation ${\cal I}$ of $\xdl$ consists of a pair ${\cal I}=(t _{{\cal I}},.^{{\cal I}})$, where
$t _{{\cal I}}$ is a $k$-ary tree and $.^{{\cal I}}$ is an
interpretation function mapping each primitive concept $A$ to a subset
$A^{{\cal I}}$ of $K^*$, each role $R$ to a subset $R^{{\cal I}}$ of
$\{(u,ud)\in K^*\times K^*:\mbox{ }d\in K\}$, so that $R^{{\cal I}}$
is functional if $R$ is an an abstract feature, and each
concrete feature $g$ to a total function $g^{{\cal I}}$:
\begin{footnotesize}
\begin{enumerate}
  \item from $K^*$ onto the set $\rtopspace$ of regions of a topological space $\topspace$, if $x=\rcc8$;
  \item from $K^*$ onto the set $\deuxdp$ of points of the 2-dimensional space, if $x=\cdalg$; and
  \item from $K^*$ onto the set $\deuxdo$ of orientations of the 2-dimensional
    space, if $x=\atra$.
\end{enumerate}
\end{footnotesize}
\end{definition}
Given an $\xdl$ interpretation ${\cal I}=(t_{{\cal I}},.^{{\cal I}})$, a feature chain
$u=f_1\ldots f_ng$, and a node $v_1$, we denote by $u^{{\cal I}}(v_1)$
the value $g^{{\cal I}}(v_2)$, where $v_2$ is the $f_1^{{\cal
    I}}\ldots f_n^{{\cal I}}$-successor of $v_1$; i.e., $v_2$ is so that there exists a sequence
$v_1=w_0,w_1,\ldots ,w_n=v_2$ verifying $(w_i,w_{i+1})\in
f_{i+1}^{{\cal I}}$,
for all $i\in\{1,\ldots ,n-1\}$.
\begin{definition}[satisfiability $\wrt$ a TBox]
Let $x\in\{\rcc8 ,\cdalg ,\atra\}$ be a spatial RA, $K=\{d_1,\ldots ,d_k\}$ a set
of $k$ directions, $C$ an $\xdl$ concept, ${\cal T}$ an $\xdl$
weakly cyclic TBox, and ${\cal I}=(t_{{\cal I}},.^{{\cal I}})$ an
$\xdl$ interpretation. The satisfiability, by a node $s$ of $t_{{\cal
    I}}$, of $C$ $\wrt$ to ${\cal T}$, denoted ${\cal I},s\models\langle C,{\cal T}\rangle$, is defined inductively as follows:
\begin{enumerate}
  \item ${\cal I},s\models\langle\top ,{\cal T}\rangle$
  \item ${\cal I},s\not\models\langle\bot ,{\cal T}\rangle$
  \item ${\cal I},s\models\langle A,{\cal T}\rangle$ $\iff$
    $s\in A^{{\cal I}}$, for all primitive concepts $A$
  \item ${\cal I},s\models\langle B,{\cal T}\rangle$ $\iff$
    ${\cal I},s\models\langle C,{\cal T}\rangle$, for all defined
    concepts $B$ defined by the axiom $B\doteq C$ of ${\cal T}$
  \item ${\cal I},s\models\langle\neg C,{\cal T}\rangle$ $\iff$
    ${\cal I},s\not\models\langle C,{\cal T}\rangle$
  \item ${\cal I},s\models\langle C\sqcap D,{\cal T}\rangle$ $\iff$
    ${\cal I},s\models\langle C,{\cal T}\rangle$ and ${\cal
      I},s\models\langle D,{\cal T}\rangle$
  \item ${\cal I},s\models\langle C\sqcup D,{\cal T}\rangle$ $\iff$
    ${\cal I},s\models\langle C,{\cal T}\rangle$ or ${\cal
      I},s\models\langle D,{\cal T}\rangle$
  \item ${\cal I},s\models\langle\exists R.C,{\cal T}\rangle$ $\iff$
    ${\cal I},s'\models\langle C,{\cal T}\rangle$, for some $s'$ such
    that $(s,s')\in R^{{\cal I}}$
  \item ${\cal I},s\models\langle\forall R.C,{\cal T}\rangle$ $\iff$
    ${\cal I},s'\models\langle C,{\cal T}\rangle$, for all $s'$ such
    that $(s,s')\in R^{{\cal I}}$
  \item ${\cal I},s\models\langle\exists (u_1)(u_2).P,{\cal T}\rangle$ $\iff$
    $P(u_1^{{\cal I}}(s),u_2^{{\cal I}}(s))$
  \item ${\cal I},s\models\langle\exists (u_1)(u_2)(u_3).P,{\cal T}\rangle$ $\iff$
    $P(u_1^{{\cal I}}(s),u_2^{{\cal I}}(s),u_3^{{\cal I}}(s))$
\end{enumerate}
A concept $C$ is satisfiable $\wrt$ a TBox ${\cal T}$ 
$\iff$ ${\cal I},s\models\langle C,{\cal T}\rangle$, for some $\xdl$
interpretation ${\cal I}$, and some state
$s\in t_{{\cal I}}$, in which case the pair $({\cal I},s)$ is a model
of $C$ $\wrt$ ${\cal T}$; $C$ is insatisfiable (has no models) $\wrt$ ${\cal T}$,
otherwise. $C$ is valid $\wrt$ ${\cal T}$ $\iff$ the negation, $\neg
C$, of $C$ is insatisfiable $\wrt$ ${\cal T}$. The satisfiability
problem and the subsumption problem are defined as follows:
\begin{enumerate}
  \item[] {\em The satisfiability problem:}
    \begin{enumerate}
      \item[$\bullet$] {\bf Input:} a concept $C$ and a TBox ${\cal T}$
      \item[$\bullet$] {\bf Problem:} is $C$ satisfiable $\wrt$ ${\cal 
          T}$?
    \end{enumerate}
  \item[] {\em The subsumption problem:}
    \begin{enumerate}
      \item[$\bullet$] {\bf Input:} two concepts $C$ and $D$ and a TBox ${\cal T}$
      \item[$\bullet$] {\bf Problem:} does $C$ subsume $D$ $\wrt$
        ${\cal T}$ (notation: $D\sqsubseteq _{{\cal T}}C$)? in other words, are all models of $C$ $\wrt$
        ${\cal T}$ also models of $D$ $\wrt$ ${\cal T}$?
    \end{enumerate}
\end{enumerate}
\end{definition}
The satisfiability problem and the subsumption problem are related to
each other, as follows: $D\sqsubseteq _{{\cal T}}C$ $\iff$
$D\sqcap\neg C$ is insatisfiable $\wrt$ ${\cal T}$.
\section{The satisfiability of an $\xdl$ concept $\wrt$ a weakly cyclic TBox}
Let $C$ be an $\xdl$ concept and ${\cal T}$ an $\xdl$ weakly cyclic TBox. We define ${\cal T}\oplus C$ as the TBox
${\cal T}$ augmented with the axiom $B_i\doteq C$, where $B_i$ is a fresh defined concept (not
occurring in ${\cal T}$):
\begin{footnotesize}
\begin{eqnarray}
{\cal T}\oplus C=\langle {\cal T}\cup\{B_i\doteq C\},B_i\rangle\nonumber
\end{eqnarray}
\end{footnotesize}
In the sequel, we refer to ${\cal T}\oplus C$ as the TBox ${\cal T}$
augmented with $C$, and to $B_i$ as the initial state of ${\cal T}\oplus C$.
The idea now is that, satisfiability of $C$ $\wrt$ ${\cal T}$ has (almost) been reduced to the emptiness
problem of ${\cal T}\oplus C$, seen as a weak alternating automaton on
$k$-ary $\Sigma$-trees, for some labelling alphabet $\Sigma$ to be
defined later, with the
defined concepts as the states of
the automaton, $B_i$ as the initial state of the automaton, the axioms
as defining the transition function,
with the accepting condition derived from those defined concepts which are not eventuality concepts, and with $k$
standing for the number of concepts of the form $\exists R.D$ in a certain closure, to be defined later, of
${\cal T}\oplus C$.
\subsection{The Disjunctive Normal Form}
The notion of Disjunctive Normal Form (DNF) of a concept $C$ $\wrt$ to a TBox ${\cal T}$, $\dnfone (C,{\cal T})$,
is crucial for the rest of the paper.
Such a form results, among other things, from the use of De Morgan's Law to decompose a concept so that, in the
final form, the negation symbol outside the scope of a (existential or universal) quantifier occurs only in front
of primitive concepts.
\begin{definition}[first DNF]\label{firstdnf}
The first Disjunctive Normal Form ($\dnfone$) of an $\xdl$ concept $C$ $\wrt$ an $\xdl$ TBox ${\cal T}$,
$\dnfone (C,{\cal T})$, is defined recursively as follows:
\begin{enumerate}
  \item for all primitive concepts $A$: $\dnfone (A,{\cal
      T})=\{\{A\}\}$, $\dnfone (\neg A,{\cal T})=\{\{\neg A\}\}$
  \item $\dnfone (\top ,{\cal T})=\{\emptyset\}$, $\dnfone (\bot ,{\cal T})=\emptyset$
  \item for all defined concepts $B$: $\dnfone (B,{\cal T})=\dnfone (E,{\cal T})$,
    $\dnfone (\neg B,{\cal T})=\dnfone (\neg E,{\cal T})$, where $E$ is the right hand side of the axiom $B\doteq E$
    defining $B$
  \item $\dnfone (C\sqcap D,{\cal T})=\prod (\dnfone (C,{\cal T}),\dnfone (D,{\cal T}))$
  \item $\dnfone (C\sqcup D,{\cal T})=\dnfone (C,{\cal T})\cup\dnfone (D,{\cal T})$
  \item $\dnfone (\exists R.C,{\cal T})=\{\{\exists R.C\}\}$
  \item $\dnfone (\forall R.C,{\cal T})=\{\{\forall R.C\}\}$
  \item $\dnfone (\exists (u_1)(u_2).P,{\cal T})=\{\{\exists (u_1)(u_2).P\}\}$
  \item $\dnfone (\exists (u_1)(u_2)(u_3).P,{\cal T})=\{\{\exists (u_1)(u_2)(u_3).P\}\}$
  \item $\dnfone (\neg (C\sqcap D),{\cal T})=\dnfone (\neg C,{\cal T})\cup\dnfone (\neg D,{\cal T})$
  \item $\dnfone (\neg (C\sqcup D),{\cal T})=\prod (\dnfone (\neg C,{\cal T}),\dnfone (\neg D,{\cal T}))$
  \item $\dnfone (\neg\exists R.C,{\cal T})=\{\{\forall R.\neg C\}\}$
  \item $\dnfone (\neg\forall R.C,{\cal T})=\{\{\exists R.\neg C\}\}$
\end{enumerate}
where $\prod$ is defined as follows:
\begin{enumerate}
  \item 
$
\prod (\{S\},\{T\})=
  \left\{
                   \begin{array}{ll}
                         \emptyset
                               &\mbox{ if }\{A,\neg A\}\subseteq S\cup T\mbox{ for some primitive concept }A,  \\
                         \{S\cup T\}
                               &\mbox{ otherwise}  \\
                   \end{array}
  \right.
$
  \item $\prod (\{S_1,\ldots ,S_n\},\{T_1,\ldots ,T_m\})=\displaystyle
    \bigcup _{i\in\{1,\ldots ,n\},j\in\{1,\ldots ,m\}}\prod (\{S_i\},\{T_j\})$
\end{enumerate}
\end{definition}
Note that the $\dnfone$ function checks satisfiability at the propositional level,
in the sense that, given a concept $C$, $\dnfone (C,{\cal T})$ is either empty, or is such
that for all $S\in\dnfone (C,{\cal T})$, $S$ does not contain both $A$ and $\neg A$, $A$
being a primitive concept. Furthermore, given a set $S\in\dnfone (C,{\cal T})$, all
elements of $S$ are concepts of either of the following forms:
\begin{enumerate}
  \item $A$ or $\neg A$, where $A$ is a primitive concept;
  \item $\exists R.D$;
  \item $\forall R.D$; or
  \item $\exists (u_1)(u_2).P$ if $x$ binary,
    $\exists (u_1)(u_2)(u_3).P$ if $x$ ternary.
\end{enumerate}
\begin{definition}
Let $C$ be an $\xdl$ concept, ${\cal T}$ and $\xdl$ TBox and $S\in\dnfone (C,{\cal T})$.
The set of concrete features of $S$, $\concretefeatures (S)$, is
defined as the set of concrete features, $g$, for which there exists a 
feature chain $u$ suffixed by $g$, such that $S$ contains a predicate
concept $\exists (u_1)(u_2).P$, with $u\in\{u_1,u_2\}$, if $x$ binary;
or $S$ contains a predicate concept $\exists (u_1)(u_2)(u_3).P$, with
$u\in\{u_1,u_2,u_3\}$, if $x$ ternary.
\end{definition}
\begin{definition}[the $\pceaone$ partition]
Let $C$ be an $\xdl$ concept, ${\cal T}$ an $\xdl$ TBox, $S\in\dnfone (C,{\cal T})$ and $N_{aF}^*$ the
language of all finite words over the alphabet $N_{aF}$. The $\pcea$ partition of $S$, $\pceaone (S)$, is
defined as
$\pceaone (S)=S_{prop}\cup S_{csp}\cup S_{\exists}\cup S_{\forall}$, where:
\begin{eqnarray}
  S_{prop}&=&\{A:A\in S\mbox{ and $A$ primitive concept}\}  \nonumber  \\
                  &&\cup\{\neg A:\neg A\in S\mbox{ and $A$ primitive concept}\}  \nonumber  \\
S_{csp}&=&
  \left\{
    \begin{array}{l}
                   \{\exists (u_1)(u_2).P:\exists (u_1)(u_2).P\in S\},\mbox{ if $x$ binary}\\
                   \{\exists (u_1)(u_2)(u_3).P:\exists (u_1)(u_2)(u_3).P\in S\},\mbox{ if $x$ ternary}
    \end{array}
  \right.
  \nonumber  \\
  S_{\exists}&=&\{\exists R.C:
  \exists R.C\in S\}  \nonumber  \\
  S_{\forall}&=&\{\forall R.C:
  \forall R.C\in S\}  \nonumber
\end{eqnarray}
\end{definition}
If $\dnfone (C,{\cal T})=\{S_1,\ldots ,S_n\}$ then $C$ is satisfiable $\wrt$ ${\cal T}$
$\iff$ for some $i=1\ldots n$, $S_i$ is satisfiable $\wrt$ ${\cal T}$. On the other
hand, the following conditions are necessary for the satisfiability of an element $S$
of $\dnfone (C,{\cal T})$:
\begin{enumerate}
  \item $S_{prop}$ does not contain $A$ and $\neg A$, where $A$ is
    a primitive concept;
  \item The CSP induced by $S$ (see the definition below) is consistent;
  \item for all concepts $\exists R.D$ in $S_{\exists}$, where $R$ is
    a general, not necessarily functional role, the conjunction
    $D\sqcap\displaystyle\sqcap _{\forall R.D'\in S_{\forall}}D'$ is a
    consistent concept (recursive call of concept
    consistency). This point is expected to clarify the reader the idea
    of distributing all $\forall R$-prefixed concepts over each
    $\exists R$-prefixed concept; and
  \item for all abstract features $f\in N_{aF}$, such that there exists a concept
    $\exists f.D$ in $S_{\exists}$, the conjunction
    $\displaystyle\sqcap _{\exists f.D\in S_{\exists}}D\sqcap\displaystyle\sqcap _{\forall f.D\in S_{\forall}}D$
    is a consistent concept (again, recursive call of concept consistency).
\end{enumerate}
With the help of the just-above explanation, given a set $S\in\dnfone (C,{\cal T})$, we can replace $S$ with the
equivalent set $S^f$ computed as follows:
\begin{enumerate}
  \item The semantics suggests that, for all general, not necessarily
    functional roles $R$, whenever $S$ contains a concept of the form
    $\exists R.D$, the tableaux method would
    create an $R$-successor $S'$ containing the concept $D$ and all concepts $E$ such that
    $\forall R.E$ belongs to $S$:
    \item[---] initialise $T$ to $S$: $T\leftarrow S$
    \item[---] for all elements of $S$ of the form $\exists R.D$:\\
      $T\leftarrow (T\setminus\{\exists R.D\})\cup
                                             \{\exists R.(D\sqcap\displaystyle
                                             \sqcap _{\forall R.E\in S_{\forall}}E)\}$
  \item A similar work has to be done for abstract features $f$ such that $S$ contains
    elements of the form $\exists f.D$, bearing in mind that abstract
    features are functional. For all such $f$, we replace the subset
    $\{\exists f.D:\exists f.D\in S\}\cup\{\forall f.D:\forall f.D\in S\}$ by the
    singleton set
    $\{\exists f.(\sqcap _{\exists f.D\in S}D\sqcap\sqcap _{\forall f.D\in S}D)\}$. The
    motivation, again, comes straight from the semantics: because abstract features are
    functional, only one $f$-successor to $S$ has to be created, which has to satisfy
    all concepts $D$ such that $\exists f.D\in S$, as well as all concepts $D$ such that
    $\forall f.D\in S$:
  \item[---] for all abstract features $f\in N_{aF}$, such that $S$ contains elements
    of the form $\exists f.D$:\\
      $T\leftarrow T\setminus
                    \{\exists f.D:\exists f.D\in S\}$\\
      $T\leftarrow T\cup
                    \{\exists f.(\sqcap _{\exists f.D\in
                      S}D\sqcap\sqcap _{\forall f.D\in S}D)\}$
  \item remove from $T$ all elements of the form $\forall R.D$:
    $T\leftarrow T\setminus S_{\forall}$
  \item $S^f\leftarrow T$
\end{enumerate}
The second $\dnf$ of a concept $C$ $\wrt$ a TBox ${\cal T}$, $\dnftwo (C,{\cal T})$, is now
introduced. This consists of the $\dnfone$ of $C$ $\wrt$ ${\cal T}$, $\dnfone (C,{\cal T})$, as
given by Definition \ref{firstdnf}, in which each element $S$ is replaced with $S^f$, computed as
shown just above. Formally:
\begin{definition}[second DNF]
Let $x\in\{\rcc8 ,\cdalg ,\atra\}$, $C$ be an $\xdl$ concept, and ${\cal T}$ an $\xdl$ TBox. The
second Disjunctive Normal Form ($\dnf2$) of $C$ $\wrt$ ${\cal T}$, $\dnftwo (C,{\cal T})$, is
defined as $\dnftwo (C,{\cal T})=\{S^f:S\in\dnfone (C,{\cal T})\}$.
\end{definition}
Given an $\xdl$ concept $C$ and an $\xdl$ TBox ${\cal T}$, we can now
use the second DNF, $\dnftwo$, to define the closure $(\tpc )^*$ of
$\tpc$, the TBox $T$ augmented with $C$. Initially, $(\tpc )^*=\tpc$,
and no defined concept in $(\tpc )^*$ is marked. Then we repeat the
following process until all defined concepts in $(\tpc )^*$ are
marked. We consider an axiom $B_1\doteq E$ of $(\tpc )^*$ such that $B_1$
is not marked. We mark $B_1$. We compute $\dnftwo (E,(\tpc )^*)$. For
all $S\in\dnftwo (E,(\tpc )^*)$, $S$ is of the form $S_{prop}\cup
S_{csp}\cup S_{\exists}$. For all such $S$, we do the
following. We consider in turn the elements $\exists R.D$ in
$S_{\exists}$. If $D$ is a defined concept of $(\tpc )^*$ then we do
nothing. Otherwise, if $(\tpc )^*$ has an axiom of the form $B_2\doteq 
D$, then we replace $D$ with $B_2$ in $\exists R.D$. Otherwise, we add to $(\tpc )^*$
the axiom $B_2\doteq D$, and we replace, in $S$, $\exists R.D$ with
$\exists R.B_2$. Formally, $(\tpc )^*$ is defined as follows.
\begin{definition}[closure of $\tpc$] Let $C$ be an $\xdl$ concept and $T$ an
  $\xdl$ TBox. The closure $(\tpc )^*$ of $\tpc$ is defined by the
  procedure of Figure \ref{dnf2procedure}. The initial defined concept of $(\tpc
  )^*$ is the same as the initial defined concept of $\tpc$.
\end{definition}
We also need the closure of a concept $C$ $\wrt$ a TBox ${\cal T}$,
$\closure (C,{\cal T})$, which is defined recursively as the union of
$\dnftwo (C,{\cal T})$, and the closures, $\wrt$ ${\cal T}$, of all
concepts $C'$ such that for some $S\in\dnftwo (C,{\cal T})$ and some
general (possibly functional) role $R$, $\exists R.C'\in S$. Formally:
\begin{definition}[closure of a concept $\wrt$ a TBox]
The closure of an $\xdl$ concept $C$ $\wrt$ an $\xdl$ TBox ${\cal T}$,
$\closure (C,{\cal T})$, is defined recursively as follows:
\begin{footnotesize}
\begin{eqnarray}
\closure (C,{\cal T})&=&\dnftwo (C,{\cal T})
               \cup\displaystyle\bigcup _{\exists R.C'\in S\in\dnftwo (C,{\cal T})}\closure (C',{\cal T})\nonumber
\end{eqnarray}
\end{footnotesize}
\end{definition}
\begin{figure}
\begin{enumerate}
  \item[] {\bf Input:} an $\xdl$ concept $C$ and an $\xdl$ TBox $T$
  \item[] {\bf Output:} the closure $\tpcf$ of $\tpc$
  \item[] Initialise $\tpcf$ to $\tpc$: $\tpcf\leftarrow\tpc$;
  \item[] Initially, no defined concept of $\tpcf$ is marked;
  \item[] {\em while}($\tpcf$ contains defined concepts that are not marked)\{
  \item[] \hskip 0.2cm consider a non marked defined concept $B_1$ from $\tpcf$;
  \item[] \hskip 0.2cm let $B_1\doteq E$ be the axiom from $\tpcf$ defining $B_1$;
  \item[] \hskip 0.2cm mark $B_1$;
  \item[] \hskip 0.2cm compute $\dnf2 (E,\tpcf )$;
  \item[] \hskip 0.2cm {\em for} all $\exists R.D\in S\in\dnf2
    (E,\tpcf )$
  \item[] \hskip 0.4cm {\em if} $D$ is not a defined concept of $\tpcf$ {\em then}
  \item[] \hskip 0.6cm {\em if}($\tpcf$ contains an axiom of the form $B_2\doteq D$) {\em then}
  \item[] \hskip 0.8cm replace $\exists R.D$ with $\exists R.B_2$ in $S$;
  \item[] \hskip 0.6cm {\em else}\{
  \item[] \hskip 0.8cm add the axiom $B_2\doteq D$ to $\tpcf$, where $B_2$ is a fresh defined concept:
  \item[] \hskip 1.6cm  $\tpcf\leftarrow\tpcf\cup\{B_2\doteq D\}$;
  \item[] \hskip 0.8cm replace $\exists R.D$ with $\exists R.B_2$ in $S$;
  \item[] \hskip 0.8cm \}
  \item[] \hskip 0.2cm \}
\end{enumerate}
\caption{Closure $\tpcf$ of a TBox $T$ augmented with a concept $C$, $\tpc$.}\label{dnf2procedure}
\end{figure}
\begin{definition}
Let $C$ be an $\xdl$ concept and ${\cal T}$ an $\xdl$ TBox. We denote by:
\begin{enumerate}
  \item $\concretefeatures (C,{\cal T})=\displaystyle\bigcup _{S\in\closure (C,{\cal T})}\concretefeatures (S)$,
    the set of concrete features of $C$ $\wrt$ ${\cal T}$;
  \item $\ncf (C,{\cal T})=|\concretefeatures (C,{\cal T})|$,
    the number of concrete features of $C$ $\wrt$ ${\cal T}$;
  \item $\abstractfeatures (C,{\cal T})=\{f\in N_{aF}:\exists D\mbox{ s. t. }\exists f.D\in S\in\closure (C,{\cal T})\}$,
    the set of abstract features of $C$ $\wrt$ ${\cal T}$;
  \item $\naf (C,{\cal T})=|\abstractfeatures (C,{\cal T})|$,
    the number of abstract features of $C$ $\wrt$ ${\cal T}$;
  \item $\primitiveconcepts (C,{\cal T})=\{A:\exists S\in\closure (C,{\cal T})\mbox{ s. t. }\{A,\neg A\}\cap S_{prop}\not =\emptyset\}$,
    the set of primitive concepts of $C$ $\wrt$ ${\cal T}$;
  \item $\definedconcepts (C,{\cal T})$ is the set of defined concepts 
    in $(\tpc )^*$;
  \item $\econcepts (C,{\cal T})$,
    the set of existential (sub)concepts of $C$ $\wrt$ ${\cal T}$, is
    the union of all $\exists R.D$ such that there exists an axiom $B\doteq E$
    in $(\tpc )^*$ and $S$ in $E$, so that $\exists R.D\in S$;
  \item $\feconcepts (C,{\cal T})=\{\exists f.D\in\econcepts (C,{\cal
      T}):\mbox{ $f$ abstract feature}\}$,
    the set of functional existential concepts of $C$ $\wrt$ ${\cal T}$;
  \item $\reconcepts (C,{\cal T})=\econcepts (C,{\cal T})\setminus\feconcepts (C,{\cal T})$,
    the set of relational existential concepts of $C$ $\wrt$ ${\cal T}$;
  \item $\fbranchingfactor (C,{\cal T})=\naf (C,{\cal T})$, the functional branching factor of $C$ $\wrt$ ${\cal T}$;
  \item $\rbranchingfactor (C,{\cal T})=|\reconcepts (C,{\cal T})|$,
    the relational branching factor of $C$ $\wrt$ ${\cal T}$;
  \item $\branchingfactor (C,{\cal T})=\fbranchingfactor (C,{\cal T})+\rbranchingfactor (C,{\cal T})$,
    the branching factor of $C$ $\wrt$ ${\cal T}$.
\end{enumerate}
\end{definition}
We suppose that the relational existential concepts in $\reconcepts (C,{\cal T})$ are ordered, and refer to the
$i$-th element of $\reconcepts (C,{\cal T})$, $i=1\ldots \rbranchingfactor (C,{\cal T})$, as $\rec _i(C,{\cal T})$. Similarly, we
suppose that the abstract features in $\abstractfeatures (C,{\cal T})$ are ordered, and refer to the
$i$-th element of $\abstractfeatures (C,{\cal T})$, $i=1\ldots
\fbranchingfactor (C,{\cal T})$, as $\af _i(C,{\cal T})$. Together,
they constitute the directions of the weak alternating automaton to be associated with he satisfiability of $C$ $\wrt$ ${\cal T}$.
\begin{definition}\label{branchingfactor}
Let $C$ be an $\xdl$ concept and ${\cal T}$ an $\xdl$ TBox. The
branching tuple of $C$ is given by the ordered $\branchingfactor (C,{\cal T})$-tuple
$\branchingtuple (C,{\cal T})=$\\
$(\rec _1(C,{\cal T}),\ldots ,\rec _{\rbranchingfactor (C,{\cal T})}(C,{\cal T}),
                      \af _1(C,{\cal T}),\ldots ,\af _{\fbranchingfactor (C,{\cal T})}(C,{\cal T}))$
                      of the $\rbranchingfactor (C,{\cal T})$ relational existential concepts in $\reconcepts (C,{\cal T})$ and
                      the $\fbranchingfactor (C,{\cal T})$ abstract features in $\abstractfeatures (C,{\cal T})$.
\end{definition}
Given an $\xdl$ concept $C$ and an $\xdl$ TBox ${\cal T}$, we will be
interested in $k$-ary $\Sigma$-trees (see Definition \ref{karymtree}), $t$,
verifying the following:
\begin{enumerate}
  \item $k=\branchingfactor (C,{\cal T})$; and
  \item $M=2^{\primitiveconcepts (C,{\cal T})}\times\Theta (\concretefeatures (C,{\cal T}),\Delta _{{\cal D}_x})$,
    where $\Theta (\concretefeatures (C,{\cal T}),\Delta _{{\cal D}_x})$ is the set of total functions
    $\theta :\concretefeatures (C,{\cal T})\rightarrow\Delta _{{\cal D}_x}$ associating with each concrete feature $g$ in
    $\concretefeatures (C,{\cal T})$ a concrete value $\theta (g)$ from the spatial concrete domain  $\Delta _{{\cal D}_x}$.
\end{enumerate}
Such a tree will be seen as representing a class of interpretations of the satisfiability of $C$ $\wrt$ ${\cal T}$:
the label $(X,\theta )$ of a node $\alpha\in\{1,\ldots ,\branchingfactor (C,{\cal T})\}^*$, with
$X\subseteq \primitiveconcepts (C,{\cal T})$ and $\theta\in\Theta(\concretefeatures (C,{\cal T}),\Delta _{{\cal D}_x})$,
is to be interpreted as follows:
\begin{enumerate}
  \item $X$ records the information on the primitive concepts that are true at $\alpha$, in all interpretations of the class; and
  \item $\theta :\concretefeatures (C,{\cal T})\rightarrow\Delta _{{\cal D}_x}$ records the values, at the abstract object
    represented by node $\alpha$, of the concrete features $g_1,\ldots ,g_{\ncf (C,{\cal T})}$
    in $\concretefeatures (C,{\cal T})$.
\end{enumerate}
The crucial question is when we can say that an interpretation of the class is a model of $C$
$\wrt$ ${\cal T}$.
To answer the question, we consider (weak) alternating automata on $k$-ary $\Sigma$-trees,
with $k=\branchingfactor (C,{\cal T})$ and $\Sigma =2^{\primitiveconcepts (C,{\cal T})}\times\Theta
(\concretefeatures (C,{\cal T}),\Delta _{{\cal D}_x})$.
We then show how to associate such an automaton with the satisfiability of an $\xdl$ concept $C$ $\wrt$ a weakly cyclic TBox ${\cal T}$, in such a way
that the models of $C$ $\wrt$ ${\cal T}$ coincide with the $k$-ary $\Sigma$-trees accepted by the automaton. The background on
alternating automata has been adapted from \cite{Muller92a}.
\section{Weak alternating automata and $\xdl$ with weakly cyclic Tboxes}
We now provide the required background on weak alternating automata, adapted from \cite{Muller92a} (see also
\cite{Isli93a,Isli96d,Muller87a,Muller95a}). We then show how to associate such an automaton with the satisfiability problem of an $\xdl$ concept $\wrt$ an $\xdl$ weakly cyclic TBox, so that the language accepted by the automaton coincides with the set of models of the concept $\wrt$ to the TBox.
\subsection{Weak alternating automata}
\begin{definition}[free distributive lattice]
Let $S$ be a set of generators. ${\cal L}(S)$ denotes the free distributive lattice generated by $S$.
${\cal L}(S)$ can be thought of as the set of logical formulas built from variables taken from
$S$ using the disjunction and conjunction operators $\vee$ and $\wedge$ (but not the negation
operator $\neg$). In other words, ${\cal L}(S)$ is the smallest set such that:
\begin{enumerate}
  \item for all $s\in S$, $s\in {\cal L}(S)$; and
  \item if $e_1$ and $e_2$ belong to ${\cal L}(S)$, then so do $e_1\wedge e_2$ and
    $e_1\vee e_2$.
\end{enumerate}
\end{definition}
Each element $e\in {\cal L}(S)$ has, up to isomorphism, a unique representation in $\gdnf$
(Disjunctive Normal Form), $e=\bigvee _iC_i$ (each $C_i$ is a conjunction of generators
from $S$, and no $C_i$ subsumes $C_k$, with $k\not =i$). We suppose, without loss of generality, that
each element of ${\cal L}(S)$ is written in such a form. If $e=\bigvee _i\bigwedge _js_{ij}$ is an
element of ${\cal L}(S)$, the dual of $e$ is the element $\tilde{e}=\bigwedge _i\bigvee _js_{ij}$
obtained by interchanging $\vee$ and $\wedge$ ($\bigwedge _i\bigvee _js_{ij}$ is not necessarily in
$\gdnf$).
\begin{definition}[set representation]\label{setrepresentation}
Let $S$ be a set of generators, ${\cal L}(S)$ the free distributive lattice generated by $S$, and
$e$ an element of ${\cal L}(S)$. Write $e$ in DNF as $\bigvee _{i=1}^n\bigwedge _{j=1}^{n_i}s_{ij}$.
The set representation
of $e$, $\setrep (e)$, is the subset of $2^S$ defined as $\{S_1,\ldots ,S_n\}$, with
$S_i=\{s_{i1},\ldots ,s_{in_i}\}$.
\end{definition}
In the following, we denote
by $K$ a set of $k$ directions $d_1,\ldots ,d_k$; by $N_P$ a set of primitive concepts;
by $x$ a spatial RA from the set $\{\rcc8 ,\cdalg ,\atra\}$;
by $N_{cF}$ a finite set of concrete features referring to objects
in $\Delta _{{\cal D}_x}$;
by $\alphabet$ the alphabet $2^{N_P}\times\Theta (N_{cF},\Delta
_{{\cal D}_x})$,
$\Theta (N_{cF},\Delta _{{\cal D}_x})$ being the set of total functions
$\theta :N_{cF}\rightarrow\Delta _{{\cal D}_x}$, associating with each concrete feature $g$ a
concrete value $\theta (g)$ from the spatial concrete domain  $\Delta
_{{\cal D}_x}$;
by $\lits (N_P)$ the set of literals derived from $N_P$ (viewed as a set of atomic
propositions):
$\lits (N_P)=N_P\cup\{\neg A:A\in N_P\}$;
by $c(2^{\lits (N_P)})$ the set of subsets of $\lits (N_P)$ which do
not contain a primitive concept and its negation: $c(2^{\lits
  (N_P)})=\{S\subset\lits (N_P):(\forall A\in N_P)(\{A,\neg
A\}\not\subseteq S)\}$;
by $\consts (x,K,N_{cF})$ the set of constraints of the form
$P(u_1,u_2)$, if $x$ binary, and $P(u_1,u_2,u_3)$, if $x$ ternary, with $P$ being
an $x$ relation, $u_1$, $u_2$ and $u_3$ $\ksgchains$ (i.e., each of
$u_1$, $u_2$ and $u_3$ is of the form $g$ or $d_{i_1}\ldots d_{i_n}g$,
$n\geq 1$ and $n$ finite, the $d_{i_j}$'s being
directions in $K$, and $g$ a concrete feature).
\begin{definition}[alternating automaton on $k$-ary $\alphabet$-trees]\label{buechialtaut}
Let $k\geq 1$ be an integer and $K=\{d_1,\ldots ,d_k\}$ a set of directions.
An alternating automaton on $k$-ary $\alphabet$-trees is a tuple
${\cal A}=({\cal L}(\lits (N_P)\cup\consts (x,K,N_{cF})\cup K\times Q),
           \alphabet ,$ $\delta ,q_0,{\cal F})$,
where
$Q$ is a finite set of states;
$\alphabet$ is the input alphabet (labelling the nodes of the input trees);
$\delta :Q\rightarrow {\cal L}(\lits (N_P)\cup\consts (x,K,N_{cF})\cup K\times Q)$ is the
transition function;
$q_0\in Q$ is the initial state; and
${\cal F}$ defines the acceptance condition:
\begin{enumerate}
  \item ${\cal F}\subseteq Q$ in case of a B\"uchi alternating automaton; and
  \item ${\cal F}\subseteq 2^Q$ in case of a Muller alternating automaton.
\end{enumerate}
\end{definition}
A weak alternating automaton is a special case of a Muller alternating automaton.
\begin{definition}[Weak alternating automaton \cite{Muller92a}]\label{weakalternatingautomaton}
Let ${\cal A}$
be a Muller alternating automaton on $k$-ary $\alphabet$-trees, as defined in
Definition \ref{buechialtaut}. ${\cal A}$ is said to be a weak alternating automaton
if there exists a partition $Q=\bigcup _{i=1}^n Q_i$ of the set $Q$ of
states, and a partial order $\geq$ on the collection of the $Q_i$'s, so that:
\begin{enumerate}
  \item the transition function $\delta$ has the property that, given two states
    $q\in Q_i$ and $q'\in Q_j$, if $q'$ occurs in $\delta (q)$ then $Q_i\geq Q_j$; and
  \item the set ${\cal F}$ giving the acceptance condition is a subset of
    $\{Q_1,\ldots ,Q_n\}$.
\end{enumerate}
\end{definition}
Let ${\cal A}$ be an alternating automaton on $k$-ary
$\alphabet$-trees, as defined in Definition \ref{buechialtaut}, and $t$ a $k$-ary
$\alphabet$-tree. Given two alphabets $\Sigma _1$ and $\Sigma _2$, we
denote by $\Sigma _1\Sigma _2$ the concatenation of $\Sigma _1$ and
$\Sigma _2$, consisting of all words $ab$, with $a\in\Sigma _1$
and $b\in\Sigma _2$. In a run $r({\cal A},t)$ of ${\cal A}$ on $t$ (see below),
which can be seen as an unfolding of a branch of the computation tree
$T({\cal A},t)$ of ${\cal A}$ on $t$, as defined in
\cite{Muller87a,Muller92a,Muller95a}, the nodes of level $n$ will represent one possibility
for choices of ${\cal A}$ up to level $n$ in $t$.
For each $n\geq 0$, we define the set of
$n$-histories to be the set
$H_n=\{q_0\}(KQ)^n$ of all $2n+1$-length words
consisting of $q_0$ as the first letter, followed by a $2n$-length
word $d_{i_1}q_{i_1}\ldots d_{i_n}q_{i_n}$, with $d_{i_j}\in K$ and $q_{i_j}\in Q$, for all
$j=1\ldots n$. If $h\in H_n$ and
$g\in KQ$ then $hg$, the
concatenation of $h$ and $g$, belongs to $H_{n+1}$. More generally, if $h\in H_n$ and
$e\in {\cal L}(KQ)$, the
concatenation $he$ of $h$ and $e$ will denote the element of ${\cal L}(H_{n+1})$ obtained by
prefixing $h$ to each generator in
$KQ$ which occurs in $e$.
Additionally, given an $n$-history $h=q_0d_{i_1}q_{i_1}\ldots d_{i_n}q_{i_n}$, with $n\geq 0$, we denote
\begin{enumerate}
  \item by $\last (h)$ the initial state $q_0$ if $h$ consists of the
    $0$-history $q_0$ ($n=0$),
and the state $q_{i_n}$ if $n\geq 1$;
  \item by $\kproj (h)$ (the $K$-projection of $h$) the empty word
    $\epsilon$ if $n=0$, and the $n$-length word $d_{i_1}\ldots d_{i_n}$ otherwise; and
  \item by $\qproj (h)$ (the $Q$-projection of $h$) the state $q_0$ if 
    $n=0$, and the $n+1$-length 
    word $q_0q_{i_1}\ldots q_{i_n}\in Q^{n+1}$ otherwise.
\end{enumerate}
The union of all $H_n$, with $n$ finite, will be referred to as the
set of finite histories of ${\cal A}$, and denoted by $\hf$. We denote by $\alphabett$ the alphabet
$2^{\hf}\times c(2^{\lits (N_P)})\times 2^{\consts (x,K,N_{cF})}$,
by $\alphabettt$ the alphabet $2^Q\times c(2^{\lits (N_P)})\times
2^{\consts (x,K,N_{cF})}$, and, in general, by $\alphabtt$ the
alphabet $S\times c(2^{\lits (N_P)})\times
2^{\consts (x,K,N_{cF})}$.

A run of the alternating automaton ${\cal A}$ on $t$ is now introduced.
\begin{definition}[Run]\label{definitionrun}
Let ${\cal A}$ be an alternating automaton on $k$-ary
$\alphabet$-trees, as defined in Definition \ref{buechialtaut}, and $t$ a $k$-ary
$\alphabet$-tree. A run, $r({\cal A},t)$, of ${\cal A}$ on $t$ is a
partial $k$-ary $\alphabett$-tree defined
inductively as follows. For all directions $d\in K$, and for all nodes $u\in K^*$ of
$r({\cal A},t)$, $u$ has at most one outgoing edge labelled with $d$, 
and leading to the $d$-successor $ud$ of $u$. The label
$(Y_{\epsilon},L_{\epsilon},X_{\epsilon})$ of the root belongs to $2^{H_0}\times c(2^{\lits (N_P)})\times 2^{\consts (x,K,N_{cF})}$
---in other words, $Y_{\epsilon}=\{q_0\}$. If $u$ is a
node of $r({\cal A},t)$ of level $n\geq 0$, with label
$(Y_u,L_u,X_u)$, then calculate
$e=\bigwedge _{h\in Y_u}\distribute (h,\delta (\last (h)))$,
where $\distribute$ is a function associating with each pair $(h_1,e_1)$
of $\hf\times {\cal L}(\lits (N_P)\cup\consts (x,K,N_{cF})\cup
K\times Q)$ an element of ${\cal L}(\lits (N_P)\cup\consts
(x,K,N_{cF})\cup\hf)$ defined inductively in the following way:\\
$
\distribute (h_1,e_1)=
  \left\{
                   \begin{array}{ll}
                         e_1
                               &\mbox{ if }e_1\in \lits (N_P)\cup\consts (x,K,N_{cF}),  \\
                         h_1dq
                               &\mbox{ if }e_1=(d,q)\mbox{, for some
                                 }(d,q)\in K\times Q,  \\
                         \distribute (h_1,e_2)\vee\distribute (h_1,e_3)
                               &\mbox{ if }e_1=e_2\vee e_3,  \\
                         \distribute (h_1,e_2)\wedge\distribute (h_1,e_3)
                               &\mbox{ if }e_1=e_2\wedge e_3  \\
                   \end{array}
  \right.
$\\
Write $e$ in $\dnf$ as
$e=\bigvee _{i=1}^r(L_i\wedge X_i\wedge Y_i)$, where the $L_i$'s are
conjunctions of literals from $\lits (N_P)$, the $X_i$'s are
conjunctions of constraints from $\consts
(x,K,N_{cF})$, and the $Y_i$'s are
conjunctions of $n+1$-histories.
Then there exists $i=1\ldots r$ such that
$L_u=\{\ell\in\lits (N_P):\ell\mbox{ occurs in }L_i\}$;
$X_u=\{x\in\consts (x,K,N_{cF}):x\mbox{ occurs in }X_i\}$;
for all $d\in K$, such that the set
$Y=\{hdq\in H_{n+1}:h\in H_n\wedge q\in Q\wedge (hdq\mbox{ occurs in }Y_i)\}$
is nonempty, and only for those $d$, $u$ has a $d$-successor, $ud$, whose label
$(Y_{ud},X_{ud},L_{ud})$ is such that
$Y_{ud}=Y$; and
the label $t(u)=({\cal P}_u,\theta _u)\in 2^{N_P}\times\Theta (N_{cF},\Delta
_{{\cal D}_x})$ of the node
$u$ of the input tree $t$ verifies the following, where, given a
        node $v$ in $t$, the notation $\theta _v$ consists of the function $\theta _v:N_{cF}\rightarrow\Delta _{{\cal D}_x}$ which is the second
        argument of $t(v)$:
    \begin{enumerate}
      \item[$\bullet$] for all $A\in N_P$: if $A\in L_u$ then $A\in {\cal P}_u$; and
        if $\neg A\in L_u$ then $A\notin {\cal P}_u$ (the elements $A$ of
        $N_P$ such that, neither $A$ nor $\neg A$ occur in $L_u$, may or may not
        occur in ${\cal P}_u$);
      \item[$\bullet$] if $x$ binary, for all $P(d_{i_1}\ldots
        d_{i_n}g_1,d_{j_1}\ldots d_{j_m}g_2)$ appearing in $X_u$,
        $P(\theta _{ud_{i_1}\ldots d_{i_n}}(g_1),$ $\theta
        _{ud_{j_1}\ldots d_{j_m}}(g_2))$ holds. In other words, the value of 
        the concrete feature $g_1$ at the $d_{i_1}\ldots
        d_{i_n}$-successor of $u$ in $t$, on the one hand, and the
        value of the concrete feature $g_2$ at the $d_{j_1}\ldots
        d_{j_m}$-successor of $u$ in $t$, on the other hand, are
        related by the $x$ relation $P$.
      \item[$\bullet$] similarly, if $x$ ternary, for all $P(d_{i_1}\ldots
        d_{i_n}g_1,d_{j_1}\ldots d_{j_m}g_2,d_{l_1}\ldots d_{l_p}g_3)$ appearing in $X_u$,
        $P(\theta _{ud_{i_1}\ldots d_{i_n}}(g_1),\theta
        _{ud_{j_1}\ldots d_{j_m}}(g_2),\theta _{ud_{l_1}\ldots d_{l_p}}(g_3))$ holds.
    \end{enumerate}
A partial $k$-ary $\alphabett$-tree $\sigma$ is a run of ${\cal A}$ if there exists a $k$-ary
$\alphabet$-tree $t$ such that $\sigma$ is a run of ${\cal A}$ on $t$.
\end{definition}
\begin{definition}[CSP of a run]\label{cspofarun}
Let ${\cal A}$ be an alternating automaton on $k$-ary
$\alphabet$-trees, as defined in Definition \ref{buechialtaut}, and $\sigma$ a run of
${\cal A}$:
\begin{enumerate}
  \item for all nodes $v$ of $\sigma$, of label
$\sigma (v)=(Y_v,L_v,X_v)\in2^{\hf}\times c(2^{\lits (N_P)})\times
2^{\consts (x,K,N_{cF})}$, the argument $X_v$ gives rise to the CSP of $\sigma$ at $v$,
    $\csp _v(\sigma )$, whose set of variables, $V_v(\sigma )$, and set of constraints,
    $C_v(\sigma )$, are defined as follows:
  \begin{enumerate}
    \item Initially, $V_v(\sigma )=\emptyset$ and $C_v(\sigma
      )=\emptyset$
    \item for all $\ksgchains$ $d_{i_1}\ldots d_{i_n}g$ appearing in
      $X_v$, create, and add to $V_v(\sigma )$, a variable $\langle vd_{i_1}\ldots d_{i_n},g\rangle$
    \item if $x$ binary, for all $P(d_{i_1}\ldots d_{i_n}g_1,d_{j_1}\ldots d_{j_m}g_2)$ in $X_v$, add the constraint\\
      $P(\langle vd_{i_1}\ldots d_{i_n},g_1\rangle,\langle vd_{j_1}\ldots d_{j_m},g_2\rangle)$ to $C_v(\sigma )$
    \item similarly, if $x$ ternary, for all $P(d_{i_1}\ldots
      d_{i_n}g_1,d_{j_1}\ldots d_{j_m}g_2,d_{l_1}\ldots d_{l_p}g_3)$
      in $X_v$,
      add the constraint
      $P(\langle vd_{i_1}\ldots d_{i_n},g_1\rangle,\langle
      vd_{j_1}\ldots d_{j_m},g_2\rangle,\langle
      vd_{l_1}\ldots d_{l_p},g_3\rangle)$ to $C_v(\sigma )$
  \end{enumerate}
  \item the CSP of $\sigma$, $\csp (\sigma )$, is the CSP whose set of variables,
    ${\cal V}(\sigma )$, and set of constraints, ${\cal C}(\sigma )$, are defined
    as ${\cal V}(\sigma )=\displaystyle\bigcup _{v\mbox{ node of }\sigma}V_v(\sigma )$ and
    ${\cal C}(\sigma )=\displaystyle\bigcup _{v\mbox{ node of }\sigma}C_v(\sigma )$.
\end{enumerate}
\end{definition}
An $n$-branch of a run $\sigma =r({\cal A},t)$ is a path of
length (number of edges) $n$ beginning at the root of
$\sigma$. A branch is an infinite path. If $u$ is the terminal node of an
$n$-branch $\beta$, then the argument $Y_u$ of the label $(Y_u,L_u,X_u)$ of
$u$ is a set of $n$-histories. Following \cite{Muller92a}, we say that
each $n$-history in $Y_u$ lies along $\beta$. An $n$-history $h$ lies
along $\sigma$ if there exists an $n$-branch $\beta$ of $\sigma$ such
that $h$ lies along $\beta$.
An (infinite) history is a sequence
$h=q_0d_{i_1}q_{i_1}\ldots d_{i_n}q_{i_n}\ldots\in\{q_0\}(KQ)^\omega$.
Given such a history,
$h=q_0d_{i_1}q_{i_1}\ldots d_{i_n}q_{i_n}\ldots\in\{q_0\}(KQ)^\omega$:
\begin{enumerate}
  \item $h$ lies along a branch $\beta$ if, for every $n\geq 1$, the prefix of $h$
    consisting of the $n$-history
    $q_0d_{i_1}q_{i_1}\ldots d_{i_n}q_{i_n}$ lies along the
    $n$-branch $\beta _n$ consisting of the first $n$ edges of
    $\beta$;
  \item $h$ lies along $\sigma$ if there exists a branch $\beta$ of
    $\sigma$ such that $h$ lies along $\beta$;
  \item $\qproj (h)$ (the $Q$-projection of $h$) is the infinite word
    $q_0q_{i_1}\ldots q_{i_n}\ldots \in Q^\omega$ such that, for all $n\geq 1$, the
    $n+1$-length prefix $q_0q_{i_1}\ldots q_{i_n}$ is the $Q$-projection of
    $h_n$, the $n$-history which is the $2n+1$-prefix of $h$.
  \item we denote by $\infinity (h)$ the set of
states appearing infinitely often in $\qproj (h)$
\end{enumerate}
The acceptance condition is now defined as follows. In the B\"uchi case, a history $h$ is accepting if
$\infinity (h)\cap {\cal F}\not =\emptyset$. In the case of a weak
alternating automaton, $h$ is accepting if $\infinity (h)\subseteq Q_i$, 
for some $Q_i\in {\cal F}$.\footnote{In the case of a weak alternating 
  automaton, if $h$ is an (infinite) history, then from some point
  onwards, all the states occurring in $h$ belong to the same element
  $Q_i$ of the partition associated with the set of states. In
  \cite{Muller92a}, the element $Q_i$ is referred to as the finality
  of $h$, and is denoted by $f(h)$. $Q_i$ is the finality of $h$ is
  equivalent to $\infinity (h)\subseteq Q_i$. This observation will be 
  made use of in the proof of Theorem \ref{frunthm}.} A branch
$\beta$ of $r({\cal A},t)$ is accepting if every history lying along
$\beta$ is accepting.

The condition for a run $\sigma$ to be accepting splits into two subconditions. The
first subcondition is the standard one, and is related to (the
histories lying along) the
branches of $\sigma$, all of which should be accepting. The second subcondition
is new and is the same for both kinds of automata: the CSP of $\sigma$, $\csp (\sigma )$,
should be consistent. ${\cal A}$ accepts a $k$-ary $\alphabet$-tree $t$ if there exists an accepting
run of ${\cal A}$ on $t$. The language ${\cal L}({\cal A})$ accepted by ${\cal A}$ is
the set of all $k$-ary $\alphabet$-trees accepted by ${\cal A}$.

Informally, a run $\sigma$ is uniform if, for all $n\geq 0$, any two
$n$-histories lying along $\sigma$, and suffixed (i.e., terminated) by the
same state, make the same transition. To define it formally, we
suppose that the transition function $\delta$ is given as a
disjunction of conjunctions, in $\dnf$.
\begin{definition}[Uniform run]
Let ${\cal A}$ be an alternating automaton on $k$-ary $\alphabet$-trees, as defined in
Definition \ref{buechialtaut}, and $\sigma$ a run of ${\cal A}$. $\sigma$ is said to
be a uniform run $\iff$ it satisfies the following.
For all $n\geq 0$, select 
for each state $q$ in
$Q$, one conjunct from $\delta (q)$, and refer to it as
$\delta (q,\sigma ,n)$. If $u$ is a
node of $\sigma$ of level $n\geq 0$, with label
$(Y_u,L_u,X_u)$, then calculate
$e=\bigwedge _{h\in Y_u}\distribute (h,\delta (\last (h),\sigma ,n))$,
where $\distribute$ is defined as in Definition \ref{definitionrun}.
Write $e$ as
$e=L\wedge X\wedge Y$, where $L$ is a
conjunction of literals from $\lits (N_P)$, $X$ is a
conjunction of constraints from $\consts
(x,K,N_{cF})$, and $Y$ is a
conjunction of $n+1$-histories.
Then
$L_u=\{\ell\in\lits (N_P):\ell\mbox{ occurs in }L\}$;
$X_u=\{x\in\consts (x,K,N_{cF}):x\mbox{ occurs in }X\}$;
for all $d\in K$, such that the set
$Z=\{hdq\in H_{n+1}:h\in H_n\wedge q\in Q\wedge (hdq\mbox{ occurs in }Y)\}$
is nonempty, and only for those $d$, $u$ has a $d$-successor, $ud$, whose label
$(Y_{ud},X_{ud},L_{ud})$ is such that
$Y_{ud}=Z$.
\end{definition}
The uniformisation theorem for alternating automata, as defined in
\cite{Muller92a,Muller95a}, states that
the existence of an accepting run of ${\cal A}$ is equivalent to the existence
of an accepting uniform run of ${\cal A}$. However, the accepting condition in
\cite{Muller92a,Muller95a} involves only the states repeated infinitely often
in the branches of the run, and this is mainly due to the fact that the input
alphabet is a simple set of symbols. In our case, as already
explained, the input alphabet is the set
$2^{N_P}\times\Theta (N_{cF},\Delta _{{\cal D}_x})$,
$\Theta (N_{cF},\Delta _{{\cal D}_x})$ being the set of total functions
$\theta :N_{cF}\rightarrow\Delta _{{\cal D}_x}$, associating with each
concrete feature $g$, at each node of a run, a
concrete value $\theta (g)$ from the spatial concrete domain  $\Delta _{{\cal D}_x}$. In addition to
the condition on the states infinitely often repeated on each branch of a run, one has
also to consider the constraints on the values of the different concrete features at
the different nodes of the run. The set of all such constraints, over the nodes of a
run, gives birth to what we have named ``CSP of the run'' (Definition \ref{cspofarun}),
which is a potentially infinite CSP. The uniformisation theorem was used in
\cite{Muller92a} to show that, a weak alternating automaton $M$ of size
(number of states) $|M|$ can be simulated by a (standard)
nondeterministic B\"uchi automaton of size $|M|4^{|M|}$. We will define a
``forgetful run" of a (weak) alternating automaton, or $\frun$ for
short, which, intuitively, is a $0$-memory run, in the sense that it
does not keep track of the histories of a branch leading to a node, but just of the states ending
such histories ---i.e., the states $q$ such that, if the node is of
level $n$, there exists an $n$-history $h_n$ lying along the branch
leading to the node, and such that $\last (h_n)=q$. We will then show
that the existence of an accepting run of weak alternating automaton
${\cal A}$ on a tree $t$, is equivalent to the existence of an $\frun$
of ${\cal A}$ on $t$. In particular, this will improve by an
exponential factor the bound on the size of the nondeterministic
B\"uchi automaton simulating a weak alternating automaton, which will be shown 
to be $2^{|M|}$, instead of the $|M|4^{|M|}$ bound in
\cite{Muller92a}.
\begin{definition}[Forgetful run]
Let ${\cal A}$ be an alternating automaton on $k$-ary
$\alphabet$-trees, as defined in Definition \ref{buechialtaut}, and $t$ a $k$-ary
$\alphabet$-tree. A forgetful run, $\fr ({\cal A},t)$, of ${\cal A}$
on $t$ is a partial $k$-ary $\alphabettt$-tree defined
inductively as follows. For all directions $d\in K$, and for all nodes $u\in K^*$ of
$\fr ({\cal A},t)$, $u$ has at most one outgoing edge labelled with $d$, 
and leading to the $d$-successor $ud$ of $u$. The label
$(Y_{\epsilon},L_{\epsilon},X_{\epsilon})$ of the root belongs to
$2^{\{q_0\}}\times c(2^{\lits (N_P)})\times 2^{\consts (x,K,N_{cF})}$
---in other words, $Y_{\epsilon}=\{q_0\}$. If $u$ is a
node of $\fr ({\cal A},t)$ of level $n\geq 0$, with label
$(Y_u,L_u,X_u)$, then calculate
$e=\bigwedge _{q\in Y_u}\delta (q)$. Write $e$ in $\dnf$ as
$e=\bigvee _{i=1}^r(L_i\wedge X_i\wedge Y_i)$, where the $L_i$'s are
conjunctions of literals from $\lits (N_P)$, the $X_i$'s are
conjunctions of constraints from $\consts
(x,K,N_{cF})$, and the $Y_i$'s are
conjunctions of \mbox{direction-state} pairs from $K\times Q$.
Then there exists $i=1\ldots r$ such that
$L_u=\{\ell\in\lits (N_P):\ell\mbox{ occurs in }L_i\}$;
$X_u=\{x\in\consts (x,K,N_{cF}):x\mbox{ occurs in }X_i\}$;
for all $d\in K$, such that the set
$Y=\{q\in Q:\mbox{ $(d,q)$ occurs in }Y_i)\}$
is nonempty, and only for those $d$, $u$ has a $d$-successor, $ud$, whose label
$(Y_{ud},X_{ud},L_{ud})$ is such that
$Y_{ud}=Y$; and
the label $t(u)=({\cal P}_u,\theta _u)\in 2^{N_P}\times\Theta (N_{cF},\Delta
_{{\cal D}_x})$ of the node
$u$ of the input tree $t$ verifies the following, where, given a
        node $v$ in $t$, the notation $\theta _v$ consists of the function $\theta _v:N_{cF}\rightarrow\Delta _{{\cal D}_x}$ which is the second
        argument of $t(v)$:
    \begin{enumerate}
      \item[$\bullet$] for all $A\in N_P$: if $A\in L_u$ then $A\in {\cal P}_u$; and
        if $\neg A\in L_u$ then $A\notin {\cal P}_u$ (the elements $A$ of
        $N_P$ such that, neither $A$ nor $\neg A$ occur in $L_u$, may or may not
        occur in ${\cal P}_u$);
      \item[$\bullet$] if $x$ binary, for all $P(d_{i_1}\ldots
        d_{i_n}g_1,d_{j_1}\ldots d_{j_m}g_2)$ appearing in $X_u$,
        $P(\theta _{ud_{i_1}\ldots d_{i_n}}(g_1),$ $\theta
        _{ud_{j_1}\ldots d_{j_m}}(g_2))$ holds. In other words, the value of 
        the concrete feature $g_1$ at the $d_{i_1}\ldots
        d_{i_n}$-successor of $u$ in $t$, on the one hand, and the
        value of the concrete feature $g_2$ at the $d_{j_1}\ldots
        d_{j_m}$-successor of $u$ in $t$, on the other hand, are
        related by the $x$ relation $P$.
      \item[$\bullet$] similarly, if $x$ ternary, for all $P(d_{i_1}\ldots
        d_{i_n}g_1,d_{j_1}\ldots d_{j_m}g_2,d_{l_1}\ldots d_{l_p}g_3)$ appearing in $X_u$,
        $P(\theta _{ud_{i_1}\ldots d_{i_n}}(g_1),\theta
        _{ud_{j_1}\ldots d_{j_m}}(g_2),\theta _{ud_{l_1}\ldots d_{l_p}}(g_3))$ holds.
    \end{enumerate}
A partial $k$-ary $\alphabettt$-tree $\sigma$ is an $\frun$ of ${\cal A}$ if there exists a $k$-ary
$\alphabet$-tree $t$ such that $\sigma$ is an $\frun$ of ${\cal A}$ on $t$.
\end{definition}
A run $\sigma$ can give rise to one and only one $\frun$, $\sigma '$, which is obtained by
replacing, in the argument $Y_u\subset\hf$ of the label of a node $u$
of $\sigma$, of, say, level $n$, each $n$-history $h_n$ by $\last
(h_n)$. We say that the run $\sigma$ is the generator of the $\frun$ $\sigma '$, and
denote this by $\sigma =\generator (\sigma ')$. The label
$(Y_u^{\sigma},L_u^{\sigma},X_u^{\sigma})$ of a node $u$ in $\sigma$, and the label
$(Y_u^{\sigma '},L_u^{\sigma '},X_u^{\sigma '})$ of the same node but in $\sigma '$, verify
$Y_u^{\sigma '}=\{\last (h):h\in Y_u^{\sigma}\}$, $L_u^{\sigma
  '}=L_u^{\sigma}$, and $X_u^{\sigma '}=X_u^{\sigma}$.

Let $\sigma$ be an $\frun$. We define a forgetful $n$-history, $n\geq 1$, of $\sigma$,
or f-$n$-history of $\sigma$ for
short, as an $n+1$-length word $u=\{q_0\}v$ over the alphabet $2^Q$
---$v\in (2^Q)^n$; and a forgetful history, or f-history for short, as 
an $\omega$-word from $\{q_0\}(2^Q)^{\omega}$. The f-$0$-history of $\sigma$ is simply
$\{q_0\}$. The f-$0$-history $\{q_0\}$ lies along the $0$-branch
reducing to the root of $\sigma$. An f-$n$-history $\Gamma
_0\ldots\Gamma _n\in (2^Q)^{n+1}$ lies along an
$n$-branch $\beta _n$ of $\sigma$ if, for all nodes $u$ of $\beta _n$, 
of level $i$, the first argument $Y_u$ of the label $\sigma (u)$ of $u$ verifies
$Y_u=\Gamma _i$. An f-history $h$ lies along a branch $\beta$ of
$\sigma$ if, for all $n\geq 0$, the f-$n$-history consisting of the
$n+1$-length prefix of $h$ lies along the $n$-branch consisting of the 
first $n$ edges of $\beta$. $\infinity (h)$ is the set of subsets of $Q$
infinitely often repeated in $h$. For all (infinite) branch $\beta$ of $\sigma$, 
there is one and only one f-history lying along $\beta$, which we
refer to as $\fnh (\beta )$. A branch $\beta$ of $\sigma$ is accepting
if the union of the elements of $\infinity (\fnh (\beta ))$ is a
subset of the union of elements of ${\cal F}$; i.e., if
$\displaystyle\bigcup _{S\in\infinity (\fnh (\beta ))}S\subseteq\displaystyle\bigcup
_{F\in {\cal F}}F$. The $\frun$ $\sigma$ is accepting $\iff$ all its
branches are accepting. We also refer to $\infinity (\fnh (\beta ))$
as $\infinity (\beta )$.
\begin{theorem}\label{frunthm}
Let ${\cal A}$ be a weak alternating automaton on $k$-ary
$\alphabet$-trees, and $t$ a $k$-ary
$\alphabet$-tree. There exists an accepting run of ${\cal A}$ on $t$
$\iff$ there exists an accepting $\frun$ of ${\cal A}$ on $t$.
\end{theorem}
{\bf Proof:} We show the following. Given an $\frun$ $\sigma$ of ${\cal A}$ 
on $t$, $\sigma$ is accepting $\iff$ the run $\generator (\sigma )$ is
accepting. The CSP of $\sigma$ and the CSP of $\generator (\sigma )$
are the same. So we only need look at the accepting subcondition
related to the states infinitely often repeated.

Suppose, to start with, that $\generator (\sigma )$ is
accepting. Consider a branch
$\beta _g$ in $\generator (\sigma )$, and the correponding branch
$\beta$ of $\sigma$. $\beta _g$ is accepting: the states infinitely
often repeated in a history $h$ lying along
$\beta _g$ are elements of a set $Q_h\in {\cal F}$.
The union of the subsets of $Q$ infinitely often repeated in the history
$\fnh (\beta )$ of $\sigma$,
is a subset of the union of all such $Q_h$ over the histories lying along
$\beta _g$: $\displaystyle\bigcup _{S\in\infinity (\fnh (\beta ))}S\subseteq\displaystyle\bigcup
_{h\mbox{ lies along }\beta _g}Q_h$. According to our definition of 
an accepting branch of an $\frun$, $\beta$ is clearly accepting,
since, for all $h$ lying along $\beta _g$, $Q_h$ belongs to ${\cal 
  F}$. It follows that $\sigma$ is accepting.

Conversely, suppose that $\sigma$ is
accepting. We need to show that $\generator (\sigma )$ is
accepting. Consider a branch $\beta _g$ of $\generator (\sigma )$, and
the corresponding branch $\beta$ in $\sigma$. Let $Q^f$ be the union
of the subsets of $Q$ infinitely often repeated 
in $\infinity (\fnh (\beta ))$. $\beta$ being accepting, $Q^f$ is a
subset of the union of all elements of ${\cal F}$:
\begin{equation}
Q^f\subseteq\displaystyle\bigcup
_{F\in {\cal F}}F\label{qfone}
\end{equation}
The key point now is that, the set $Q^f$ can also be 
seen as the union of the $\infinity (h)$'s over the histories $h$ lying along $\beta
_g$:
\begin{equation}
Q^f=\displaystyle\bigcup _{h\mbox{ lies along }\beta
  _g}\infinity (h)\label{qftwo}
\end{equation}
From (\ref{qfone}) and (\ref{qftwo}), we get:
\begin{equation}
\displaystyle\bigcup _{h\mbox{ lies along }\beta
  _g}\infinity (h)\subseteq\displaystyle\bigcup
_{F\in {\cal F}}F\label{qfthree}
\end{equation}
Suppose now that there exists a history $h_*$ lying
along $\beta _g$, which is not accepting. In concrete terms, this
would mean that:
\begin{equation}
\forall F\in {\cal F}, \infinity (h_*)\not\subseteq F\label{qffour}
\end{equation}
Given the partial order $\geq$ associated with the partition
$Q=\displaystyle\bigcup _{i=1}^nQ_i$ of the set of states $Q$ of ${\cal A}$,
and the decreasing property of the transition function $\delta$, that, given $q\in
Q_i$ and $q'\in Q_j$, if $q'\in\delta (q)$ then $Q_i\geq Q_j$, it
follows that the set $\infinity (h)$ of states infinitely often
repeated in a history $h$, should be a subset of some element $Q_i$
of the partition, $i=1\ldots n$. For $h_*$, in particular, we should
have:
\begin{equation}
\exists i=1\ldots n,\infinity (h_*)\subseteq Q_i\label{qffive}
\end{equation}
The conjunction of (\ref{qffive}) and (\ref{qffour}) implies that,
$\infinity (h_*)$ is disjoint from each of the elements in ${\cal F}$:
\begin{equation}
\forall F\in {\cal F}, \infinity (h_*)\cap F=\emptyset\label{qfsix}
\end{equation}
(\ref{qfsix}) clearly contredicts (\ref{qfthree}). All histories lying 
along $\beta _g$ are thus accepting, and the run $\generator (\sigma
)$ is accepting. \cqfd

A B\"uchi nondeterministic automaton on $k$-ary $\alphabet$-trees can
be thought of as a special case of a B\"uchi alternating automaton: as 
one that sends at most one copy per direction in a run. In other
words, as a B\"uchi alternating automaton with the property that,
there is one and only one history lying on any branch of any run of the
automaton.
\begin{definition}[B\"uchi nondeterministic automaton on $k$-ary $\alphabet$-trees]\label{buechinondetaut}
A B\"uchi nondeterministic automaton on $k$-ary $\alphabet$-trees is a tuple\\
${\cal B}=(Q,K,\lits (N_P),\consts (x,K,N_{cF}),
           \alphabet ,\delta ,q_0,q_{\#},{\cal F})$,
where
$Q$ is a finite set of states;
$K=\{d_1,\ldots ,d_k\}$ ($k\geq 1$) is a set of directions;
$\lits (N_P)$, $\consts (x,K,N_{cF})$ and
           $\alphabet$
are as in Definition \ref{buechialtaut};
$q_0\in Q$ is the initial state;
$q_{\#}\in Q$ is a state indicating that no
copy has to be sent in the corresponding direction;
${\cal F}$ defines the acceptance condition; and
$\delta :Q\rightarrow {\cal P}(2^{\lits (N_P)}\times 2^{\consts
  (x,K,N_{cF})}\times (Q\cup\{q_{\#}\})^k)$ is the
transition function.
\end{definition}
\begin{definition}[Run of a B\"uchi automaton]
Let ${\cal B}$ be a B\"uchi nondeterministic automaton on $k$-ary
$\alphabet$-trees, as defined in Definition \ref{buechinondetaut}, and $t$ a $k$-ary
$\alphabet$-tree. A run, $r ({\cal B},t)$, of ${\cal B}$ on $t$ is a
partial $k$-ary $\alphabt$-tree\footnote{$\alphabt =Q\times c(2^{\lits (N_P)})\times
2^{\consts (x,K,N_{cF})}$.} defined
inductively as follows. For all directions $d\in K$, and for all nodes $u\in K^*$ of
$r({\cal B},t)$, $u$ has at most one outgoing edge labelled with $d$, 
and leading to the $d$-successor $ud$ of $u$. The label
$(Y_{\epsilon},L_{\epsilon},X_{\epsilon})$ of the root belongs to $\{q_0\}\times c(2^{\lits (N_P)})\times 2^{\consts (x,K,N_{cF})}$
---in other words, $Y_{\epsilon}=q_0$. If $u$ is a
node of $r({\cal B},t)$ of level $n\geq 0$, with label
$(Y_u,L_u,X_u)$, then let
$e=\delta (Y_u)\subseteq 2^{\lits (N_P)}\times 2^{\consts
  (x,K,N_{cF})}\times (Q\cup\{q_{\#}\})^k$.
Then there exists $(L,X,(q_{i_1},\ldots ,q_{i_k}))\in\delta (Y_u)$ such that
$L_u=L$;
$X_u=X$;
for all $j=1\ldots k$, such that
$q_{i_j}\not = q_{\#}$, and only for those $j$, $u$ has a $d_j$-successor, $ud_j$, whose label
$(Y_{ud_j},X_{ud_j},L_{ud_j})$ is such that
$Y_{ud_j}=q_{i_j}$; and
the label $t(u)=({\cal P}_u,\theta _u)\in 2^{N_P}\times\Theta (N_{cF},\Delta
_{{\cal D}_x})$ of the node
$u$ of the input tree $t$ verifies the following, where, given a
        node $v$ in $t$, the notation $\theta _v$ consists of the function $\theta _v:N_{cF}\rightarrow\Delta _{{\cal D}_x}$ which is the second
        argument of $t(v)$:
    \begin{enumerate}
      \item[$\bullet$] for all $A\in N_P$: if $A\in L_u$ then $A\in {\cal P}_u$; and
        if $\neg A\in L_u$ then $A\notin {\cal P}_u$ (the elements $A$ of
        $N_P$ such that, neither $A$ nor $\neg A$ occur in $L_u$, may or may not
        occur in ${\cal P}_u$);
      \item[$\bullet$] if $x$ binary, for all $P(d_{i_1}\ldots
        d_{i_n}g_1,d_{j_1}\ldots d_{j_m}g_2)$ appearing in $X_u$,
        $P(\theta _{ud_{i_1}\ldots d_{i_n}}(g_1),$ $\theta
        _{ud_{j_1}\ldots d_{j_m}}(g_2))$ holds. In other words, the value of 
        the concrete feature $g_1$ at the $d_{i_1}\ldots
        d_{i_n}$-successor of $u$ in $t$, on the one hand, and the
        value of the concrete feature $g_2$ at the $d_{j_1}\ldots
        d_{j_m}$-successor of $u$ in $t$, on the other hand, are
        related by the $x$ relation $P$.
      \item[$\bullet$] similarly, if $x$ ternary, for all $P(d_{i_1}\ldots
        d_{i_n}g_1,d_{j_1}\ldots d_{j_m}g_2,d_{l_1}\ldots d_{l_p}g_3)$ appearing in $X_u$,
        $P(\theta _{ud_{i_1}\ldots d_{i_n}}(g_1),\theta
        _{ud_{j_1}\ldots d_{j_m}}(g_2),\theta _{ud_{l_1}\ldots d_{l_p}}(g_3))$ holds.
    \end{enumerate}
A partial $k$-ary $\alphabt$-tree $\sigma$ is a run of ${\cal B}$ if there exists a $k$-ary
$\alphabet$-tree $t$ such that $\sigma$ is a run of ${\cal B}$ on $t$.
\end{definition}
An $n$-branch and a branch of a run of a B\"uchi nondeterministic
automaton are defined as in the alternating case. Given an $n$-branch
$\beta$, one and only one $n$-history lies along $\beta$, which is
$h=q_0d_{i_1}q_{i_1}\ldots d_{i_n}q_{i_n}\in\{q_0\}(KQ)^n$,
such that, the node given by $\kproj (h)$ is the
terminal node of the $n$-branch, and $q_{i_j}$, $j=1\ldots n$, is the first
argument of the label of the $j$-th node of the $n$-branch.
An (infinite) history
$h=q_0d_{i_1}q_{i_1}\ldots d_{i_n}q_{i_n}\ldots\in\{q_0\}(KQ)^\omega$ lies along a branch $\beta$ if, for every $n\geq 1$, the prefix of $h$
    consisting of the $n$-history
    $q_0d_{i_1}q_{i_1}\ldots d_{i_n}q_{i_n}$ lies along the
    $n$-branch $\beta _n$ consisting of the first $n$ edges of
    $\beta$. A history $h$ is accepting if $\infinity (h)\cap {\cal
      F}\not =\emptyset$. A branch is accepting if the history lying along it is
    accepting. A run is accepting if all its branches are accepting.
The following corollary is a direct consequence of Theorem \ref{frunthm}.
\begin{corollary}\label{ftheorem}
Let ${\cal A}$ be a weak alternating automaton on $k$-ary
$\alphabet$-trees, and $Q$ the set of states of ${\cal A}$. There exists a B\"uchi nondeterministic automaton
simulating ${\cal A}$, with a number of states bounded by $2^{|Q|}$,
$|Q|$ being the size (number of states) of ${\cal A}$.
\end{corollary}
{\bf Proof:}
Let ${\cal A}=({\cal L}(\lits (N_P)\cup\consts (x,K,N_{cF})\cup K\times Q),
           \alphabet ,\delta ,q_0,{\cal F})$ be an alternating automaton on $k$-ary
$\alphabet$-trees, as defined in Definition \ref{buechialtaut}, and
suppose that ${\cal A}$ is weak (Definition
\ref{weakalternatingautomaton}). From the proof of Theorem \ref{frunthm}, the following B\"uchi
nondeterministic automaton,
${\cal B}=(2^Q,K,\lits (N_P),\consts (x,K,N_{cF}),
           \alphabet ,$ $\delta _{{\cal B}},\{q_0\},\{q_{\#}\},{\cal F} _{{\cal B}})$, simulates ${\cal 
             A}$. In particular, the set of states of ${\cal B}$ is
           the set of subsets of $Q$, and the initial state of ${\cal
             B}$ is the singleton subset $\{q_0\}$ of $Q$. The only
           parameters that are not obvious are the transition function
           $\delta _{{\cal B}}$ and the set ${\cal F} _{{\cal B}}$
           providing the acceptance condition. For all $Q_1\in 2^Q$, $\delta
_{{\cal B}}(Q_1)$ is obtained as follows.
An element $(L,X,(Q_{i_1},\ldots ,Q_{i_k}))$ of $2^{\lits (N_P)}\times 2^{\consts
  (x,K,N_{cF})}\times (2^Q\cup\{\{q_{\#}\}\})^k$ belongs to $\delta
_{{\cal B}}(Q_1)$ $\iff$ there exists an $\frun$ $\sigma$ of ${\cal
  A}$, and a node $u$ of $\sigma$, so that, the label $(Y_u,X_u,L_u)$
satisfies $Y_u=Q_1$, $X_u=X$ and $L_u=L$, and
for all $j=1\ldots k$, such that
$Q_{i_j}\not =\{q_{\#}\}$, and only for those $j$, $u$ has a $d_j$-successor, $ud_j$, whose label
$(Y_{ud_j},X_{ud_j},L_{ud_j})$ is such that
$Y_{ud_j}=Q_{i_j}$. The set ${\cal F} _{{\cal B}}$ is ${\cal F}
_{{\cal B}}=\displaystyle\bigcup _{F\in {\cal F}}F$. The condition for 
a history $h$ to be accepting is not $\infinity (h)\cap {\cal F}
_{{\cal B}}\not =\emptyset$, rather $\displaystyle\bigcup _{Q_1\in\infinity (h)}Q_1\subseteq {\cal F}
_{{\cal B}}$ (the explanation lies in the proof of Theorem
\ref{frunthm}). \cqfd

Let $\sigma$ be an $\frun$. Given a branch $\beta$ of $\sigma$,
$\infinity (\beta )$ denotes, as we have seen, the set of subsets of $Q$ infinitely often
repeated in $\beta$: in other words, the set of $Q_1\in 2^Q$ such
that, there exist infinitely many nodes $u$ of $\beta$ so that, the
label $\sigma (u)=(Y_u,X_u,L_u)$ of $u$ verifies $Y_u=Q_1$. By
$\infinity (\sigma )$, we denote the union of all $\infinity (\beta )$
along the branches $\beta$ of $\sigma$: $\infinity (\sigma )=\displaystyle\bigcup _{\beta\mbox{ branch
      of }\sigma}\infinity (\beta )$. The following
corollary is also a direct consequence of Theorem \ref{frunthm}.
\begin{corollary}
Let ${\cal A}$ be a weak alternating automaton on $k$-ary
$\alphabet$-trees, and $t$ a $k$-ary
$\alphabet$-tree.
An $\frun$ $\sigma$ of ${\cal A}$ on $t$ is accepting $\iff$ the union of
the elements of $\infinity (\sigma )$ is a subset of the union of
the elements in ${\cal F}$; in other words, $\iff$ $\displaystyle\bigcup
_{Q_1\in\infinity (\sigma )}Q_1\subseteq\displaystyle\bigcup
_{F\in {\cal F}}F$.
\end{corollary}

{\bf Proof:} Let $\sigma$ be an $\frun$ as described in the
theorem. Suppose, to start with, that $\sigma$ is accepting. As a
consequence, for all branches $\beta$ of $\sigma$, we have $\displaystyle\bigcup
_{Q_1\in\infinity (\beta )}Q_1\subseteq\displaystyle\bigcup
_{F\in {\cal F}}F$. This straightforwardly leads to $\displaystyle\bigcup
_{Q_1\in\infinity (\sigma )}Q_1\subseteq\displaystyle\bigcup
_{F\in {\cal F}}F$. To show the other direction of the corollary,
suppose that $\displaystyle\bigcup
_{Q_1\in\infinity (\sigma )}Q_1\subseteq\displaystyle\bigcup
_{F\in {\cal F}}F$. As an immediate consequence, for all
branches $\beta$ of $\sigma$, we have $\displaystyle\bigcup
_{Q_1\in\infinity (\beta )}Q_1\subseteq\displaystyle\bigcup
_{F\in {\cal F}}F$, which clearly means that the $\frun$ $\sigma$ is
accepting. \cqfd

Deciding whether a standard B\"uchi nondeterministic automaton on $k$-ary
$\Sigma$-trees (see, for instance, \cite{Rabin69a,Rabin70a}) accepts a nonempty language is trivial. The intuitive
idea is to build a partial run, whose size is linear in the number of
states, and with the property that no state appears more than once in
the label of an internal node, though it may appear more than once at
the level of leaves. The kind of B\"uchi automata we are dealing with
is more complicated, due mainly to the use of feature chains to
relate values of different concrete features at different nodes of the
run, which gives rise to what we have named ``CSP of a run",
which is potentially infinite. The rest of the section will show how to extend the method, so
that it can handle the emptiness problem of this new kind of B\"uchi
automata. Thanks to Theorem \ref{frunthm} and Corollary
\ref{ftheorem}, we transform the problem into how to check whether an
$\frun$ of a weak alternating automaton
is accepting. The method is constructive and can easily be used to
derive effective tableaux methods for the problem of deciding
satisfiability of an $\xdl$ concept $\wrt$ an $\xdl$ weakly cyclic TBox. Some additional vocabulary is needed.
\begin{definition}[prefix and lexicographic order]
Let $\Sigma =\{a_1,\ldots ,a_n\}$ be an ordered alphabet, with $a_1<a_2<\cdots <a_n$, and
$u,v\in\Sigma ^*$. The relations ``$u$ is prefix of $v$'', denoted by $\prefix (u,v)$,
and ``$u$ is lexicographically smaller than $v$'', denoted by $u\lleq v$, are defined in
the following obvious manner:
\begin{enumerate}
  \item $\prefix (u,v)$ $\iff$ $v=uw$, for some $w\in\Sigma ^*$
  \item $u\lleq v$ $\iff$, either $\prefix (u,v)$; or $u=w_1aw_2$ and $v=w_1bw_3$, for
    some $w_1,w_2,w_3\in\Sigma ^*$ and $a,b\in\Sigma$, with $a<b$.
\end{enumerate}
\end{definition}
We will also need the derived relations ``$u$ is a strict prefix of $v$'', ``$u$ is
lexicographically strictly smaller than $v$'', and ``$u$ and $v$ are incomparable'',
which we denote, respectively, by $\sprefix (u,v)$, $u\slleq v$ and
$\incomparable (u,v)$:
\begin{enumerate}
  \item $\sprefix (u,v)$ $\iff$ $\prefix (u,v)$ and $u\not = v$
  \item $u\slleq v$ $\iff$ $u\lleq v$ and $u\not = v$
  \item $\incomparable (u,v)$ $\iff$ $\neg\prefix (u,v)$ and $\neg\prefix (v,u)$
\end{enumerate}
\begin{definition}[subtree]
Let $K=\{d_1,\ldots ,d_k\}$ be a set of $k$ directions, $t$ a
partial $k$-ary $\Sigma$-tree,
and $u\in K^*$ a node of $t$. The subtree of $t$ at $u$, denoted $t/u$, is the
partial $k$-ary $\Sigma$-tree $t'$, whose nodes are of the form $v$, so that $uv$ is a 
node of $t$, and, for all such nodes, $t'(v)=t(uv)$ ---i.e., the
label of $v$ in $t'$, is the same as the one of $uv$ in $t$.
\end{definition}
\begin{definition}[substitution]
Let $K=\{d_1,\ldots ,d_k\}$ be a set of $k$ directions, $t$ and $t'$ two
partial $k$-ary $\Sigma$-trees, and
$u\in K^*$ a node of $t$.
The substitution of $t'$ to the subtree of $t$ at $u$, or $u$-substitution of $t'$
in $t$, denoted $t(u\leftarrow t')$, is the partial $k$-ary $\Sigma$-tree $t''$ such
that, the nodes are of the form $v$, with $v$ node of $t$ of which $u$ 
is not a prefix, or of the form $uv$, with $v$ a node of $t'$. The
label $t''(v)$ of $v$ in $t''$ is defined as follows:
$
t''(v)=
  \left\{
                   \begin{array}{ll}
                         t'(w)
                               &\mbox{ if }v=uw\mbox{, for some node
                                 }w\mbox{ of }t',  \\
                         t(v)
                               &\mbox{ otherwise}  \\
                   \end{array}
  \right.
$
\end{definition}
\begin{definition}[cut]
Let $K=\{d_1,\ldots ,d_k\}$ be a set of $k$ directions, $t$ a
partial $k$-ary $\Sigma$-tree, and
$u\in K^*$ a node of $t$.
The cut in $t$ of the subtree at $u$, or $u$-cut
in $t$, denoted $c(u,t)$, is the partial $k$-ary $\Sigma$-tree $t'$
whose nodes are those nodes $v$ of $t$ of which $u$  
is not a strict prefix ---i.e., such that $\neg\sprefix (u,v)$. The
label $t'(v)$ of any node $v$ in $t'$ is the same as $t(v)$, the label of the same node in $t$.
\end{definition}
\begin{figure}
\begin{enumerate}
  \item {\bf Input:} an accepting $\frun$ $\sigma$ of a weak alternating automaton ${\cal A}$.
  \item {\bf Output:} a finite representation, $t$, of a regular
    $\frun$ generated from $\sigma$.
  \item\label{lone} Initialise $t$ to $\sigma$: $t\leftarrow\sigma$;
  \item\label{ltwo} Initially, no node of $t$ is marked;
  \item\label{lthree} {\bf repeat} while possible\{
  \item\label{lfour} \hskip 0.2cm Let $u$ be the smallest node of $t$ such that there exists a 
    non marked node $v$, so that $\slleq (u,v)$
    \underline{and} $Y_u=Y_v$ \underline{and} $\backconstraints (\sigma 
    ,u)=\backconstraints (\sigma ,v)$;
  \item\label{lfive} \hskip 0.2cm choose $v$ as small as possible, $\wrt$ to the
    lexicographic order $\lleq$;
  \item\label{lsix} \hskip 0.2cm if $\neg\prefix (u,v)$\{
  \item\label{lseven} \hskip 0.6cm $t\leftarrow c(v,t)$;
  \item\label{leight} \hskip 0.6cm $\backnode (v)\leftarrow u$;
  \item\label{lnine} \hskip 0.6cm mark $v$;
  \item\label{lten} \hskip 0.6cm \}
  \item\label{leleven} \hskip 0.2cm else \underline{\% $\prefix (u,v)$ \%}
  \item\label{ltwelve} \hskip 0.6cm if nodes $w$ between $u$ and $v$ (i.e., so that $\lleq (u,w)\wedge\lleq (w,v)$) all verify
    $Y_w\subseteq\displaystyle\bigcup _{F\in {\cal F}}F$
    then\{
%
%\footnote{By $w$ between $u$ and $v$, we mean $\lleq (u,w)$ and $\lleq (w,v)$.}
%
  \item\label{lthirteen} \hskip 1cm $t\leftarrow c(v,t)$;
  \item\label{lfourteen} \hskip 1cm $\backnode (v)\leftarrow u$;
  \item\label{lfifteen} \hskip 1cm mark $v$;
  \item\label{lsixteen} \hskip 1cm \}
  \item\label{lseventeen} \hskip 0.6cm else\{
  \item\label{leighteen} \hskip 1cm $t'\leftarrow t/v$;
  \item\label{lnineteen} \hskip 1cm $t\leftarrow t(u\leftarrow t')$;
  \item\label{ltwenty} \hskip 1cm \}
  \item\label{ltwentyone} \hskip 0.2cm \} \underline{\% end {\bf repeat} \%}
\end{enumerate}
\caption{The order $d_1<\ldots <d_k$ is assumed on the directions in $K$.}\label{finitekarytree}
\end{figure}
The last step of our walk towards decidability of the satisfiability
of an $\xdl$ concept $\wrt$ an $\xdl$ weakly cyclic TBox, is to show how to
handle the CSP of an $\frun$, which is potentially infinite. For the
purpose, we need another kind of $\frun$, regular $\frun$, which is
based on a function $\backconstraints$: given an $\frun$ $\sigma$ and
a node $u$ of $\sigma$, $\backconstraints (\sigma ,u)$ consists,
intuitively, of those constraints that are still
unfulfilled at $u$, and which were solicited at nodes $v$ that are
prefixes of $u$. Formally, the function is defined as follows for
the case of $x$ being binary: $\backconstraints (\sigma ,u)=\backconstraints
_l(\sigma ,u)\cup\backconstraints _r(\sigma ,u)$, with
\begin{eqnarray}
\backconstraints _l(\sigma ,u)&=&\{(n,P(d_{i_1}\ldots d_{i_n}v_1g_1,v_2g_2)):(\exists
u_1\in K^*)(u=u_1d_{i_1}\ldots d_{i_n}\wedge  \nonumber  \\
                            & &P(d_{i_1}\ldots
d_{i_n}v_1g_1,v_2g_2)\in X_{u_1})\}  \nonumber  \\
\backconstraints _r(\sigma ,u)&=&\{(n,P(v_1g_1,d_{i_1}\ldots d_{i_n}v_2g_2)):(\exists
u_1\in K^*)(u=u_1d_{i_1}\ldots d_{i_n}\wedge  \nonumber  \\
                            & &P(v_1g_1,d_{i_1}\ldots
d_{i_n}v_2g_2)\in X_{u_1})\}  \nonumber
\end{eqnarray}
\begin{definition}[regular $\frun$]
Let ${\cal A}$ be a weak alternating automaton on $k$-ary
$\alphabet$-trees, as defined in Definition \ref{buechialtaut}, and
$\sigma$ an $\frun$ of ${\cal A}$. $\sigma$ is regular if,
for all nodes $u$ and $v$ of $\sigma$ verifying $\backconstraints
(\sigma ,u)=\backconstraints (\sigma ,v)$, and whose labels $\sigma
(u)=(Y_u,L_u,X_u)$ and $\sigma
(v)=(Y_v,L_v,X_v)$ verify $Y_u=Y_v$, the following holds:
\begin{enumerate}
  \item $L_u=L_v$;
  \item $X_u=X_v$;
  \item for all $d\in K$, $u$ has a $d$-successor $\iff$ $v$ has a
    $d$-successor; and
  \item for all $d\in K$ such that, each of $u$ and $d$ has a
    $d$-successor, it is the case that $Y_{ud}=Y_{vd}$.
\end{enumerate}
\end{definition}
\begin{theorem}\label{gafrunthm}
Let ${\cal A}$ be a weak alternating automaton on $k$-ary
$\alphabet$-trees, and $t$ a $k$-ary
$\alphabet$-tree. There exists an accepting $\frun$ of ${\cal A}$ on $t$
$\iff$ there exists an accepting regular $\frun$ of ${\cal A}$ on $t$.
\end{theorem}
{\bf Proof:} A regular $\frun$ is a particular $\frun$, which 
means that the existence of an accepting regular $\frun$
implies the existence of an accepting $\frun$. To show the other
direction of the equivalence, suppose the existence of an accepting
$\frun$, say $\sigma$. From $\sigma$, we first build a finite partial
$k$-ary $\alphabett$-tree, $t$. We then show how to use $t$ to get an
accepting regular $\frun$ of ${\cal A}$. The tree $t$ is built by the
procedure of Figure \ref{finitekarytree}. The details of the procedure are as
follows:
\begin{enumerate}
  \item[$\bullet$] $u$ and $v$ are chosen so that $\slleq (u,v)$ and $Y_u=Y_v$ and 
    $\backconstraints (\sigma ,u)=\backconstraints (\sigma ,v)$ (line (\ref{lfour}))
  \item[$\bullet$] if $u$ is not prefix of $v$: given that
    $\backconstraints (\sigma ,u)=\backconstraints (\sigma ,v)$, we
    can substitute the subtree of $t$ at $u$ to the subtree of $t$ at
    $v$, and get a run with all branches accepting, and with a global CSP
    consistent. The procedure, however, does not do the
    substitution. Instead, it cuts the subtree at $v$, and marks $u$
    as the successor of $v$, information which will be used in the
    building of the accepting regular run (lines
    (\ref{lseven})-(\ref{leight})-(\ref{lnine}))
  \item[$\bullet$] if $u$ is a (strict) prefix of $v$ then there are
    two possibilities:
    \begin{enumerate}
      \item[\#] if all nodes $w$ between $u$ and $v$ are so that
        $Y_w\subseteq\displaystyle\bigcup _{F\in {\cal F}}F$ (line
        (\ref{ltwelve})) then cutting $t$ at $v$, and then repeating 
        the subtree at $u$ of the resulting tree, will lead to an
        accepting $\frun$, again thanks to $\backconstraints (\sigma
        ,u)=\backconstraints (\sigma ,v)$. What the procedure does in
        this case: it cuts the subtree at $v$, and sets $v$ as a
        repetition of internal node $u$ (lines
    (\ref{lthirteen})-(\ref{lfourteen})-(\ref{lfifteen}))
      \item[\#] the other possibility corresponds to the case when the 
        segment $[u,v]$ does contain nodes $w$ which do not have the
        property $Y_w\subseteq\displaystyle\bigcup _{F\in {\cal
            F}}F$. The procedure shortens the distance to segments
        $[u,v]$ with all nodes $w$ verifying the
        property $Y_w\subseteq\displaystyle\bigcup _{F\in {\cal
            F}}F$ (lines
    (\ref{leighteen})-(\ref{lnineteen})).
    \end{enumerate}
\end{enumerate}
The output tree $t$ of the procedure of Figure \ref{finitekarytree} is so that, each marked node, $v$, is a leaf and is so
that, there is one and only one internal node, $u$, of $t$ such that
$Y_u=Y_v$ and $\backconstraints (\sigma ,u)=\backconstraints (\sigma ,v)$. For each such node $v$, we refer to the corresponding
internal node $u$ as $i_v$, and to the subtree of $t$ at $u$ as
$t/i_v$. From $t$, we now build an accepting regular $\frun$, $\sigma$,
which, intuitively, consists of ``pasting'' infinitely many times such subtrees at the
matching leaves.
\begin{enumerate}
  \item Step $0$: $\sigma _0\leftarrow t$
  \item Step $1$:
  \item  \hskip 0.3cm initialise $\sigma _1$ to $\sigma _0$: $\sigma
    _1\leftarrow\sigma _0$
  \item  \hskip 0.3cm repeat while possible\{
  \item  \hskip 0.6cm consider a marked node $v_1$ of $t$
  \item  \hskip 0.6cm if $\sigma _0$ and $\sigma _1$ have a (same) leaf node $v_2$
    which is marked and so that $Y_{v_2}=Y_{v_1}$\{
  \item  \hskip 0.9cm $\sigma _1\leftarrow\sigma _1 (v_2\leftarrow t/i_{v_1})$
  \item  \hskip 0.9cm if a leaf $i_{v_1}v_3$ of $t$ is marked then mark the
    corresponding leaf $v_2v_3$ of $\sigma _1$
  \item  \hskip 0.9cm \}
  \item  \hskip 0.6cm \}
  \item Step $n$ $(n\geq 2)$:
  \item \hskip 0.3cm initialise $\sigma _n$ to $\sigma _{n-1}$: $\sigma
    _n\leftarrow\sigma _{n-1}$
  \item \hskip 0.3cm repeat while possible\{
  \item \hskip 0.6cm consider a marked node $v_1$ of $t$
  \item \hskip 0.6cm if $\sigma _{n-1}$ and $\sigma _n$ have a (same) leaf node $v_2$
    which is marked and so that $Y_{v_2}=Y_{v_1}$\{
  \item \hskip 0.9cm $\sigma _n\leftarrow\sigma _n(v_2\leftarrow t/i_{v_1})$
  \item \hskip 0.9cm if a leaf $i_{v_1}v_3$ of $t$ is marked then mark the
    corresponding leaf $v_2v_3$ of $\sigma _n$
  \item  \hskip 0.9cm \}
  \item  \hskip 0.6cm \}
\end{enumerate}
The accepting regular $\frun$ $\sigma$ we are looking for now, is nothing else than the partial
$k$-ary $\Sigma$-tree $\sigma _n$ when $n$ tends to $+\infty$. By
construction, $\sigma$ is an $\frun$ of ${\cal A}$. Given that, in
$t$, each marked node $v$ verifying $\lleq (i_v,v)$ is so that, the union of all
$Y_w$, over the nodes $w$ between $i_v$ and $v$, is a subset of
$\displaystyle\bigcup _{F\in {\cal F}}F$, all branches of $\sigma$ are
accepting. And, finally, given that, in $t$, each marked node $v$
verifies $\backconstraints (t,v)=\backconstraints (t,i_v)$, the CSP of 
$\sigma$, $\csp (\sigma )$, is consistent. \cqfd

The following corollary is a direct consequence of Theorem \ref{gafrunthm}.
\begin{corollary}\label{xdlsatisfiability}
There exists a nondeterministic exponential-time algorithm deciding
whether an $\xdl$ concept is satisfiable $\wrt$ an $\xdl$ weakly cyclic TBox.
\end{corollary}
{\bf Proof:} The number of nodes of the output tree $t$ of
the procedure of Figure \ref{finitekarytree}, that are not
marked\footnote{The others are leaf nodes, and are repetitions of
  unmarked, internal nodes.}, is bounded by $2^{|Q|}\times\ell
_{fc}\times 2^{n_c}$, where $Q$ is the set of states of ${\cal A}$, and
$\ell _{fc}$ and $n_c$ are, respectively, the length of the longest $\ksgchain$ and the number of
constraints from $\consts (x,K,N_{cF})$ appearing in the transition
function $\sigma$ of ${\cal A}$. We can thus in nondeterministic exponential-time
build such a tree, if it exists, or report its inexistence, otherwise. \cqfd
\subsection{Associating a weak alternating automaton with the satisfiability of a concept $\wrt$ a weakly cyclic TBox}
We are now ready to describe how to effectively associate with the
satisfiability, $\wrt$ an $\xdl$ weakly cyclic TBox ${\cal T}$,
of an $\xdl$ concept $C$, a weak
alternating automaton ${\cal A}_{C,{\cal T}}$, so that the set of
models of $C$ $\wrt$ ${\cal T}$ coincides with the language accepted
by ${\cal A}_{C,{\cal T}}$ ---in particular, $C$ is insatisfiable
$\wrt$ ${\cal T}$ $\iff$ the language accepted by ${\cal A}_{C,{\cal
    T}}$ is empty.
\begin{definition}
Let $x\in\{\rcc8 ,\cdalg ,\atra\}$,
$C$ an $\xdl$ concept,
${\cal T}$ an $\xdl$ weakly cyclic TBox, $\tpc$ the TBox $T$ augmented 
with $C$, and $B_i$ the initial defined concept of $\tpc$.
With the satisfiability of $C$ $\wrt$ ${\cal T}$, we associate the
weak alternating automaton
${\cal A}_{C,{\cal T}}=({\cal L}(\lits (N_P)\cup\consts (x,K,N_{cF})\cup K\times Q),
           \alphabet ,\delta ,q_0,{\cal F})$ on $k$-ary $\alphabet$-trees such that $C$
is satisfiable $\wrt$ ${\cal T}$ $\iff$ the language ${\cal L}({\cal A}_{C,{\cal T}})$
accepted by ${\cal A}_{C,{\cal T}}$ is nonempty. The parameters of the automaton are as
follows:
\begin{enumerate}
  \item $N_P=\primitiveconcepts (C,{\cal T})$,
    $N_{cF}=\concretefeatures (C,{\cal T})$,
    $Q=\definedconcepts (C,{\cal T})$, $q_0=B_i$
  \item $K$ is the set of concepts appearing as arguments in the branching tuple of $C$ $\wrt$ ${\cal T}$:
    $K=\{d_1,\ldots ,d_n:(d_1,\ldots ,d_n)=\branchingtuple (C,{\cal T})\}$ (Definition
    \ref{branchingfactor})
  \item $\delta (B)$ is obtained from the axiom $B\doteq E$ in $(\tpc
    )^*$ defining $B$, as follows. $E$ is of the form $\{S_1,\ldots
    ,S_n\}$, with $S=S_{prop}\cup S_{csp}\cup S_{\exists}$, for all
    $S\in\{S_1,\ldots ,S_n\}$. We transform $S_{\exists}$ into
    $S_{\exists}'=\{(\exists R.D,D):\exists R.D\in
    S_{\exists}\cap\reconcepts (C,{\cal T})\}\cup\{(f,D):\exists f.D\in
    S_{\exists}\cap\feconcepts (C,{\cal T})\}$. We transform $S_{csp}$ into
    $S_{csp}'=\{P(u_1,u_2):u_1,u_2\in K^*N_{cF}\mbox{ and }\exists
      (u_1)(u_2).P\in S_{csp}\}$, if $x$ binary, and into $S_{csp}'=\{P(u_1,u_2,u_3):u_1,u_2,u_3\in K^*N_{cF}\mbox{ and }\exists
      (u_1)(u_2)(u_3).P\in S_{csp}\}$, if $x$ ternary. We get $S'=S_{prop}\cup
    S_{csp}'\cup S_{\exists}'$. Finally, $\delta
    (B)=\displaystyle\bigvee _{S\in E}\bigwedge _{X\in S'}X$.
  \item The remaining part is to determine the right partition of $Q$;
    the right partial order $\geq$ on the elements of the partition; and
    those elements of the partition that are accepting, i.e.,
    constituting the set ${\cal F}$ (see Definition 
    \ref{weakalternatingautomaton}). Without loss of generality, we assume that the axioms of
    ${\cal T}$ are given in the form $B\doteq\{S_1,\ldots ,S_n\}$, with $S=S_{prop}\cup
    S_{csp}\cup S_{\exists}\cup S_{\forall}$, for all
    $S\in\{S_1,\ldots ,S_n\}$. In other words, we suppose that in $\tpc$, each of the
    axioms, $B\doteq E$, has gone through the process of computing the
    first DNF, $\dnfone$, of the right-hand side, $E$. Going from $\dnfone$ to $\dnftwo$
    involves transforming $S=S_{prop}\cup S_{csp}\cup S_{\exists}\cup
    S_{\forall}$ into $S^f=S_{prop}\cup S_{csp}\cup S_{\exists}^f$,
    with each $\exists R.C$ in $S_{\exists}^f$ verifying
    $C=C_1\sqcap\ldots\sqcap C_{m_1}\sqcap C_{m_1+1}\sqcap\ldots\sqcap
    C_{m_2}$, such that $\{\exists R.C_1,\ldots ,\exists R.C_{m_1},\forall R.C_{m_1+1},\ldots
    ,\forall R.C_{m_2}\}\subseteq S_{\exists}\cup S_{\forall}$ ($m_1\geq
    1$, if $R$ is functional, and $m_1=1$, otherwise). This
    decreasing property implies that no defined concept in $(\tpc )^*$, other than
    the ones in $\tpc$, ``directly uses'' itself. However, if in $\tpc$, a defined
    concept $B$ ``uses'' (specifically, ``directly uses'') itself, it
    might give birth in $(\tpc )^*$ to a new defined concept which
    (directly) ``uses'' $B$, leading thus to a cycle of length strictly
    greater than one ---all defined concepts of such a cycle will
    constitute one element of the partition. The right partition is computed by the procedure 
    of Figure \ref{partition}. The right partial order, $\geq$, is as
    follows: given two elements $Q_i$ and $Q_j$ of the partition, $Q_i\geq Q_j$ if
    there exist $B_1\in Q_i$ and $B_2\in Q_j$, such that $B_1$
    ``uses'' $B_2$ in $(\tpc )^*$. The set ${\cal F}$ of accepting
    subsets of $Q$ is the set of all elements $Q_i$ of the partition,
    such that $Q_i$ contains no state consisting of an
    eventuality defined concept of $(\tpc )^*$ ---in other words, all
    states in such a $Q_i$ are noneventuality defined concepts of
    $(\tpc )^*$.
\end{enumerate}
\end{definition}
\begin{figure}
\begin{enumerate}
  \item[] {\bf Input:} the closure $(\tpc )^*$ of a TBox ${\cal T}$
    augmented with a concept $C$, $\tpc$
  \item[] {\bf Output:} partition of the set of defined concepts in $(\tpc )^*$
  \item[] Initially, no defined concept of $(\tpc )^*$ is marked;
  \item[] $k=1$;
  \item[] {\em while}($(\tpc )^*$ contains defined concepts that are not marked)\{
  \item[] \hskip 0.2cm consider a non marked defined concept $B_1$
    from $(\tpc )^*$;
  \item[] \hskip 0.2cm mark $B_1$;
  \item[] \hskip 0.2cm $\usesb1\leftarrow\displaystyle\bigcup
    _{B_2\mbox{ ``uses'' }B_1}\{B_2\}$;
  \item[] \hskip 0.2cm {\em if} $B_1\in\usesb1$\{
  \item[] \hskip 0.4cm $\partition [k]\leftarrow\usesb1$;
  \item[] \hskip 0.4cm mark all defined concepts of $(\tpc )^*$
    occurring in $\usesb1$;
  \item[] \hskip 0.4cm \}
  \item[] \hskip 0.2cm {\em else} $\partition [k]\leftarrow\{B_1\}$;
  \item[] \hskip 0.2cm $k\leftarrow k+1$;
  \item[] \hskip 0.2cm \}
\end{enumerate}
\caption{Partition of the set of defined concepts in the closure $(\tpc )^*$ of a TBox ${\cal T}$
  augmented with a concept $C$, $\tpc$.}\label{partition}
\end{figure}
\section{Discussion 1}\label{discone}
Theorem \ref{gafrunthm} and Corollary \ref{xdlsatisfiability} provide a tableaux-like procedure for 
the satisfiability of an $\xdl$ concept $\wrt$ an $\xdl$ weakly cyclic 
TBox. To understand how such a procedure will work in practice,
suppose that the $\rcc8$-like spatial constraints at the different
nodes of the output tree $t$ of the procedure of Figure \ref{finitekarytree}, are
not processed while the tree is being built. In other words, we
suppose that we first build the tree, and then process the (global)
CSP of the tree, which is given by the sets $S_{csp}$, over the sets
$S$ labelling the nodes of $t$. The building of $t$ is clear from
Theorem \ref{gafrunthm} and Corollary \ref{xdlsatisfiability}, which, among other things, offer
a bound on the size of $t$, as well as what is referred to in standard 
tableaux-like algorithms as a blocking condition. If such a tree is
successfully built, according to Theorem \ref{gafrunthm}, it can be extended to 
a full $\frun$ which satisfies the accepting subcondition related to
the states infinitely often repeated in the branches. We now need to
check the other accepting subcondition, which is the consistency of
the CSP of $t$, whose definition can be derived from that of the
(global) CSP of a run, as follows:
\begin{enumerate}
  \item For all nodes of $t$ that are not marked, and for all
    directions $d\in K$ such that $u$ has a $d$-successor $v$ in $t$, the
    non marked $d$-successor of $u$ in $t$ is $v$, if
    $v$ is not a marked node, and $i_v$, otherwise. If $u$ has a
    $d_{i_1}\ldots d_{i_n}$-successor in $t$, $n\geq 2$, then the non
    marked $d_{i_1}\ldots d_{i_n}$-successor of $u$ in $t$ is the non
    marked $d_{i_2}\ldots d_{i_n}$-successor of $v$ in $t$, where $v$
    is the non marked $d_{i_1}$-successor of $u$ in $t$.
  \item for all nodes $v$ of $t$, of label
$t (v)=(Y_v,L_v,X_v)\in2^{\hf}\times c(2^{\lits (N_P)})\times
2^{\consts (x,K,N_{cF})}$, the argument $X_v$ gives rise to the CSP of $t$ at $v$,
    $\csp _v(t)$, whose set of variables, $V_v(t)$, and set of constraints,
    $C_v(t)$, are defined as follows:
  \begin{enumerate}
    \item Initially, $V_v(t)=\emptyset$ and $C_v(t)=\emptyset$
    \item for all $\ksgchains$ $d_{i_1}\ldots d_{i_n}g$ appearing in
      $X_v$, create, and add to $V_v(t)$, a variable $\langle
      w,g\rangle$, where $w$ is the non marked $d_{i_1}\ldots
      d_{i_n}$-successor of $v$ in $t$
    \item if $x$ binary, for all $P(d_{i_1}\ldots d_{i_n}g_1,d_{j_1}\ldots d_{j_m}g_2)$ in $X_v$, add the constraint\\
      $P(\langle w_1,g_1\rangle,\langle w_2,g_2\rangle)$ to $C_v(t)$,
      where $w_1$ is the non marked $d_{i_1}\ldots
      d_{i_n}$-successor of $v$ in $t$, and $w_2$ the non marked $d_{j_1}\ldots
      d_{j_m}$-successor of $v$ in $t$
    \item similarly, if $x$ ternary, for all $P(d_{i_1}\ldots
      d_{i_n}g_1,d_{j_1}\ldots d_{j_m}g_2,d_{l_1}\ldots d_{l_p}g_3)$
      in $X_v$,
      add the constraint
      $P(\langle w_1,g_1\rangle,\langle
      w_2,g_2\rangle,\langle
      w_3,g_3\rangle)$ to $C_v(t)$, where $w_1$, $w_2$ and $w_3$ are,
      respectively, the non marked $d_{i_1}\ldots
      d_{i_n}$-successor, the non marked $d_{j_1}\ldots
      d_{j_m}$-successor, and the non marked $d_{l_1}\ldots
      d_{l_p}$-successor of $v$ in $t$
  \end{enumerate}
  \item the CSP of $t$, $\csp (t)$, is the CSP whose set of variables,
    ${\cal V}(t)$, and set of constraints, ${\cal C}(t)$, are defined
    as ${\cal V}(t)=\displaystyle\bigcup _{v\mbox{ node of }t}V_v(t)$ and
    ${\cal C}(t)=\displaystyle\bigcup _{v\mbox{ node of }t}C_v(t)$.
\end{enumerate}
Contrary to the CSP of a run, which is potentially infinite, the CSP
of $t$ is finite, and can thus be checked for consistency using
existing algorithms, as explained in Subsection
\ref{admissibility}. In particular, if the constraints, of the form
$P(u_1,u_2)$, if $x$ binary, or $P(u_1,u_2,u_3)$, if $x$ ternary,
appearing in the sets $X_v$, where $v$ is a node of $t$, are $x$
atomic relations, then the CSP of $t$ can be solved in deterministic polynomial time 
in the number of its variables. In general, however, one has to use a
search algorithm, which needs nondeterministic polynomial time in the number of
variables (again, the reader is referred to
Subsection \ref{admissibility}, and to the pointers to the appropriate 
litterature).

The described tableaux-like procedure can be made much more efficient by
pruning the search space with constraint propagation during the construction of $t$,
as sketched at the end of Subsection \ref{overview}. The basic idea is 
that, we do not wait until the completion of the construction of $t$,
to process the CSP. Instead, whenever new constraints arise, resulting 
from the adding of a new node to the tree being built, we propagate them to 
the constraints already existing. In particular, this will potentially 
detect more dead-ends, and consequently shorten the search space. The reader is
referred to Subsection \ref{overview} for details.
\section{Discussion 2: Continuous spatial change}\label{disctwo}
I open the section with my answer to the question of whether AI should
reconsider, or revise its challenges:
\begin{flushright}
\begin{scriptsize}
``{\bf AI
to the servive of the Earth as the Humanity's global,
continuous environment: the role of continuous (spatial) change in
building a lasting global, locally plausible democracy:}\\
(Cognitive) AI, which is guided by cognitively plausible assumptions on the physical world, such
as, e.g., {\em ``the continuity of (spatial) change''}, will start touching at its actual
success, the day it will have begun to serve, in return, as a source of inspiration for lasting
solutions to challenges such as, a World's globalisation respectful of local, regional believes
and traditions. One of the most urgent steps, I believe, is the implementation, in the Humanity's
global mind, of the idea of ``continuous change'', before any attempt of discontinuous
globalisation of our continuous Earth reaches a point of non return.''\\
A(mar)I(sli) 2003.
\end{scriptsize}
\end{flushright}
\subsection{The abstract objects as time intervals and the roles as
  the {\em meets} relation}
Instead of interpreting the nodes of the $k$-ary structures as time points, and the roles as
discrete-time immediate-successor relations, we can interpret the former as time intervals, the latter as
the {\em meets} relation of Allen's time interval Relation Algebra (RA) \cite{Allen83b}. In
particular, the satisfiability of a concept $\wrt$ a weakly cyclic TBox will remain decidable.
We get then a new family of languages for (continuous) spatial change. This new
spatio-temporalisation of $\alcd$ can be summarised as follows:
\begin{enumerate}
  \item Temporalisation of the roles, so that they consist of $n+l$
    Allen's {\em meets} relations
    $m_1,\ldots ,m_n,m _{n+1},\ldots ,m_{n+l}$,
    of which the $m_i$'s, with $i\leq n$, are general, not necessarily functional relations,
    and the $m _i$'s, with $i\geq n+1$, functional relations.
  \item Spatialisation of the concrete domain ${\cal D}$, in a similar way as we did for the
    first family: the concrete domain is is  generated  by a spatial RA such as the
    Region-Connection Calculus RCC8 \cite{Randell92a}.
\end{enumerate}
In Examples \ref{xdlzocda}, \ref{xdlztrcc}, \ref{xdlzorcc} and \ref{xdlzoatra}, instead of interpreting 
the roles (including the abstract features) as discrete-time
accessibility relations, we can, and do in the rest of the section,
interpret them as durative-time meets relations \cite{Allen83b}. In
Example \ref{xdlztrcc}, for instance, the abstract features $f_1$ and $f_2$
can be so interpreted. The motion of the corresponding spatial scene
(Figure \ref{sectorstwo}), when it reaches, for instance, Submotion B, remains in that
configuration for a (durative) while, before reaching, without
discontinuing, Submotion C in which it remains another while: in this respect, Submotion B ``meets'' Submotion C, which
is indicated with the abstract feature $f_1$.
\subsection{The properties of durativeness, continuity and density of (spatial) change}
The discussion is aimed at clarifying cognitively plausible assumptions on spatial change in
the physical world: {\em durativeness}, {\em continuity} and {\em density}. In
the particular case of motion of a spatial scene, for instance, this intuitively means that, on
the one hand, once the scene has reached a certain configuration, it remains in that
configuration for a (durative) while, before eventually reaching a distinct configuration; and,
on the other hand, the transition from a configuration c to the very first future configuration
c' distinct from c, respects some continuity condition, so that c' is a neighbour of c, in a
sense to be explained shortly. The transition also fulfils a {\em density} criterion, in the
sense that, there is no temporal gap in the spatial scene, between the end of configuration c
and the beginning of configuration c'. The discussion is related to continuity as discussed in
\cite{Galton97a}.

The theory of conceptual neighbourhoods is well-known in qualitative spatial and temporal
reasoning (QSTR) ---see, for instance, \cite{Freksa92a}. QSTR constraint-based languages
consist mainly of RAs. In the spatial case, for instance, the atoms of such an RA, in finite
number, are built by defining an appropriate partition of the spatial domain at hand, on
which the RA is supposed to represent knowledge, as constraints on n-tuples of
objects, where n is the arity of the relations. We say appropriate partition,
in the sense that the partition has to fulfil some requirements, such as
cognitive adequacy criteria, so that the obtained RA reflects, for instance,
the common-sense reasoning, or the reasoning required by the task the RA
is meant to be used for, as much as possible. The regions of the partition are
generally continuous, and each groups together elements of the universe which do
not need to be distinguished, because, for instance, the task at hand does not need,
or Humans do not make, such a distinction. Given two
atoms $r_1$ and $r_2$ of such an RA, $r_2$ is said to be a conceptual neighbour
of $r_1$, if the union of the corresponding regions in the partition is continuous,
so that one can move from one to the other without traversing a third region of
the partition. The conceptual neighbourhood of $r_1$ is nothing else than the set
of all its conceptual neighbours, including $r_1$ itself. The conceptual
neighbourhood of a general relation, which is a set of atoms, is the union of the
conceptual neighbourhoods of its atoms. The meets relation in Allen's
RA \cite{Allen83b}, for instance, has two conceptual neighbours other than
itself, which are before ($\allenb$) and overlaps ($\alleno$); the $\rcc8$ relation
$\rcctpp$ has $\rccpo$, $\rcceq$ and $\rccntpp$ as conceptual neighbours. The
conceptual neighbourhoods of the atoms, $e$, $l$, $o$ and $r$, of the $\apra$ binary
RA of 2-dimensional orientations in \cite{Isli00b} are, respectively, $\{e,l,r\}$,
$\{e,l,o\}$, $\{l,o,r\}$ and $\{e,o,r\}$. Concerning the ternary RA $\atra$ in
\cite{Isli00b}, an atom $b_1'b_2'b_3'$ is a conceptual neighbour of an atom
$b_1b_2b_3$, where the $b_i$'s and the $b_i'$'s are $\apra$ atoms, if and only if the
$\apra$ atoms $b_i$ and $b_i'$, $i=1\ldots 3$, are conceptual
neighbourhoods of each other. In Example \ref{xdlztrcc}, for instance, it is the case that the relation
on any pair of the involved objects
(the objects o1, o2 and o3 in Subscene 1; and
the objects q1, q2 and q3 in Subscene 2), when
moving from the current
atomic submotion to the next, either remains the same, or changes to a relation
that is a conceptual neighbour.
The transition from
Submotion $A$ to Submotion $B$ involves only the change of the $\rcc8$ relation on
the pair (o2,o3) from TPP to its conceptual neighbour NTPP; one might then argue
that, because the distinction between the TPP and NTPP relations involves only a 0- or
1-dimensional region, it might happen that the transition from the TPP
configuration to the NTPP configuration of the pair (o2,o3) is not durative. Nevertheless, even in such
extreme situations, the time required for the scene's motion to achieve the
transition is considered as an interval.

Without loss of generality, we restrict the remainder of the discussion to one
member of our $\xdl$ family of theories
for continuous spatial change, which is $\xdlrcc$, whose concrete domain is generated by $\rcc8$.
We denote by $M$ a motion of a spatial scene ${\cal S}$ composed on $n$
objects, $O_1,\ldots ,O_n$.\footnote{The objects $O_1,\ldots ,O_n$ are regions of
a topological space.} For all $i,j\in\{1,\ldots ,n\}$, $i<j$, we denote
by ${\cal S}_{ij}$ the subscene of ${\cal S}$ composed of objects $O_i$ and $O_j$;
by $M_{ij}$ the restriction of motion $M$ to subscene ${\cal S}_{ij}$;
by $M_{ij}^1,\ldots ,M_{ij}^{n_{ij}}$ the $n_{ij}\geq 1$ atomic submotions of $M_{ij}$;
by $I_{ij}^k$, $k\in\{1,\ldots ,n_{ij}\}$, the interval during which atomic submotion
$M_{ij}^k$ takes place; and
by $r_{ij}^k$, $k\in\{1,\ldots ,n_{ij}\}$, the $\rcc8$ relation of the pair $(O_i,O_j)$
during Submotion $M_{ij}^k$ of Subscene ${\cal S}_{ij}$.
Submotions $M_{ij}^k$ and $M_{ij}^{k+1}$, $k\in\{1,\ldots ,n_{ij}-1\}$, of $M_{ij}$ are
such that $M_{ij}^{k+1}$ immediately follows $M_{ij}^k$; in other words, the
intervals during which they hold are related by the {\em meets} relation:
$\allenm (I_{ij}^k,I_{ij}^{k+1})$.
\begin{definition}[maximal atomic submotion]
The atomic submotion $M_{ij}^k$, $k\in\{1,\ldots ,n_{ij}\}$, of subscene
${\cal S}_{ij}$ is said to be maximal if $M_{ij}$ has no atomic submotion that stricly
subsumes $M_{ij}^k$.
\end{definition}
\begin{definition}[continuous motion of a 2-object subscene]The motion of Subscene
${\cal S}_{ij}$ is said to be continuous if it can be decomposed into $n_{ij}$ maximal
atomic submotions $M_{ij}^1,\ldots ,M_{ij}^{n_{ij}}$ such that $r_{ij}^{k+1}$ is a
conceptual neighbourhood of $r_{ij}^k$, for all $k=1,\ldots ,n_{ij}-1$.
\end{definition}
\begin{definition}[continuous motion of scene ${\cal S}$]The motion of Scene
${\cal S}$ is said to be continuous if its restriction to any of its 2-object
subscenes is continuous.
\end{definition}
\subsection{Qualitative probabilistic decision making}
The previous discussion clearly suggests that one can design a qualitative probablistic
decision maker which, given a physical system such as, for instance, the one in Example
\ref{xdlztrcc}, would decide which Submotion the system should enter next. The
continuity assumption implies that the submotion to enter next should fulfil the
conceptual neighbourhood condition. If we assume a uniform probability distribution,
then we can easily give the conditional probabilities governing the motion restricted
to any pair (o1,o2) of the involved objects. For the purpose, we denote by $p(r'(o1,o2)|r(o1,o2))$ the probability  for the relation on the pair (o1,02), to be $r'$ at the next submotion of the scene, given that it is currently $r$. If $n$ is the
number of conceptual neighbours of $r$, then clearly:\\
$
p(r'(o1,o2)|r(o1,o2))=
  \left\{
    \begin{array}{l}
                   \frac{1}{n},\mbox{ if $r'$ is a conceptual neighbour of $r$}\\
                   0,\mbox{ otherwise}
    \end{array}
  \right.
$
\section{Discussion 3}\label{discthree}
As explained in Section \ref{discone}, the construction of the tree $t$,
ouput of the procedure of Figure \ref{finitekarytree}, can be done in such a way
that, the solving of the CSP of $t$ 
is entirely left until the end of the construction of the tree itself
---a two-step construction, with no constraint propagation during the
first step. The CSP of $t$ is a finite conjunction of constraints of the form 
$P(x_1,\ldots ,x_n)$, $P$ being a predicate of the concrete domain,
and $x_1,\ldots ,x_n$ variables. Consistency of the CSP is decidable
thanks to the admissibility of the concrete domain. We restricted the work 
to admissible concrete domains generated by $\rcc8$-like qualitative spatial
languages \cite{Randell92a,Egenhofer91b}, for we wanted such languages 
to be combined with modal temporal logics in a way leading to flexible 
domain-specific languages for spatial change in general, and for
motion of spatial scenes in particular. The reader should easily see that our
decidability results (Theorem \ref{gafrunthm} and Corollary \ref{xdlsatisfiability}) extend to any admissible concrete domain, in the
sense of ``concrete domain'' and ``admissibility'' which we have been 
using \cite{Baader91a} (Definition \ref{cddefinition} and \ref{cdadmissibility}).
\section{Discussion 4: adding ``atemporal'' roles to $\xdl$}
Another important point to mention is that, the roles of $\xdl$ are
temporal. The $\dnf2$ of a concept $C$  is of the form
$\dnf2(C)=S_{prop}\cup S_{csp}\cup S_{\exists}$, where, in
particular, $S_{prop}$ is a set of primitive concepts and negated
primitive concepts (literals), representing a
conjunction of propositional knowledge. It is known that $\alcd$ with an
admissible concrete domain and an acyclic TBox is
decidable \cite{Baader91a}; i.e., if the concrete domain $D$ is admissible then, satisfiability of an $\alcd$ concept $\wrt$ an
acyclic TBox is decidable. If we denote by $\alcf$ the DL $\alc$ \cite{Schmidt-Schauss91a}
augmented with abstract features, then clearly $\alcf$ is a sublanguage
of $\alcd$, and thus satisfiability of an $\alcf$ concept with respect
to an $\alcf$ acyclic TBox is decidable. $\alcf$ is particularly
important for the representation of structured data, such as data in
$\xml$ documents (see, e.g., \cite{Buneman01a,Fan02a}), thanks, among
other things, to its abstract features, which allow it to access
specific paths. We can add ``atemporal''
roles to $\xdl$, so that $\alcf$ gets subsumed, again without compromising
our decidability results. Such an extension would offer a
representational tool for the history of structured data in general,
and of data in $\xml$ documents in particular. It would also offer a
representational tool for event models in high-level computer vision
(see, e.g., \cite{Badler75a,Neumann83a}), which are ``a representation 
of classes of events and a tool to recognise events in a given scene
\cite{Neumann83a}''. The concepts of such a language are the
atemporal concepts and the temporal concepts, as given by the
definition below:
\begin{definition}
Let $x$ be an RA from the set $\{\rcc8 ,\cdalg ,\atra\}$. Let $N_C$,
$N_R^a$, $N_R^t$
and $N_{cF}$ be mutually disjoint and countably infinite sets of concept
names, atemporal role names, temporal role names, and concrete
features, respectively;
$N_{aF}^a$ a countably infinite subset of $N_R^a$ whose elements are
atemporal abstract features; and
$N_{aF}^t$ a countably infinite subset of $N_R^t$ whose elements are
temporal abstract features. A temporal (concrete)
feature chain is any finite composition $f_1^t\ldots f_n^tg$ of $n\geq
0$ temporal abstract
features $f_1^t,\ldots ,f_n^t$ and one concrete feature $g$. The set of
atemporal concepts and the set of temporal concepts are the smallest sets such that:
\begin{enumerate}
  \item\label{defxdlconceptsone} $\top$ and $\bot$ are
    atemporal concepts
  \item\label{defxdlconceptstwo} a concept name is an atemporal
    concept
  \item\label{defxdlconceptsthree} if
    $C^a$ and $D^a$ are atemporal concepts;
    $C^t$ and $D^t$ are temporal concepts;
    $R^a$ is an atemporal role (in general, and an atemporal abstract feature in particular);
    $R^t$ is a temporal role (in general, and a temporal abstract feature in particular);
    $g$ is a concrete feature;
    $u_1^t$, $u_2^t$ and $u_3^t$ are temporal feature chains; and
    $P$ is a predicate,
    then:
    \begin{enumerate}
      \item $\neg C^a$,
            $C^a\sqcap D^a$,
            $C^a\sqcup D^a$,
            $\exists R^a.C^a$,
            $\forall R^a.C^a$ are atemporal concepts;
      \item $\neg C^t$,
            $C^t\sqcap D^t$,
            $C^t\sqcup D^t$,
            $\exists R^t.C^t$,
            $\forall R^t.C^t$ are temporal concepts;
      \item $C^a\sqcap C^t$,
            $C^a\sqcup C^t$,
            $\exists R^t.C^a$,
            $\forall R^t.C^a$ are temporal concepts; and
      \item $\exists (u_1^t)(u_2^t).P$, if $x$ binary,
            $\exists (u_1^t)(u_2^t)(u_3^t).P$, if $x$ ternary, are
            temporal concepts.
    \end{enumerate}
\end{enumerate}
\end{definition}
A TBox $T$ is now weakly cyclic if it satisfies the following two conditions:
\begin{enumerate}
  \item Whenever $A$ uses $B$ and $B$ uses $A$, we have $B=A$.
  \item All possible occurrences of a defined concept $B$ in the right
    hand side of the axiom defining $B$ itself, are within subconcepts of 
    $C$ of the form $\exists R.D$ or $\forall R.D$, $C$ being the right 
    hand side of the axiom, $B\doteq C$, defining $B$, with $R$ being
    a temporal role ---in other words, we do not allow any cyclicity
    in any atemporal part of the TBox.
\end{enumerate}
\section{Conclusion}\label{conclusion}
We have described how to enhance the expressiveness of modal temporal logics with
qualitative spatial constraints. The theoretical framework consists of a
spatio-temporalisation of the $\alcd$ family of description logics with a concrete
domain \cite{Baader91a}, obtained by temporalising the roles, so that
they consist of $m+n$ immediate-successor (accessibility) relations,
the first $m$ being general, not necessarily functional roles, the
other $n$ abstract features; and spatialising the concrete domain,
which is generated by an $\rcc8$-like qualitative spatial language
\cite{Randell92a,Egenhofer91b}. The result is a family $\xdl$ of
languages for qualitative spatial change in general, and for motion of 
spatial scenes in particular. We considered $\xdl$ weakly cyclic
TBoxes, expressive enough to capture most of existing modal temporal
logics -which was shown for Propositional Linear Temporal Logic
$\pltl$, and for the $\ctl$ version of the full branching modal temporal logic $\ctlstar$
\cite{Emerson90a}. We proved that satisfiability of a concept $\wrt$ such a TBox
is decidable, by reducing it to the emptiness problem of a weak
alternating automaton \cite{Muller92a} augmented with qualitative
spatial constraints. 
The accepting condition of a run of such an augmented weak
alternating automaton involves, additionally to the states infinitely 
often repeated, consistency of a CSP (Constraint Satisfaction Problem)
potentially infinite.
Nevertheless, the emptiness problem was shown to remain decidable.
A tableaux-like procedure for the satisfiability of an $\xdl$ concept
$\wrt$ an $\xdl$ weakly cyclic TBox, which we have discussed, is
straightforwardly obtainable from our results.

We also discussed various extensions of the work. In particular, we discussed that, if, instead of interpreting the nodes of the
$k$-ary structures as time points, and the roles as
immediate-successor relations, we interpreted the former as time
intervals, the latter as the {\em meets} relation of Allen's time
interval Relation Algebra (RA) \cite{Allen83b}, the satisfiability of
a concept $\wrt$ a weakly cyclic TBox remained decidable. This led 
to a new family of languages for continuous spatial change.
\section*{Acknowledgement}
I had discussions with Bernd Neumann on issues related to the presented work.
\bibliographystyle{style-subdirectory/acmtrans}
\bibliography{biblio}
\end{document}